%% file: ace_brain05.tex
\documentclass[]{academic_template}
\usepackage[toc,page,header]{appendix}

\usepackage{marvosym}

\usepackage{bbding}
\usepackage{soul}

\usepackage{amssymb}
\usepackage{amsmath}

\usepackage{makecell}
\usepackage{algorithm}      
\usepackage{algpseudocode}  
\usepackage{tabularx}
\usepackage{booktabs}
\usepackage{graphicx}
\usepackage{array}
\newcolumntype{C}{w{c}{3em}}
\usepackage{listings}
\usepackage{hyperref}
\usepackage{url}
\usepackage{booktabs}
\usepackage{wrapfig}

\usepackage{colortbl}
\usepackage{caption}
\usepackage{subcaption}

\usepackage{amsthm}
\newtheorem{theorem}{Theorem}

\setlogoheight{8mm}   

\setlogospacing{0mm}
\setsjtublue 

\settitlerulethickness{3pt}

\abstractboxon 
\setabstractframecolor{gray} 
\setabstractbgcolor{gray!10} 
\setlogotolineshift{0mm}

\settoprulethickness{2.5pt}  
\setbottomrulethickness{1.5pt}

\usepackage{dblfloatfix}
\usepackage{float}

\usepackage{tabularx}

\usepackage{multicol}
\usepackage{multirow}
\usepackage[table]{xcolor}

\usepackage{natbib}
\setcitestyle{numbers}
\setcitestyle{square}
\usepackage{graphicx}

\usepackage{xspace}
\newcommand{\myname}{ACE-Brain-0.5\xspace}

\newcommand{\ourMethod}{ACE-Brain-0.5\xspace}

\title{{\ourMethod}: A Unified Embodied Foundational Model for Physical Agentic AI}

\author{ACE-Brain Team}

\affiliation{
Please see \hyperref[sec:contribution_author_list]{Contributions and Author List}
for more author details.
}

\abstract{

Embodied AI is moving from isolated perception or action modules toward physical agents that can understand the world, plan under goals, act through robot bodies, monitor whether their behavior is making progress, and improve from accumulated experience. Existing systems address different parts of this loop in isolation: end-to-end policies (vision-language-action or world-action) are effective at generating robot actions but often provide limited spatial reasoning, long-horizon planning, and execution assessment; robot-agent systems can orchestrate multiple tools or specialist models but do not learn a single shared robot representation. This fragmentation limits the development of general Physical Agentic AI. We present \myname{}, a unified embodied foundation model for Physical Agentic AI that organizes robot intelligence into five tightly coupled cognitive functions: spatial perception, decision making, embodied interaction, self-monitoring, and self-improvement. Built on ACE-Brain-0, which established spatial intelligence as a shared scaffold across heterogeneous robot platforms, \myname{} extends an understanding-centric embodied model into a closed-loop embodied foundation model. A single 8B backbone directly instantiates the first four functions within a closed loop: it grounds objects and affordances, reasons over 3D and egocentric spatial relations, decomposes high-level instructions into executable subgoals, generates navigation and manipulation actions, and estimates execution progress for verification and recovery. To unify these heterogeneous capabilities without cross-task interference, we introduce SSR+, which extends Scaffold–Specialize–Reconcile with a lightweight Reactivate stage after task-vector merging. The fifth function, self-improvement, is realized through a companion framework that incrementally updates an external execution state, i.e., task schemas, spatial memory, and failure-recovery cases, from accumulated rollout experience, enabling deployment-time adaptation. Across more than fifteen benchmarks spanning spatial cognition, grounding, navigation, manipulation, and progress evaluation, \myname{} improves over ACE-Brain-0 on 14 out of 18 spatial perception and grounding benchmarks, achieves competitive navigation and manipulation performance, and provides strong progress-estimation ability under both in-distribution and out-of-distribution settings. These results show that spatial perception, decision making, embodied interaction, self-monitoring, and self-improvement can be unified within a single robot foundation model, marking an early step toward general Physical Agentic AI.}

\date{\today}

\checkdata[Project Page]{\url{https://ace-brain-team.github.io/ACE-Brain-0.5/}}
\checkdata[Code]{\url{https://github.com/ACE-BRAIN-Team/ACE-Brain-0.5}}
\checkdata[Hugging Face]{\url{https://huggingface.co/ACE-Brain/ACE-Brain-0.5-8B}}

\begin{document}

\maketitle

\newpage
\tableofcontents

\newpage

\input{contexts/1_intro}

\input{contexts/2_related}
\input{contexts/3_method}

\input{contexts/4_exp}
\input{contexts/5_data}

\input{contexts/6_conclusion}


\clearpage
\bibliographystyle{unsrtnat}
\bibliography{references}

\clearpage
\input{contributions_author_list}

\clearpage

\appendix
\section{Appendix}

\input{appendix/appendix}
\clearpage

\end{document}

%% file: contexts/1_intro.tex
\section{Introduction}

Physical Agentic AI aims to build autonomous robotic agents that can perceive and understand the physical world, plan under long-horizon goals, execute actions through robot bodies, and continuously monitor whether their behavior is making progress~\cite{duan2022survey,xu2024robotics,chen2025robotwin20scalabledata}. Achieving such closed-loop intelligence requires perception, reasoning, planning, action, and self-assessment to operate as a unified system rather than as independent capabilities. Recent advances in embodied foundation models have substantially improved individual capabilities, including visual understanding~\cite{internvl3,bai2025qwen3vltechnicalreport}, language-conditioned reasoning~\cite{qwen3,gemini}, navigation~\cite{zhang2026qwennav,zhang2025navfom}, robot manipulation and control~\cite{black2024pi_0,ye2026world}, and execution assessment~\cite{lee2026roboreward,robometer2026}. Despite these advances, these capabilities remain fragmented and are largely developed in isolation rather than as a unified cognitive architecture, hindering the realization of general Physical Agentic AI.

\begin{figure}[!htb]
    \centering
    \includegraphics[width=\linewidth]{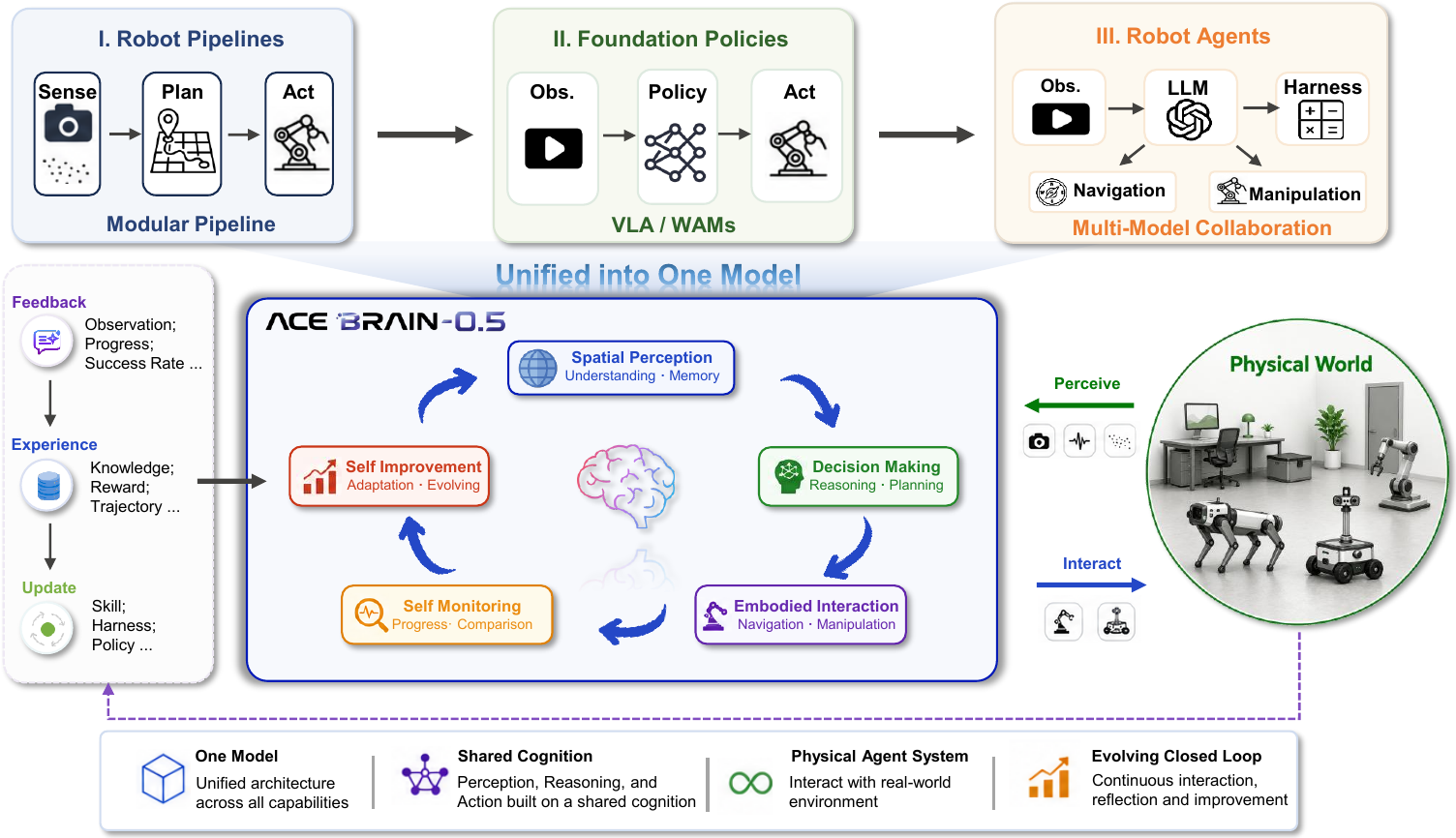}
    \caption{The evolution toward a Unified Embodied Foundation Model. Embodied intelligence has progressed from modular robot pipelines to end-to-end policies and multi-model robot agents. We envision the next paradigm as a Unified Robot Brain, in which all core cognitive functions are integrated within a single foundation model, enabling closed-loop robot cognition from spatial perception and decision making to embodied interaction, self-monitoring, and ultimately continual self-improvement.}
    \label{fig:paradigm}
\end{figure}

To realize general Physical Agentic AI, embodied intelligence has evolved through several paradigms, as illustrated in Fig.~\ref{fig:paradigm}. Classical robotics adopts a modular Sense--Plan--Act pipeline~\cite{nilsson1984shakey,brooks1986robust,murphy2000introduction}, where perception, planning, and control are independently designed. While interpretable and reliable, such systems suffer from pipeline fragmentation and limited generalization to open-world environments. End-to-end Vision-Language-Action (VLA) and World-Action models~\cite{pi05_2025,ye2026world,interleave-vla} instead learn robot policies directly from observations, substantially improving manipulation through large-scale training. However, they remain primarily action-centric, with limited support for spatial reasoning, long-horizon planning, and execution assessment. Robot-agent systems~\cite{ahn2022saycan, driess2023palme, liang2023code, huang2023voxposer,robocodex2024,xu2026roboagent} further extend autonomy by integrating planning, tool use, and specialist models, but rely on system-level orchestration rather than a shared cognitive representation. Consequently, existing paradigms advance different aspects of embodied intelligence, yet none learns a unified model for closed-loop robot autonomy.


Despite rapid progress, existing embodied foundation models remain fragmented in both capability coverage and system design (Table~\ref{tab:robot_brain_comparison}). 
End-to-end policy models, including the $\pi$ series~\cite{black2024pi_0, pi05_2025},
DreamZero~\cite{ye2026world}, GR00T~\cite{nvidia2025gr00tn1, nvidia2025gr00tn15}, and QwenVLA~\cite{wang2026qwen}, are strong at producing robot actions but offer limited spatial perception and task planning. 
Robot agents such as QwenRobot~\cite{zhang2026qwen} and ABot~\cite{yang2026abot} handle complex tasks through system-level orchestration but do not learn a unified robot model. Multimodal embodied models such as RynnBrain~\cite{dang2026rynnbrain} offer strong spatial understanding but cannot generate executable actions or monitor execution progress, whereas Cosmos 3~\citep{agarwal2026cosmos} unifies action generation and perceptual understanding within a single world-model architecture yet lacks a mechanism for self-evaluation.
Overall, there is no existing system that integrates spatial perception, decision making, embodied interaction, and self-monitoring within a single robot foundation model.

We argue that the next step is a \textbf{Unified Embodied Foundation Model}: a single foundation model
organized around five tightly coupled cognitive functions.
\textit{Spatial Perception} builds a representation of the environment through perception,
spatial reasoning, and memory.
\textit{Decision Making} performs long-horizon reasoning and task planning.
\textit{Embodied Interaction} carries out decisions through navigation and manipulation.
\textit{Self Monitoring} tracks execution progress and feeds failures back to replanning.
\textit{Self Improvement} enables continual adaptation from accumulated experience.
The key enabler across all these functions is \textbf{spatial intelligence}: the ability to localize objects, reason about 3D relationships, understand scenes from egocentric and allocentric views, and identify actionable regions. Because spatial understanding is required by navigation, manipulation, and instruction-following alike~\cite{liang2023code, ahn2022saycan, huang2023voxposer}, it serves as the shared representation that ties these functions together across different robot platforms.

Our prior work, ACE-Brain-0~\cite{acebrain0}, established this spatial foundation across spatial cognition, autonomous driving, low-altitude sensing, and embodied understanding,
using a Scaffold--Specialize--Reconcile (SSR) training strategy.
However, ACE-Brain-0 was primarily an understanding model: it perceived and reasoned but lacked a unified action interface and had no mechanism to monitor its own execution. In this work, we present \myname{}, the instantiation of the Unified Robot Brain paradigm. Built on ACE-Brain-0, \myname{} integrates Decision Making, Embodied Interaction, and Self Monitoring into a single model. It strengthens spatial perception through object grounding, affordance understanding, and spatial reasoning; supports long-horizon task planning; generates executable actions for both navigation and manipulation; and estimates execution progress for self-monitoring and error recovery. To integrate these heterogeneous capabilities without cross-task interference, we extend SSR with a \textit{Reactivate} stage (SSR+), combining task vector merging with lightweight fine-tuning to align outputs across grounding, navigation, manipulation, and progress estimation within one model. Extensive experiments spanning spatial cognition, embodied grounding, navigation, manipulation, and execution monitoring show that jointly training all cognitive functions
in one model consistently outperforms task-specific baselines, providing an early foundation toward more general and self-improving robotic systems.

\input{tables/ability_comparison}


\noindent\textbf{The main contributions of this work are summarized as follows:}

\begin{itemize}

\item We propose the \textbf{Unified Embodied Foundation Model} paradigm, which organizes robot intelligence into five tightly coupled cognitive functions: \textit{Spatial Perception}, \textit{Decision Making}, \textit{Embodied Interaction}, \textit{Self Monitoring}, and \textit{Self Improvement}.
\myname{} is presented as the first instantiation of this paradigm.

\item We develop \myname{}, a unified embodied foundation model that extends ACE-Brain-0 from a spatial understanding model to a closed-loop cognitive system, jointly supporting spatial perception, long-horizon decision making, executable navigation and manipulation, and execution monitoring within a single model.

\item We introduce the \textbf{SSR+} training strategy, which extends
Scaffold--Specialize--Reconcile with a Reactivate stage.
Building on recent advances in model merging~\cite{yang2026model},
SSR+ combines task vector merging with targeted fine-tuning to unify spatial reasoning, grounding, navigation, manipulation, and progress estimation in one model without cross-task interference.

\item Extensive experiments across more than fifteen benchmarks spanning five embodied intelligence domains demonstrate consistent improvements over prior models and task-specific baselines, validating the benefit of jointly training all cognitive functions within a single robot foundation model.

\end{itemize}

%% file: tables/ability_comparison.tex
\begin{table*}[t]
\centering

\definecolor{rbourscol}{HTML}{EAF3FF}
\definecolor{rbheadgray}{HTML}{F4F6F8}
\definecolor{rbfullgreen}{HTML}{2F6B3F}
\definecolor{rbpartialorange}{HTML}{A5671F}
\definecolor{rbnonered}{HTML}{9E3B3B}

\newcommand{\rbfull}{%
\textcolor{rbfullgreen}{\ensuremath{\checkmark}}%
}
\newcommand{\rbpartial}{%
\textcolor{rbpartialorange}{\ensuremath{\bullet}}%
}
\newcommand{\rbnone}{%
\textcolor{rbnonered}{\ensuremath{\times}}%
}

\newcommand{\rbhead}[1]{%
\parbox[c][4.5em][c]{2.05cm}{%
\centering
\bfseries
\fontsize{8.2}{8.8}\selectfont
#1
}%
}

\caption[
Comparison of recent embodied foundation models and robot systems.
]{
Comparison of recent embodied foundation models and robot systems
in terms of core robotic capabilities.
}
\label{tab:robot_brain_comparison}

\setlength{\tabcolsep}{2.2pt}
\renewcommand{\arraystretch}{1.16}

\resizebox{0.98\textwidth}{!}{%
\begin{tabular}{@{}
>{\raggedright\arraybackslash}m{5cm}
*{7}{>{\centering\arraybackslash}m{2.05cm}}
@{}}
\toprule

\rowcolor{rbheadgray}
\parbox[c][4.5em][c]{5cm}{%
    \raggedright
    \bfseries
    Model / System
}
&
\rbhead{Object\\Perception}
&
\rbhead{Spatial\\Understanding}
&
\rbhead{Task\\Planning}
&
\rbhead{Navigation}
&
\rbhead{Manipulation}
&
\rbhead{Execution\\Monitoring}
&
\rbhead{Self-\\Improving}
\\

\midrule

\rowcolor{rbourscol}
\textbf{ACE-BRAIN-0.5}
& \rbfull
& \rbfull
& \rbfull
& \rbfull
& \rbfull
& \rbfull
& \rbfull
\\

\textbf{Pelican-Unified 1.0} \cite{zhang2026pelican}
& \rbfull
& \rbpartial
& \rbpartial
& \rbnone
& \rbfull
& \rbnone
& \rbnone
\\


\textbf{HY-Embodied-0.5} \cite{team2026hy}
& \rbfull
& \rbfull
& \rbfull
& \rbnone
& \rbpartial
& \rbnone
& \rbnone
\\

\textbf{Cosmos3} \cite{agarwal2026cosmos}
& \rbfull
& \rbfull
& \rbfull
& \rbnone
& \rbfull
& \rbnone
& \rbnone
\\

\textbf{RynnBrain} \cite{dang2026rynnbrain}
& \rbfull
& \rbfull
& \rbfull
& \rbnone
& \rbnone
& \rbnone
& \rbnone
\\

\textbf{QwenVLA} \cite{wang2026qwen}
& \rbpartial
& \rbpartial
& \rbpartial
& \rbfull
& \rbfull
& \rbnone
& \rbnone
\\

\textbf{\(\pi\) Series} \cite{black2024pi_0, pi05_2025, intelligence2604pi0, intelligence2025pi}
& \rbfull
& \rbnone
& \rbfull
& \rbfull
& \rbfull
& \rbnone
& \rbnone
\\

\textbf{Embodied-R1.5} \cite{yuan2026embodied}
& \rbfull
& \rbfull
& \rbfull
& \rbnone
& \rbpartial
& \rbfull
& \rbpartial
\\



\textbf{MolmoAct2} \cite{fang2026molmoact2}
& \rbfull
& \rbfull
& \rbpartial
& \rbnone
& \rbfull
& \rbnone
& \rbnone
\\

\textbf{QwenRobot} \cite{zhang2026qwen, yuan2026qwen, zhang2026qwennav}
& \rbnone
& \rbnone
& \rbpartial
& \rbfull
& \rbfull
& \rbnone
& \rbnone
\\

\textbf{ABot} \cite{yang2026abot,chu2026abotn0}
& \rbnone
& \rbnone
& \rbpartial
& \rbfull
& \rbfull
& \rbnone
& \rbnone
\\

\textbf{Gemini Robotics-ER 1.6} \cite{graesser2026geminiroboticser}
& \rbfull
& \rbfull
& \rbfull
& \rbnone
& \rbfull
& \rbnone
& \rbnone
\\

\textbf{Helix 02} \cite{figure2026helix02}
& \rbfull
& \rbpartial
& \rbfull
& \rbfull
& \rbfull
& \rbpartial
& \rbnone
\\

\bottomrule

\end{tabular}%
}

\vspace{0.35em}
\begin{minipage}{0.98\textwidth}
\footnotesize
\textbf{Legend.}
\rbfull\ denotes full or explicit support;
\rbpartial\ denotes partial, restricted, or indirectly demonstrated
support; and
\rbnone\ denotes capabilities that are not explicitly supported or
not clearly demonstrated in publicly available materials.
Navigation refers to goal-directed or semantic navigation rather than
locomotion alone.
Execution monitoring requires explicit success detection, progress
tracking, failure identification, or error recovery.
Self-improving refers to the ability to improve model behavior, task
performance, or execution policy through feedback-driven adaptation,
post-training, self-correction, or iterative experience-based refinement.
\myname{} is the first embodied model with capabilities of object
perception, spatial understanding, task planning, end-to-end navigation,
end-to-end manipulation, execution monitoring, and self-improving
adaptation.
\end{minipage}

\end{table*}

%% file: contexts/2_related.tex
\section{Related Work}

\subsection{General Multimodal Models, Spatial Reasoning, and Embodied Planning}
Recent multimodal large language models have shown strong visual understanding and reasoning ability~\cite{qwen2.5-vl,bai2025qwen3vltechnicalreport,llava,chatgpt4o,claude4}, providing a general foundation for embodied perception. Early embodied MLLMs such as PaLM-E~\cite{driess2023palme} further connected language models with continuous sensor observations, enabling vision-language reasoning for robotic tasks. 
A subsequent line of work focuses on spatial grounding, where models map language instructions to physical entities, points, regions, or 3D locations. Representative systems~\cite{chen2024spatialvlm} and related grounding-based~\cite{yuan2024robopoint,zhou2025roborefer} methods improve object localization and affordance prediction. Beyond grounding, recent robot brain and embodied-agent models begin to integrate high-level reasoning with planning-oriented outputs, such as task decomposition, spatial pointing, trajectory prediction, or progress estimation~\cite{robobrain,robobrain2_5,mimo-embodied,dang2026rynnbrain,yuan2026embodied,team2026hy}. 
However, most existing models~\cite{vebrain,yuan2026embodied,agarwal2026cosmos,robobrain2,vlaser,pelican-vl} still emphasize either general visual reasoning, spatial grounding, or planning assistance, while action generation and execution monitoring are often handled by separate policies, tools, critics, or downstream modules. \myname is designed to close this gap by integrating physical-world perception, spatial reasoning, planning, action generation, and progress evaluation within a single embodied foundation model.

\subsection{End-to-End Manipulation Policy}

A prominent line of work maps perception and language directly to robot actions through end-to-end vision-language-action (VLA) models. The paradigm was established by RT-1~\cite{brohan2022rt1} and RT-2~\cite{zitkovich2023rt2}, and subsequently scaled by open generalist policies such as Octo~\cite{octo2024} and OpenVLA~\cite{kim2024openvla}, the latter trained on Open X-Embodiment~\cite{open_x_embodiment_rt_x_2023} across many embodiments. Hierarchical designs such as RT-H~\cite{belkhale2024rth} decompose actions into language-intermediate representations, while interleaved vision-text-action pretraining~\cite{eo1} unifies perception and control in a single autoregressive model. Subsequent designs explore flow-matching action generation, diffusion action experts, and hierarchical or ``fast-slow'' architectures, including $\pi_0$~\cite{black2024pi_0}, GR00T~N1~\cite{nvidia2025gr00tn1}, and CogACT~\cite{li2024cogact}. The most recent generation continues to push generalization and dexterity: $\pi_{0.5}$ extends flow-based VLAs toward open-world generalization~\cite{pi05_2025}, GR-2 and GR-3 scale generalist manipulation policies with large-scale video and robot data~\cite{cheang2024gr2,gr3_2025}, ABot-M0 studies unified manipulation through action manifold learning~\cite{yang2026abot}, $\pi_{0.7}$ further improves steerable generalist control with multimodal context conditioning~\cite{intelligence2604pi0}, and recent approaches inject perceptual priors such as gaze-regularized attention~\cite{pani2026gaze} into the action model. A parallel line, often termed world-action models~\cite{sun2026vlajepa,kim2026cosmospolicy,ye2026gigaworld,guo2026xwam,ye2026world,yuan2026fastwam}, couples action generation with predictive world modeling, forecasting future observations or latent states to learn environment dynamics that in turn improve control. Recent systems such as  DreamVLA, Being-H0.7, RynnVLA-002, and ABot-M0.5 further explore joint video-action generation, latent future reasoning and unified VLA-world-model training~\cite{zhang2026dreamvla,luo2026being,cen2025rynnvla,chen2026abotm05}. While these models excel at producing low-level control, they are typically specialized for manipulation and trained primarily on action or paired video data, which limits their general spatial perception, long-horizon reasoning, and self-assessment. \myname instead couples direct end-to-end action generation with broad perceptual and reasoning competencies in a single model.

\subsection{Foundation Models for Embodied Navigation}
A parallel body of research treats navigation as the core embodied capability.
Early vision-and-language navigation studies mainly operate in discrete
graph-based environments, where agents follow instructions by selecting
viewpoints or primitive actions~\cite{r2r,chen2022think}. VLN-CE further extends
language-guided navigation to continuous environments, making perception,
actuation, and error accumulation more realistic~\cite{vlnce}. With the rise of
large language and vision-language models, recent approaches increasingly use
LLM/VLM reasoning for instruction following, planning, and embodied decision
making~\cite{zhang2024vlnsurvey}. Representative agentic systems such as
NavGPT introduce explicit language-based planning and progress reasoning for
navigation~\cite{zhou2024navgpt}.

More recent work formulates navigation as a video or action-conditioned
foundation-model problem. NaVid, NaVILA, Uni-NaVid, and StreamVLN unify
egocentric observations with navigation actions under video-based or VLA-style
interfaces~\cite{zhang2024navid,cheng2024navila,zhang2024uninavid,wei2025streamvln}. Beyond task-specific
VLN, generalist navigation models such as OctoNav, NavFoM, and ABot-N0 extend
navigation across tasks, embodiments, and instruction types~\cite{gao2026octonav,zhang2025navfom,chu2026abotn0}.
AgentVLN further explores a VLM-as-Brain agentic framework with modular skills,
active exploration, and self-correction~\cite{xin2026agentvln}, while SID-VLN constructs
self-improving demonstrations for goal-oriented language-guided navigation~\cite{sidvln}.
In parallel, spatial-cognitive and semantic-scene-graph-based methods study how
agents can reduce spatial hallucination and build semantic world models beyond
pure geometry~\cite{bai2025endowing,kueble2026ssg}. However, these systems are
generally optimized around navigation-specific objectives, action interfaces, or
embodiment assumptions. \myname supports navigation and manipulation within one
framework, treating them as coordinated outputs of shared spatial understanding.

\subsection{Progress-based Reward Models}
Another line of research explores progress estimation as a reward signal for embodied agents. Instead of relying on hand-crafted rewards or privileged state information, these methods estimate task progress directly from visual observations and language instructions. Early approaches use pretrained VLMs as zero-shot progress or success estimators~\cite{roboclip, ma2023vipuniversalvisualreward, ma2024generative, rocamonde2024visionlanguage,qing2022survey}, while others train domain-specific progress predictors~\cite{ma2023liv, yang2024rank, zhang2025rewind, chen2025sarm}; however, zero-shot signals are often noisy and specialized models may overfit to particular tasks, embodiments, or scenes. TOPReward~\cite{chen2026topreward} further shows that pretrained video VLMs can provide zero-shot robotic reward signals by extracting task-progress information from internal token probabilities rather than prompting the model to output explicit numerical scores. Recent methods focus on fine-tuning VLMs or VLAs on robot trajectories: VLAC~\cite{zhai2025vision} co-trains action prediction with relative progress estimation, $\pi^*_{0.6}$~\cite{intelligence2025pi} and related value-learning methods model distance-to-goal progress, RoboReward~\cite{lee2026roboreward} learns discretized progress rewards from real-robot data, and RoboDopamine~\cite{tan2026robodopamine} explores process reward modeling with goal-image conditioning. RoboMeter~\cite{robometer2026} scales this direction by introducing RBM-1M, combining frame-level progress supervision with trajectory-level preference comparisons over expert, suboptimal, and failed rollouts. Building on this paradigm, \myname leverages RoboMeter-style data and evaluation with a stronger multimodal foundation model, improving progress-based reward estimation and enabling self-assessment to be integrated with broader embodied perception and action capabilities.

\subsection{Self-Improving Robotic Agents}

A growing line of work studies agents that improve through their own experience. Early LLM agents mainly relied on verbal reflection or trajectory feedback, turning failures into reusable hints for later attempts~\cite{shinn2023reflexion} or distilling corrections into model weights~\cite{yuan2025agentr}. More recent work~\cite{skillopt,skilllens,continualharness,kong2024tptu,kong2024qpo} treats the agent's external execution state as the object of optimization, such as reusable skills, prompts, memory, tools, or harnesses~\cite{skillopt,continualharness}. In robotics, many excellent works are also inspired by agentic engineering to build embodied agents~\cite{gca}, e.g., Code-as-Policy~\cite{liang2023code}, VoxPoser~\cite{huang2023voxposer}, RoVI~\cite{rovi}, and CaP-X~\cite{capx} use MLLMs to generate executable policy code, spatial value maps, visual instructions, or control programs that connect high-level reasoning with physical actions. However, closing the self-improvement loop in the physical world remains harder than in virtual environments, since robot rollouts are costly, resets are nontrivial, and success signals are often sparse. Recent robotic systems, therefore, introduce separate evaluators, critics, reward models, or progress predictors to provide feedback for self-directed practice~\cite{ghasemipour2025selfimproving,tian2025seear1,rise2026,li2025reflective,intelligence2025pi,xu2026roboagent,qing2024a2po,qing2025bitrajdiff,liu2025curricular,kong2025mastering}. \myname integrates the prerequisite capability of task-progress estimation and execution-state assessment into the same backbone that performs grounding, planning, and action generation, providing an internal self-assessment interface for future self-improving robotic agents.

%% file: contexts/3_method.tex
\section{Overview}

\subsection{Model Architecture}

\begin{figure}
    \centering
    \includegraphics[width=1.\linewidth]{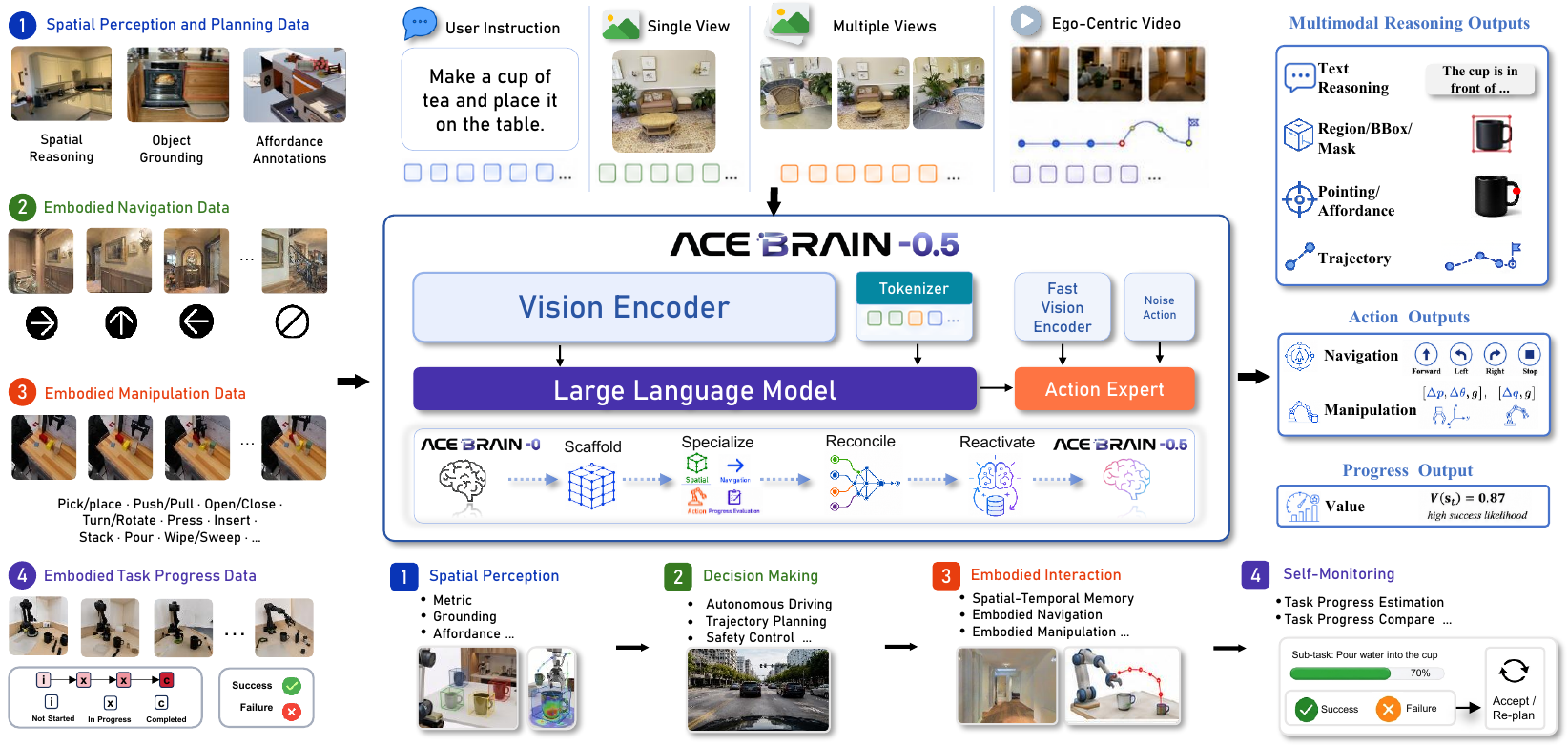}
    \caption{
    Overall architecture of \myname{}. Specifically, 
        first, omni-vision inputs (single-view images, multi-view images, and video) are processed
        by a shared Vision Encoder. At the same time, language instructions are tokenized. Both streams are fused by the LLM Decoder to produce a shared embodied state. 
        For low-latency closed-loop control, a lightweight Fast Vision Encoder~(DINOv3) encodes
        the most recent multi-view observations into real-time features injected directly into a flow-matching Action Expert, which integrates a noise-initialized action conditioned on the decoder's high-level context into executable continuous and discrete action chunks. Within a single mixture-of-transformer model, \myname{} decodes the shared embodied state into four multimodal output types: bounding boxes/masks~(Spatial Perception),
        language subgoals (Decision Making), navigation actions and manipulation chunks~(Embodied Interaction), and progress estimates~(Self Monitoring), forming the
        Perception-Planning-Action-Evaluation closed loop.}
    \label{fig:arch}
\end{figure}

\myname{} is built upon ACE-Brain-0~\citep{acebrain0}, an embodied foundation model initialized from Qwen3-VL~\citep{bai2025qwen3vltechnicalreport} for physical agentic AI. ACE-Brain-0 provides a strong spatial-intelligence scaffold through large-scale pretraining across spatial cognition, autonomous driving, low-altitude sensing, and embodied scene understanding. \myname{} extends this understanding-centric backbone toward the Unified Embodied Foundation Model paradigm, directly instantiating four cognitive functions within a single model: \textit{Spatial Perception}, \textit{Decision Making}, \textit{Embodied Interaction}, and \textit{Self Monitoring}. These four functions are realized as a closed-loop Perception--Planning--Action--Evaluation process: the model perceives the scene and grounds objects, plans executable subgoals, acts through navigation or manipulation, and evaluates its own execution progress. The fifth function, \textit{Self Improvement}, is enabled through the self-improving framework described in Section~\ref{sec:self_improve}.

At its core, \myname{} adopts an all-in-one mixture-of-transformer design.
A shared embodied backbone encodes heterogeneous inputs and maintains a unified
scene-and-task representation, while task-specific interfaces decode this representation
into multimodal outputs spanning language, structured spatial predictions, and continuous
robot actions.
Fig.~\ref{fig:arch} illustrates the overall architecture.

\noindent\textbf{Omni-Vision Encoder.} The Vision Encoder is shared across all visual modalities, encoding single-view images,
multi-view observations, and videos through a unified pathway. Single-view images support efficient object-level perception. Multi-view observations provide cross-view geometric constraints for spatial relation reasoning and 3D grounding. Videos preserve temporal context for progress estimation and sequential decision making.
The encoded visual tokens are projected into the LLM token space via a shared projection layer.

\noindent\textbf{LLM Decoder.} Language instructions are tokenized and fused with the visual tokens in the LLM Decoder, which serves as the unified backbone for all downstream capabilities. Given a language instruction~$\ell$, visual observations~$o_t$, and optional robot
proprioceptive state~$q_t$ at timestep~$t$, the LLM Decoder $F_\theta$ produces a shared embodied state:
\begin{equation}
s_t = F_\theta(\ell,\, o_t,\, q_t).
\label{eq:shared_state}
\end{equation}
This shared state serves as the common context for all task-specific decoding heads and as the high-level semantic input to the Action Expert.

\noindent\textbf{Action Expert and Fast Vision Pathway.} To translate the shared embodied state into executable robot control, \myname{} introduces a dedicated {Action Expert} implemented as a flow-matching~\citep{lipman2023flow} policy head, following the action-expert design of $\pi_0$~\citep{black2024pi_0}. Conditioned jointly on the LLM Decoder output $s_t$ and real-time visual features $z_t$, the Action Expert integrates a noise-initialized action along the learned flow into an executable action chunk:
\begin{equation}
a_t = \mathrm{ActionExpert}(s_t,\, z_t).
\end{equation}
To meet the strict latency requirements of real-time control, the Action Expert is paired
with a lightweight {Fast Vision Pathway}.
Specifically, a DINOv3~\citep{simeoni2025dinov3} encoder processes the most recent
multi-view observations and produces real-time perceptual features:
\begin{equation}
z_t = E_{\mathrm{fast}}(o_t),
\end{equation}
which are injected directly into the Action Expert, bypassing the heavier LLM Decoder at control frequency. This yields a \textit{two-timescale control interface}: the LLM Decoder computes high-level embodied context at a lower frequency, while the Fast Vision Pathway supplies up-to-date perceptual features at the control frequency. The Action Expert combines the cached decoder context with the latest Fast Vision features to generate reactive action chunks, avoiding the need to route every low-level control update through the full LLM Decoder while keeping action generation grounded in the shared backbone representation. During training, the VLM backbone is frozen and only the Fast Vision Pathway and the Action Expert are updated. This decoupling enables efficient manipulation learning without perturbing the backbone's spatial reasoning and language grounding capabilities.

\noindent\textbf{Task-Specific Decoding.}
Within this single mixture-of-transformer architecture, each cognitive function is realized
through a dedicated decoding path conditioned on the shared embodied state $s_t$:

\noindent\textit{Spatial Perception.}
The LLM Decoder autoregressively predicts bounding boxes, segmentation masks, and
spatial point coordinates, grounding target objects, regions, and affordances directly
in the visual observation.

\noindent\textit{Decision Making.}
Task planning is formulated as autoregressive language generation. The model reasons over the current scene state and decomposes high-level instructions into a sequence of executable subgoals expressed in natural language.

\noindent\textit{Embodied Interaction.}
For navigation, the model predicts discrete actions conditioned on the instruction, observation history, and decoded scene state. For manipulation, the Action Expert produces continuous action chunks in the robot's end-effector or joint space, conditioned jointly on $s_t$ and $z_t$ as defined above.

\noindent\textit{Self Monitoring.}
Task progress estimation is formulated as frame-wise sequence prediction. Given a trajectory of visual observations and the task instruction, the model outputs a normalized progress sequence $\hat{p}_t \in [0, 1]$ and a pairwise trajectory preference judgment, enabling execution monitoring and failure recovery within the same unified framework.

Together, these decoding paths instantiate the Perception--Planning--Action--Evaluation closed loop. Token-based perception, reasoning, and planning share a unified token space, while real-time control is routed through the Action Expert with low latency. \myname{} thus realizes a unified embodied foundation model through shared context and coordinated decoding, while retaining output-specific interfaces for the heterogeneous signals required in physical interaction.


\subsection{Training Strategy: Scaffold, Specialize, Reconcile, and Reactivate (SSR+)}
\label{sec:ssr}

Extending ACE-Brain-0 from an understanding-centric model to a unified embodied foundation model
introduces a fundamental optimization challenge: the heterogeneity of supervision signals. Specifically,
1)~Spatial QA and task planning are trained through textual generation; 2)~Grounding requires structured region or point outputs; 3)~Navigation requires sequential action prediction under egocentric observations; 4)~Manipulation requires continuous action chunks; 5)~Progress estimation requires temporally grounded evaluation of execution states.
Directly mixing all data sources in a single supervised fine-tuning~(SFT) stage leads to cross-interface interference: the model may retain the semantic knowledge for a task but fail
to follow the correct output convention for that interface, or specialize one interface at
the cost of others. Conversely, training capabilities in complete isolation yields strong task-specific checkpoints but does not produce a single unified model.

To resolve this tension, we extend the Scaffold--Specialize--Reconcile (SSR) strategy
of ACE-Brain-0~\citep{acebrain0} with a fourth \textit{Reactivate} stage,
yielding the \textbf{SSR+} training paradigm.
The four stages are designed to:
(1) provide a spatially capable initialization,
(2) develop individual task-specific capabilities independently,
(3) consolidate them into a single parameter space through task vector merging, and
(4) calibrate output conventions across all interfaces with lightweight fine-tuning.

\noindent\textbf{Stage~1: Scaffold.} Let $\theta$ denote the Qwen3-VL-8B-Instruct~\citep{bai2025qwen3vltechnicalreport} initialization.
The scaffold is provided by the ACE-Brain-0 checkpoint $\theta_0$, which supplies the broad spatial and embodied understanding representations inherited from the first-generation
model. $\theta_0$ captures transferable spatial representations that are shared across grounding,
navigation, manipulation, and progress estimation, and it serves as the starting point for
all subsequent specialization.

\textbf{Stage~2: Specialize.}
Starting from $\theta$, we independently train four task-specialized checkpoints,
each optimized on its corresponding data distribution:
1)~$\theta_{\mathrm{qa}}$: spatial question answering and chain-of-thought task planning;
2)~$\theta_{\mathrm{grd}}$: 2D and 3D object grounding with structured coordinate outputs;
3)~$\theta_{\mathrm{nav}}$: egocentric language-guided navigation action prediction;
4)~$\theta_{\mathrm{prog}}$: frame-wise progress estimation and pairwise trajectory preference.
Training each capability in isolation prevents supervision from one interface from degrading the output convention of another.

\textbf{Stage~3: Reconcile.} 
Naive data mixing across heterogeneous interfaces often causes cross-task interference, where optimizing one capability degrades another.
We therefore reconcile the specialized checkpoints through optimization-based task vector merging~\cite{shen2025efficient}.
Each task vector $\tau_i = \theta_i - \theta$ captures the weight-space displacement induced by specialization. 
Following the merging practice of ACE-Brain-0~\citep{acebrain0} and recent multi-task fusion methods~\citep{tangFusionBenchUnifiedLibrary2025}, we adopt a layer-wise objective that minimizes task interference across all experts.
For layer $l$, the objective is:
\begin{equation}
\theta^*_{\mathrm{merge},l}
= \arg\min_{\theta_{\mathrm{merge},l}}
\sum_{i=1}^{K}
\mathbb{E}_{x_{i,l}\sim\mathcal{D}_{m_i,l}}
\bigl\|\theta_{i,l}\,x_{i,l} - \theta_{\mathrm{merge},l}\,x_{i,l}\bigr\|_2^2,
\label{eq:merge_obj}
\end{equation}
where $\mathcal{D}_{m_i,l}$ is the approximated data distribution of the $i$-th expert at layer $l$.
Decomposing into task vectors and upper-bounding the residual interference~\citep{acebrain0,shen2025efficient}, Eq.~\eqref{eq:merge_obj} is approximately solved as:
\begin{equation}
\theta^*_{\mathrm{merge},l}
\approx
\theta_{\mathrm{pre},l}
+
\arg\min_{\tau_{\mathrm{merge},l}}
\sum_{i=1}^{K}
\frac{1}{\|\tau_{i,l}\|_F^2}
\bigl\|(\tau_{\mathrm{merge},l} - \tau_{i,l})\,\tau_i^\top\bigr\|_F^2.
\label{eq:merge_approx}
\end{equation}
The optimization runs for 1{,}000 data-free iterations with the Adam optimizer via the FusionBench framework~\citep{tangFusionBenchUnifiedLibrary2025}.
The overall reconciliation step is therefore:
\begin{equation}
\theta_{\mathrm{merge}}=
\operatorname{Merge}
\!\left(
\theta;\;
\theta_0,\,
\theta_{\mathrm{qa}},\,
\theta_{\mathrm{grd}},\,
\theta_{\mathrm{nav}},\,
\theta_{\mathrm{prog}}
\right).
\label{eq:merge}
\end{equation}
This allows $\theta_{\mathrm{merge}}$ to inherit the broad spatial scaffold of $\theta_0$ while incorporating the decoding interfaces for grounding, navigation, and progress estimation, all within a single unified parameter space.

\textbf{Stage~4: Reactivate.}
The merged checkpoint $\theta_{\mathrm{merge}}$ is not used directly as the final model. 
A central empirical finding of SSR+ is that $\theta_{\mathrm{merge}}$ can \textit{rapidly recover} strong performance on each task with only a small number of SFT steps---far fewer than the Specialize stage requires from $\theta$.
This fast-recovery phenomenon reveals a structural property of the merging objective: because Eq.\eqref{eq:merge_obj} minimizes the per-layer output discrepancy between $\theta_{\mathrm{merge}}$ and each expert $\theta_i$ on their respective input distributions, the merged model already produces near-identical intermediate representations to each specialist for typical inputs.
In other words, the task-specific semantic knowledge---the ability to localize objects, predict navigation actions, or estimate progress---is encoded in the merged weights and survives the reconciliation step.
What is lost is not capability but \textit{interface convention}: different experts have learned to express results in different output formats (e.g., structured coordinates for grounding vs. discrete action tokens for navigation vs. scalar progress values), and weight-space averaging can temporarily desynchronize these format conventions even while preserving the underlying representations.
This finding aligns with post-merge fine-tuning observations in the model merging literature, where a brief warm-up on mixed data reliably restores or even improves upon individual task performance~\cite{komatsuzaki2022sparse}.

We therefore apply a lightweight Reactivate stage: $\theta_{\mathrm{merge}}$ is fine-tuned on a compact mixed SFT dataset $\mathcal{D}_{\mathrm{mix}}$ for a small number of updates,
\begin{equation}
\theta_{\mathrm{0.5}}
=
\operatorname{SFT}
\!\left(
\theta_{\mathrm{merge}},\,
\mathcal{D}_{\mathrm{mix}}
\right).
\label{eq:reactivate}
\end{equation}
$\mathcal{D}_{\mathrm{mix}}$ contains a small representative sample from each task interface, providing the output-format supervision needed to re-synchronize the conventions across all interfaces.
Because the underlying representations are already aligned, this stage requires orders of magnitude fewer steps than a cold-start Specialize run: the gradient signal targets only surface-level format calibration rather than deep representational learning.
Unlike the Specialize stage, Reactivation does not re-learn each capability from scratch. 
It serves as a post-merge calibration step that unlocks the task knowledge already encoded in the merged weights, resolves output-format conflicts, and restores a consistent interface switching across heterogeneous embodied tasks.
As a result, $\theta_{\mathrm{0.5}}$ preserves the spatial and embodied understanding capabilities of ACE-Brain-0~\citep{acebrain0} while extending them to grounding, navigation, manipulation, and execution monitoring within a single unified model.

\subsection{Self-Improving Framework}
\label{sec:self_improve}

\myname{} is designed to be embedded into a closed-loop self-improving framework for real-world embodied deployment. Rather than treating a physical task as a fixed, hand-engineered pipeline, we represent execution as a Perception--Planning--Action--Evaluation loop in which the agent repeatedly perceives the scene, plans the next executable subgoal, performs navigation or manipulation, and evaluates its own execution progress. This loop is task-agnostic: a concrete task such as cloth washing is one instantiation of this general framework.

Given a language instruction $\ell$, visual observations $o_t$, and optional proprioceptive state $q_t$, \myname{} computes the shared embodied state $s_t$
(Eq.~\ref{eq:shared_state}) and decodes four coordinated outputs:
\begin{equation}
(g_t,\; p_t,\; a_t,\; e_t) = D_\theta(s_t),
\end{equation}
where $g_t$ denotes the grounded scene state~(Spatial Perception), $p_t$ denotes the executable subgoal plan~(Decision Making), $a_t$ denotes the navigation or manipulation action~(Embodied Interaction),
and $e_t$ denotes the execution evaluation signal~(Self Monitoring). These four outputs directly correspond to the four implemented cognitive functions of \myname{},
instantiating the unified perception--planning--action--evaluation loop within a single model.

The self-improving component operates over an external execution state $\mathcal{H}$, which encodes accumulated task-level knowledge including task schemas, spatial memory, failure recovery cases, tool-use constraints, and reusable skill descriptions. 
After each rollout, the agent stores an experience tuple:
\begin{equation}
\xi = (\ell,\; \tau,\; p_1,\; a_1,\; e_1),
\end{equation}
where $\tau$ is the observed trajectory.
Successful rollouts, corrected failures, and progress-labeled trajectories are selected
as feedback for updating the external execution state:
\begin{equation}
\mathcal{H}_{k+1} = \mathcal{U}\!\left(\mathcal{H}_k,\; \{\xi_i\}_{i=1}^{N}\right).
\label{eq:self_improve}
\end{equation}

This formulation keeps self-improvement practical for physical deployment. First, the agent improves through lightweight updates to the external execution
state $\mathcal{H}$, which requires no model retraining and can be deployed incrementally
after each rollout. Then, model-level adaptation can be performed only when sufficient validated experience has been accumulated.
In this way, \myname{} provides the unified perception, planning, action, and self-monitoring
interface required to support the Self Improvement cognitive function and to serve as a
foundation for future self-improving embodied agents.

%% file: contexts/4_exp.tex
\section{Experiments}


To comprehensively evaluate ACE-Brain-0.5, we conduct experiments across the core cognitive functions required by a unified embodied brain, including spatial perception, decision making, embodied interaction, and self-monitoring. Specifically, \ourMethod{} is assessed along four capability axes. First, we evaluate \textbf{embodied spatial perception and planning}, covering spatial QA, 3D scene understanding, language-guided 3D grounding, embodied-centric spatial reasoning, affordance localization, pointing, and trajectory prediction on benchmarks such as VSI~\cite{vsi}, MMSI~\cite{mmsi}, MindCube~\cite{mindcube}, ScanQA~\cite{scanqa}, SQA3D~\cite{sqa3d}, Scan2Cap~\cite{scan2cap}, ScanRefer~\cite{scanrefer}, Multi3DRef~\cite{multi3drefer}, EmbSpatial~\cite{embspatial}, ERQA~\cite{gemini_robotics}, RoboSpatial~\cite{robospatial}, RefSpatial~\cite{zhou2025roborefer}, PointArena~\cite{pointarena}, RoboAfford~\cite{roboafford}, and ShareRobot-Traj~\cite{robobrain}. Second, we evaluate \textbf{language-guided navigation} on VLN-CE, testing instruction grounding, egocentric observation integration, and sequential decision making under partial observability. Third, we evaluate \textbf{robotic manipulation} on SimplerEnv-Bridge~\citep{li2024simplerenv} and LIBERO~\citep{liu2023libero}, measuring whether the model can generate executable actions for object-centric manipulation. Finally, we evaluate \textbf{progress evaluation and reward modeling}, focusing on task-progress estimation and execution-state assessment in closed-loop embodied settings. All tasks are handled by a single unified architecture without task-specific architectural modifications, demonstrating the generalization of \ourMethod{} across the perception--planning--action--evaluation pipeline of embodied intelligence.

\subsection{Spatial Perception}
\label{sec}

Spatial intelligence is the foundation for physical intelligence: an embodied brain must perceive the spatial state of the physical world and use it to support planning-oriented decisions. We therefore evaluate ACE-Brain-0.5 from two complementary perspectives: embodied spatial perception, which measures object/region grounding, 3D scene understanding, egocentric spatial relations, multi-view memory, and affordance perception; and planning-oriented spatial reasoning, which measures whether spatial representations can be converted into action-relevant targets, interaction regions, and trajectories.

As shown in Table~\ref{perception_planning}, ACE-Brain-0.5 achieves strong performance on physical-world spatial perception benchmarks. On VSI, which evaluates egocentric video spatial reasoning, ACE-Brain-0.5 obtains 62.2\%, slightly underperforming ACE-Brain-0 by 0.9\%, while outperforming GPT-5.4 52.6\% and Gemini-2.5-Pro 43.4\%. On MMSI, which focuses on multi-image spatial reasoning, ACE-Brain-0.5 reaches 35.5\%, improving over ACE-Brain-0 32.2\%. On MindCube, which tests spatial mental modeling under limited observations, ACE-Brain-0.5 achieves 86.3\%, improving over ACE-Brain-0 82.1\% and substantially surpassing RynnBrain-8B 56.6\%. These results suggest that ACE-Brain-0.5 preserves the spatial scaffold of the previous generation while strengthening spatial memory and mental modeling.

For 3D scene perception and grounding, ACE-Brain-0.5 also shows clear gains. It reaches 99.2 on ScanQA and 62.6\% on SQA3D, improving over ACE-Brain-0 on both benchmarks. On language-grounded 3D perception, ACE-Brain-0.5 achieves 83.3 on Scan2Cap, 70.2\% on ScanRefer, and 72.4\% on Multi3DRef, improving over ACE-Brain-0 by 8.1\%, 8.8\%, and 16.5\%, respectively. These benchmarks evaluate dense 3D captioning, object localization, and multi-object reference resolution, indicating that ACE-Brain-0.5 can align language with physical entities and regions in 3D space. On broader embodied reasoning benchmarks, ACE-Brain-0.5 obtains 75.9\% on EmbSpatial and 46.3\% on ERQA, showing competitive performance across diverse spatial reasoning settings.

We further evaluate whether ACE-Brain-0.5 can use spatial perception for planning-oriented embodied reasoning. On RoboSpatial, which measures robot-centric spatial relations, reference frames, and spatial compatibility, ACE-Brain-0.5 achieves 60.1\%, improving over ACE-Brain-0 55.6\% and outperforming closed-source MLLMs including GPT-5.4 53.5\%, Gemini-2.5-Pro 53.7\%, and Claude-Sonnet-4.6 37.3\%. On RefSpatial, which requires multi-step spatial reference resolution, ACE-Brain-0.5 reaches 55.6\%, largely improving over ACE-Brain-0 26.0\% and surpassing GPT-5.4 15.7\%, Gemini-2.5-Pro 36.5\%, and Cosmos3-Nano 53.1\%. On PointArena, which evaluates language-guided pointing to physical targets, ACE-Brain-0.5 obtains 68.5\%, outperforming ACE-Brain-0 44.7\% and Gemini-2.5-Pro 62.8\%, while remaining comparable to Cosmos3-Nano 69.4\% and Embodied-R1.5 71.4\%.

For affordance and trajectory-level planning, ACE-Brain-0.5 further demonstrates action-relevant spatial reasoning. On RoboAfford, it achieves 75.1\%, improving over ACE-Brain-0 56.5\% by 18.6\% and surpassing RoboBrain-2.5-8B 74.9\%, while remaining comparable to Embodied-R1.5 80.0\%. On ShareRobot-Traj, where lower error is better, ACE-Brain-0.5 reduces the trajectory prediction error to 0.32, improving over ACE-Brain-0 0.46, matching GPT-5.4, and outperforming Gemini-2.5-Pro 0.34 and Claude-Sonnet-4.6 0.39.

Overall, ACE-Brain-0.5 advances spatial intelligence from physical-world perception to planning-oriented spatial reasoning. The gains on 3D grounding, reference resolution, target pointing, affordance understanding, and trajectory prediction indicate that the model not only recognizes spatial structures, but can also convert them into action-relevant representations for downstream navigation, manipulation, and closed-loop embodied interaction.

\input{tables/spatial_grounding}

\subsection{Decision Making}

Autonomous driving provides a decision-centric evaluation setting for embodied brain models~\cite{rawdrive,llm4drive,drivemoe,drivegpt4,drivelm}. Different from static spatial QA, driving requires the model to understand road topology, traffic participants, ego-vehicle state, map constraints, and temporal scene dynamics, and then convert such information into safe and feasible high-level decisions. We therefore evaluate ACE-Brain-0.5 on autonomous driving benchmarks covering real-world scene understanding, action-level decision prediction, and driving QA.

As shown in Table~\ref{perception_planning}, ACE-Brain-0.5 maintains competitive driving-scene understanding. On MME-RealWorld, it obtains 66.6\%, outperforming most embodied baselines and remaining comparable to stronger general MLLMs. On MAPLM, which emphasizes map-aware and traffic-structured reasoning, ACE-Brain-0.5 achieves 71.3\%, substantially surpassing GPT-5.4 57.9\%, Gemini-2.5-Pro 26.1\%, Claude-Sonnet-4.6 56.7\%, Cosmos3-Nano 26.7\%, and RoboBrain-2.0-7B 31.7\%. This indicates that the model can associate visual observations with structured driving context rather than only recognizing isolated traffic objects.

For action-level decision making, ACE-Brain-0.5 reaches 78.2\% on DriveAction, showing comparable performance to several driving or embodied models such as Embodied-R1.5 77.1\%, RynnBrain-8B 76.3\%, Vlaser-8B 78.1\%, Pelican-VL-7B 77.2\%, and VeBrain-7B 78.3\%. On driving QA benchmarks, ACE-Brain-0.5 obtains 44.7\% on NuScenesQA and 86.7\% on NuPlanQA, demonstrating strong multi-agent and multi-view scene reasoning. On LingoQA, it achieves 55.6\%, remaining comparable to Pelican-VL-7B 56.0\% and VeBrain-7B 55.0\%, although long-context driving video reasoning remains challenging.

Overall, ACE-Brain-0.5 does not aim to be a driving-specialized model, and it shows moderate drops from ACE-Brain-0 on several driving-specific benchmarks. Nevertheless, its performance indicates that the unified embodied brain preserves strong decision-making ability across real-world perception, map-aware reasoning, action prediction, and driving QA, while being optimized for broader spatial grounding and planning-oriented embodied interaction.

\subsection{Embodied Interaction}

\paragraph{Navigation}
As shown in Table~\ref{tab:vlnce_open_source}, we evaluate ACE-Brain-0.5 on embodied navigation in the VLN-CE setting~\cite{vlnce}, covering instruction grounding, egocentric scene understanding, temporal observation integration, action prediction, and navigation decision-making under partial observability.
\input{tables/navigation}

On R2R Val-Unseen, the unified ACE-Brain-0.5 model achieves 57.4\% SR, 63.7\% OS, and 4.8 NE. It substantially outperforms NaVid and Uni-NaVid, and remains competitive with recent open-source navigation models such as StreamVLN, NavFoM, RynnBrain-Nav, and Qwen-VLA-Instruct. In particular, ACE-Brain-0.5 obtains lower NE than all compared general baselines, showing that the unified embodied model can accurately localize target goals in unseen environments. The specialist variant further improves the main navigation metrics to 62.2\% SR and 4.2 NE, achieving the best SR and NE among all compared methods, while also maintaining a strong OS of 67.4\%. These results show that ACE-Brain-0.5 already provides a strong unified navigation capability, and navigation-specific specialization further strengthens goal-reaching accuracy.

On RxR Val-Unseen, ACE-Brain-0.5 shows even stronger advantages. The unified model achieves the best NE of 4.3 and the highest SR of 63.8\%, outperforming all open-source baselines including NavFoM, RynnBrain-Nav, and Qwen-VLA-Instruct. This demonstrates that ACE-Brain-0.5 generalizes well to longer and more detailed navigation instructions, where robust instruction grounding and temporal observation integration are essential. The specialist variant maintains a comparable SR of 63.6\% and obtains the best nDTW of 67.1, indicating stronger trajectory alignment with the reference path.

Overall, ACE-Brain-0.5 achieves strong navigation performance as a unified embodied model, reaching competitive results among open-source VLN-CE baselines. ACE-Brain-0.5-Specialist further improves the key R2R navigation metrics and strengthens path consistency on RxR. These results indicate that the spatial perception and planning representations learned by ACE-Brain-0.5 can be effectively transferred to instruction-grounded sequential decision making, while task-specialized tuning can further enhance navigation-specific goal reaching.


\paragraph{Manipulation}
\label{sec:embodied_manipulation}

Embodied manipulation evaluates whether the spatial and task-level representations learned by ACE-Brain-0.5 can be converted into executable robot actions. We validate ACE-Brain-0.5 from two complementary perspectives.

First, to support real-time low-level control and maintain the perception--planning ability, we build directly on the ACE-Brain-0.5 architecture, freeze the backbone, and train only the FastVision module and a lightweight flow-matching action expert, then evaluate on LIBERO. This design preserves the perception--planning representation of the ACE-Brain-0.5 family while making closed-loop manipulation inference more efficient. LIBERO measures language-conditioned manipulation across spatial, object, goal, and long-horizon task suites, making it well-suited for assessing the data-fitting capacity of a policy built on ACE-Brain-0.5. As shown in Table~\ref{tab:vla_libero}, ACE-Brain-0.5 achieves an average success rate of 98.2\%, outperforming Qwen-VLA-Instruct 97.9\%, OpenVLA-OFT 97.1\%, GR00T N1.6 97.0\%, $\pi_{0.5}$
 96.9\%, and other strong VLA baselines. It achieves 100.0\% on both the Spatial and Object suites, showing strong object-centric perception and spatial grounding for manipulation. On the Long suite, ACE-Brain-0.5 obtains 97.0\%, surpassing VITA 96.8\%, OpenVLA-OFT 94.5\%, and GR00T N1.6 94.4\%, indicating that the learned representation also supports multi-step manipulation execution. On the Goal suite, ACE-Brain-0.5 reaches 96.0\%, remaining competitive with the strongest baselines.
 
Second, we follow the action-expert design of the $\pi$ series and instantiate a compact VLA variant, ACE-Brain-0.5-VLA. This variant does not use FastVision; instead, it fully fine-tunes the ACE-Brain-0.5 VLM and attaches a lightweight flow-matching action head that progressively denoises and generates action chunks conditioned on visual observations and language instructions. We evaluate ACE-Brain-0.5-VLA on SimplerEnv-Bridge, which tests object manipulation under a fixed maximum inference horizon. Here, given that there is sufficient training data on the Bridge dataset~\citep{walke2023bridgedata}, we did not use the pre-training weights of ACE-Brain-0.5 on manipulation. As shown in Table~\ref{tab:simplerenv}, ACE-Brain-0.5-VLA achieves the best (SOTA) average success rate of 82.3\%, outperforming GTA-VLA 81.2\%, X-VLA 76.0\%, Qwen-VLA-Instruct 73.7\%, and Uni-VLA 69.8\%. It obtains the best performance on Cube 75.0\% and Eggplant 100.0\%, while remaining competitive on Carrot 79.2\%. Although it is lower than GTA-VLA and X-VLA on Spoon, the overall average demonstrates strong cross-object manipulation ability.
Together, these two settings show that ACE-Brain-0.5 can translate embodied spatial perception and task-level reasoning into executable manipulation actions, providing the embodied interaction component of the unified robot brain.

\input{tables/VLA_LIBERO}

\subsection{Self Monitoring}
As shown in Table~\ref{tab:reward_rbm_1m}, we evaluate the progress estimation ability of \myname through robotic progress estimation. Given a task instruction and a trajectory, the model is required to predict temporally consistent progress scores that reflect whether the execution is moving toward task completion. We report VOC, which measures the correlation between predicted progress values and ground-truth temporal progress.

We evaluate on RBM-EVAL-ID and RBM-EVAL-OOD from Robometer~\citep{robometer2026}, and further introduce refined variants of both splits. The standard RBM-EVAL setting mainly evaluates whether models can recover the upward progress trend in normal trajectories. However, this may allow a shortcut: a model can obtain a reasonable score by recognizing later-stage visual states or simply regressing to a monotonic increasing pattern, without truly understanding whether the task is being executed in the correct temporal direction. To test this more directly, our refined splits add reversed successful trajectories as negative controls, where the visual content remains unchanged but the temporal order is inverted. This setting requires the model to distinguish moving toward the goal from undoing the task.

\myname achieves the best performance across all four test suites, reaching 0.94 and 0.96 VOC on the standard ID and OOD splits, and 0.80 and 0.88 on the refined splits. The refined setting causes clear performance drops for most baselines, especially general VLMs and reward models that are sensitive to static success cues. In contrast, \myname remains consistently strong, outperforming Robometer by 0.02 on RBM-EVAL-ID-Refined and 0.07 on RBM-EVAL-OOD-Refined. These results suggest that \myname does not merely fit an increasing progress prior, but better captures temporal task evolution and action-direction under both in-distribution and out-of-distribution settings.
\input{tables/Simplerenv_Bridge}

\input{tables/reward_rbm_1m}

\begin{figure}[!t]
    \centering
    \includegraphics[width=1.\linewidth]{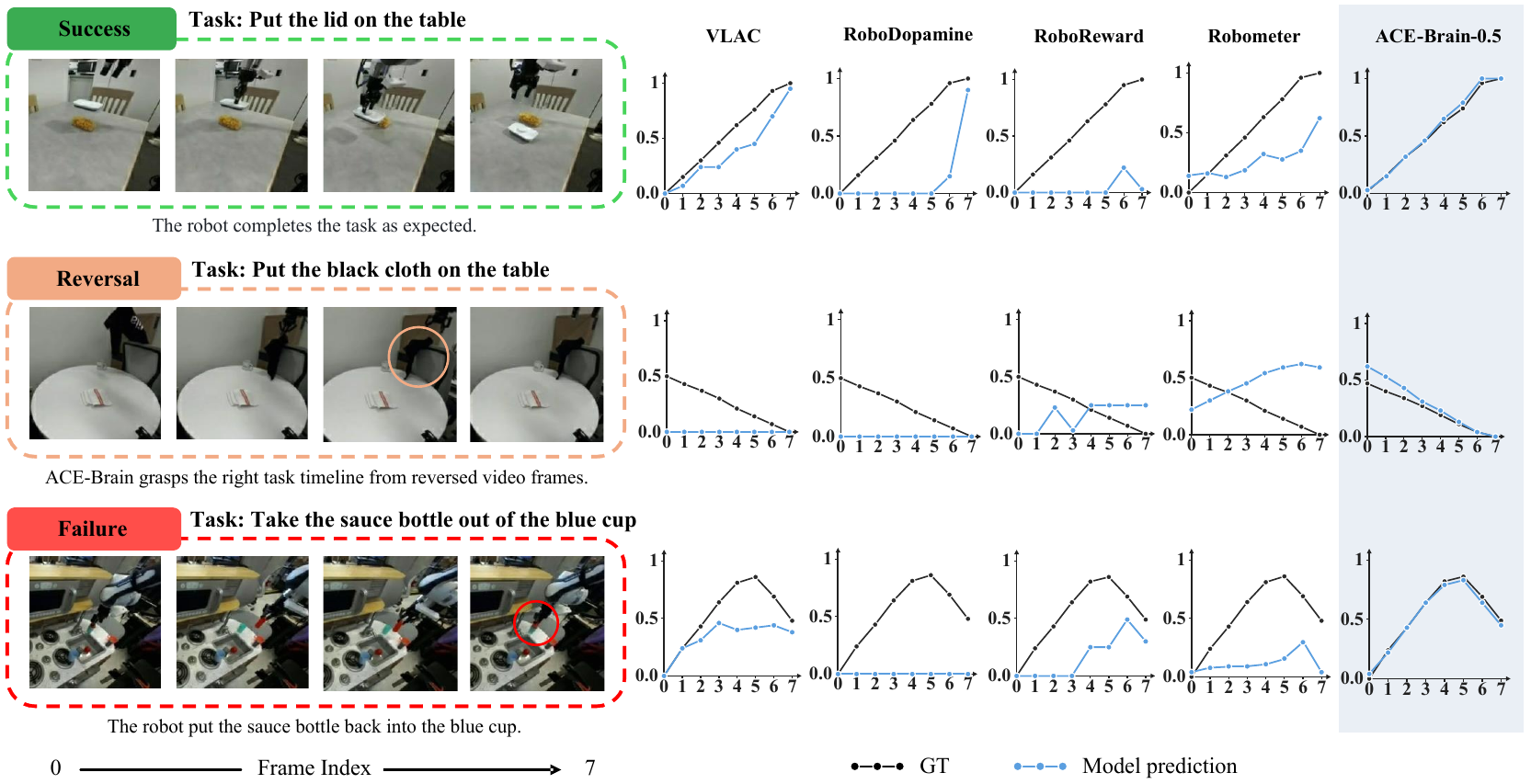}
    \caption{
    Qualitative visualization of task progress prediction by ACE-Brain-0.5.
    Each row shows a typical trajectory setting.
    The model predicts temporally consistent progress scores from visual observations.
    For success, reversal, and failure trajectory samples, the progress curves predicted by ACE-Brain-0.5 closely match the true task evolution.
    The model captures the overall task completion trend and also detects small changes in intermediate states.
    These results show that ACE-Brain-0.5 has strong temporal understanding.
    They also support its potential as a reward model for robotic task execution.
    }
    \label{fig:progress_vis}
\end{figure}

As shown in Figure~\ref{fig:progress_vis}, 
we further provide qualitative visualizations of the predicted progress curves produced by ACE-Brain-0.5. The figure contains three rows, each corresponding to a different trajectory case, and illustrates how the model assigns progress scores along the temporal execution process. ACE-Brain-0.5 consistently produces curves that align well with the actual task evolution, not only capturing the overall completion trend in complex scenes but also reflecting fine-grained changes between intermediate visual states. Such behavior suggests that ACE-Brain-0.5 has strong potential to serve as a reward model for robotic learning and evaluation.

\subsection{Self Improvement}
To move beyond static trajectory imitation, we introduce self improvement that converts the model's own closed-loop rollouts into corrective navigation supervision. Static demonstrations mainly cover near-optimal states, while a learned policy inevitably visits off-policy states during execution, especially near ambiguous intersections, visually similar corridors, and instruction-progress transitions. We therefore let the current policy interact with the environment and use its failures to expose states that require correction.

Starting from a cold-start policy trained on navigation demonstrations, the model performs closed-loop rollouts under its own predictions. An oracle navigation teacher is used to detect deviations based on navigation progress and path consistency, rather than treating every action mismatch as an error. A deviation is triggered when the predicted action increases the distance to the goal, reduces path alignment, causes premature stopping, or accumulates local error. Once a deviation is detected, the oracle takes over and completes the remaining trajectory, producing an oracle-recovered rollout from the model-induced state. This stage constructs corrective navigation experiences:
\begin{equation}
  \mathcal{D}_{\mathrm{evo}}
  =
  \left\{
  (q, h_t, o_t, \hat{a}_t, a_t^{*}, \rho_t)
  \right\}.
\end{equation}
where $\hat{a}_t$ is the model-predicted action, $a_t^{*}$ is the oracle recovery action, and $\rho_t$ records offline deviation metadata, such as progress state, distance-to-goal change, path-alignment status, and accumulated local error. This metadata is used only for diagnosis, filtering, and supervision construction, and is not provided to the policy during evaluation.

The oracle-recovered trajectories are filtered by goal completion and path-efficiency criteria, and then merged with the original demonstrations to train the self-improved policy. In this data flywheel, the policy discovers its own failure modes, the oracle converts them into executable correction trajectories, and the model learns from an expanded closed-loop state distribution. Compared with static imitation, this improves robustness to off-policy states and reduces compounding errors in long-horizon navigation.

\input{tables/nav_improve}
\begin{figure}[H]
    \centering
    \includegraphics[width=0.95\linewidth]{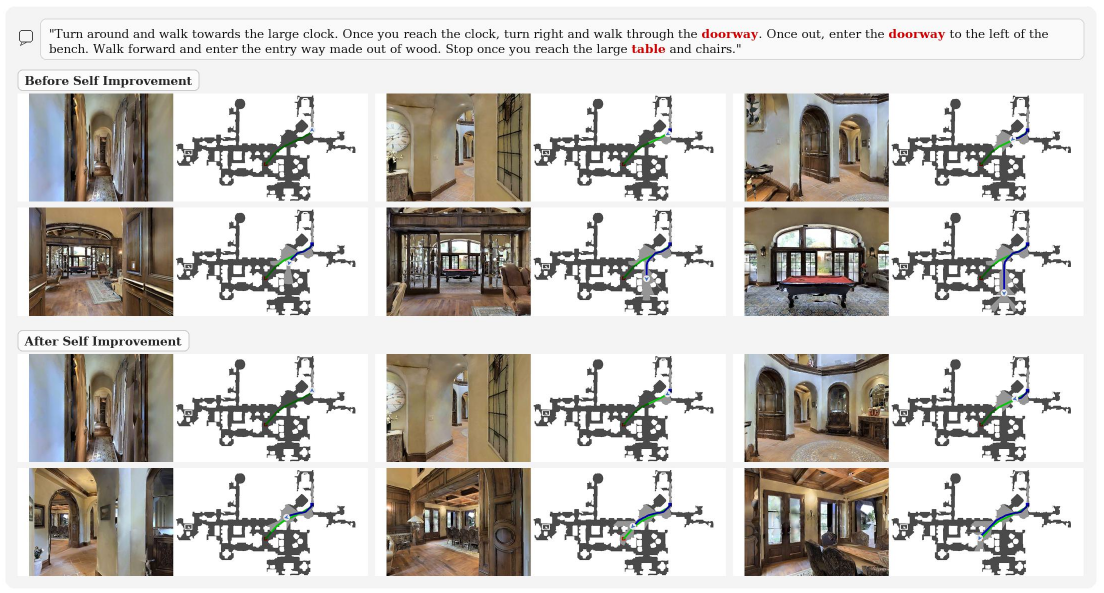}
    \caption{
    \textbf{Visualization of navigation self improvement.}
    The self-improved model corrects an intermediate navigation error and follows the instruction to the target.
    }
    \label{fig:self_improve}
\end{figure}

%% file: tables/spatial_grounding.tex

\definecolor{ourscol}{HTML}{EAF3FF}
\definecolor{rankonecol}{HTML}{F3B46D}
\definecolor{ranktwocol}{HTML}{F7D29A}
\definecolor{rankthreecol}{HTML}{FBE8C8}

\providecommand{\na}{--}
\providecommand{\mhead}[1]{\rotatebox[origin=c]{90}{\makecell[c]{\bfseries #1}}}
\providecommand{\benchgroup}[1]{\rotatebox[origin=c]{90}{\makecell[c]{\bfseries #1}}}
\providecommand{\ourcell}[1]{\cellcolor{ourscol}#1}
\providecommand{\rankone}[1]{\cellcolor{rankonecol}#1}
\providecommand{\ranktwo}[1]{\cellcolor{ranktwocol}#1}
\providecommand{\rankthree}[1]{\cellcolor{rankthreecol}#1}

\begin{table*}[!t]
\centering
\scriptsize
\setlength{\tabcolsep}{2.2pt}
\setlength{\extrarowheight}{1pt}
\renewcommand{\arraystretch}{1.1}
\caption{Benchmark comparison across embodied spatial question answering, grounding, and driving benchmarks. Highlighted column denotes ACE-Brain-0.5. Orange-shaded cells denote the best, second-best, and third-best results among Embodied Brain MLLMs for each benchmark, with darker shades indicating higher rank.}
\resizebox{\textwidth}{!}{%
\begin{tabular}{
  @{}c l @{\hskip 3pt}
  *{3}{C}
  !{\vrule width .45pt}
  c *{10}{c}
  @{}
}
\toprule
\multicolumn{2}{c}{
  \raisebox{-3.2em}[0pt][0pt]{
    \makecell[c]{\bfseries Benchmark}
  }
} &
\multicolumn{3}{c}{\textit{Closed-source MLLMs}} &
\multicolumn{11}{c}{\textit{Embodied Brain MLLMs}} \\
\cmidrule(lr){3-5}
\cmidrule(l){6-16}
\multicolumn{2}{c}{} &
\mhead{GPT\\5.4} &
\mhead{Gemini\\2.5 Pro} &
\mhead{Claude\\Sonnet 4.6} &
\ourcell{\mhead{ACE-Brain\\0.5-8B}} &
\mhead{ACE-Brain\\0-8B} &
\mhead{Cosmos3\\Nano} &
\mhead{Embodied\\R1.5} &
\mhead{RynnBrain\\8B} &
\mhead{Vlaser\\8B} &
\mhead{MiMo\\Embodied-7B} &
\mhead{Pelican\\VL-7B} &
\mhead{VeBrain\\7B} &
\mhead{RoboBrain\\2.5-8B} &
\mhead{RoboBrain\\2.0-7B} \\
\midrule

\multirow{13}{*}{\benchgroup{Embodied Spatial QA\\13 benchmarks}}
& VSI
& 52.6 & 43.4 & 17.6
& \rankthree{62.2} & \ranktwo{63.1} & 54.9 & 56.1 & \rankone{71.0} & 60.3 & 48.5 & 52.8 & 39.9 & 41.0$^{*}$ & 46.6 \\

& MMSI
& 31.3 & 38.0 & 30.9
& \rankthree{35.5} & 32.2 & \ranktwo{36.2} & 29.5 & \rankone{39.6} & 27.2 & 31.7$^{*}$ & 26.0$^{*}$ & 27.3$^{*}$ & 29.3$^{*}$ & 30.2 \\

& MindCube
& 45.3 & 57.6 & 43.7
& \rankone{86.3} & \ranktwo{82.1} & 34.8 & 29.4 & \rankthree{56.6} & 34.6 & 32.3$^{*}$ & 31.0 & 30.1$^{*}$ & 28.1$^{*}$ & 31.2$^{*}$ \\

& ScanQA
& 78.3 & 67.0 & 31.5
& \ranktwo{99.2} & \rankthree{97.3} & 60.9 & 67.1 & 60.2 & 55.5 & 83.7 & 65.6 & \rankone{101.5} & 63.1 & 59.7 \\

& SQA3D
& 45.8 & 37.3 & 14.0
& \rankone{62.6} & \rankthree{54.5} & 44.2 & 48.2 & 43.3 & 44.6 & 54.2 & 42.5 & \ranktwo{61.6} & 28.1 & 45.1 \\

& Scan2Cap
& 14.0 & 16.8 & 8.3
& \ranktwo{83.3} & \rankthree{75.2} & 9.2 & 0.3 & 2.8 & 0.2 & 61.8 & 0.5 & \rankone{89.7} & 4.8 & 3.2 \\

& ScanRefer
& 61.7 & 65.9 & 51.5
& \rankone{70.2} & 61.4 & 5.4 & 5.1 & 5.4 & 3.1 & \rankthree{64.7} & 5.4 & \ranktwo{66.4} & 2.3 & 5.4 \\

& Multi3DRef
& 45.1 & 54.4 & 40.2
& \rankone{72.4} & \rankthree{55.9} & 8.1 & 7.7 & 8.1 & 8.2 & 35.9 & 7.9 & \ranktwo{67.8} & 8.1 & 8.1 \\

& SparBench
& 46.1 & 46.2 & 31.2
& \ourcell{39.7} & \rankthree{44.2} & \rankone{54.8} & 40.3 & \ranktwo{49.8} & 41.2 & 11.1 & 38.7 & 32.5 & 42.5 & 42.5 \\

& MMSIVideo
& 32.8 & 34.5 & 29.8
& \ourcell{\rankthree{30.4}} & 26.4 & 27.2 & 27.7 & 27.5 & 27.1 & 27.5 & 29.5 & 28.4 & \rankone{31.3} & \ranktwo{30.5} \\



& EmbSpatial
& 73.2 & 78.7 & 87.9
& \ourcell{75.9} & 77.8 & \rankone{82.9} & \rankthree{78.1} & \ranktwo{80.0} & 75.1 & 76.2 & 73.2 & 30.6 & 75.6 & 76.3 \\

& ERQA
& 50.5 & 55.7 & 41.0
& \ranktwo{46.3} & 41.5 & \rankthree{46.0} & \rankthree{46.0} & \rankone{46.8} & 41.0 & \rankone{46.8} & 39.8 & 33.3 & 22.3 & 39.3 \\

& SAT
& 73.3 & 78.7 & 64.7
&\ranktwo{82.7} & \rankone{92.0} & \rankthree{80.7} & 62.7 & 78.0 & 66.7 & 78.7 & 67.3 & 73.3 & 63.3 & 75.3 \\

\midrule

\multirow{5}{*}{\benchgroup{Grounding\\5 benchmarks}}
& RoboSpatial
& 53.5 & 53.7 & 37.3
& \ourcell{60.1} & 55.6 & 61.0 & \rankthree{69.7} & \rankone{73.1} & 61.7 & 61.8 & 57.5 & 44.9 & \ranktwo{73.0} & 54.2 \\

& RefSpatial
& 15.7 & 36.5 & 4.0
& \rankthree{55.6} & 26.0 & 53.1 & 54.2 & \ranktwo{59.2} & \ranktwo{59.2} & 48.0 & 22.3 & 9.8 & \rankone{60.5} & 32.5 \\

& PointArena
& 37.9 & 62.8 & 14.9
& \rankthree{68.5} & 44.7 & \ranktwo{69.4} & \rankone{71.4} & 65.8$^{*}$ & 60.3 & 3.9 & 24.6 & 12.1 & 59.3$^{*}$ & 15.7 \\

& RoboAfford
& 29.4 & 15.0 & 15.2
& \rankthree{75.1} & 56.5 & \rankone{84.0} & \ranktwo{80.0} & 50.3$^{*}$ & 18.3 & 69.8 & 26.0 & 18.9 & 74.9$^{*}$ & 24.4 \\

& ShareRobot-Traj.
& 0.32 & 0.34 & 0.39
& \ourcell{0.32} & 0.46 & \rankthree{0.30} & 0.31 & 0.35 & 0.35 & \rankone{0.15} & 0.46 & 0.38 & \ranktwo{0.24$^{*}$} & 0.55 \\





\midrule

\multirow{6}{*}{\benchgroup{Driving\\6 benchmarks}}
& MME-RealWorld
& 32.4 & 67.0 & 24.4
& \ourcell{\rankthree{66.6}} & \rankone{71.2} & 57.0 & 61.7 & \ranktwo{68.7} & 41.6 & 60.3 & 57.9 & 60.1 & 60.0 & 59.6 \\

& MAPLM
& 57.9 & 26.1 & 56.7
& \ourcell{\rankthree{71.3}} & \rankone{77.8} & 26.7 & 63.1 & 68.2 & 29.1 & \ranktwo{74.5} & 24.4 & 22.9 & 22.5 & 31.7 \\

& DriveAction
& 82.2 & 73.5 & 80.0
& \ourcell{78.2} & \rankone{81.3} & 79.9 & 77.1 & 76.3 & 78.1 & \ranktwo{81.0} & 77.2 & 78.3 & 80.5 & \rankthree{80.9} \\

& NuScenesQA
& 37.1 & 16.1 & 22.9
& \ourcell{\rankthree{44.7}} & \rankone{58.8} & 16.6 & 33.5 & 29.5 & 33.1 & \ranktwo{56.7} & 14.8 & 29.3 & 33.2 & 32.3 \\

& NuPlanQA
& 87.5 & 64.2 & 78.2
& \ourcell{\ranktwo{86.7}} & \rankone{91.7} & 82.5 & 82.0 & 81.8 & 78.3 & 73.7 & \rankthree{83.4} & 82.9 & 79.3 & 82.8 \\

& LingoQA
& 76.8 & 64.1 & 60.0
& \ourcell{55.6} & \rankthree{65.8} & \rankone{71.4} & 61.0 & 57.8 & 59.6 & \ranktwo{69.9} & 56.0 & 55.0 & 48.0 & 39.2 \\

\bottomrule
\end{tabular}%
}
\label{perception_planning}
\end{table*}

%% file: tables/navigation.tex
\begin{table}[!htb]
\centering
\caption{\textbf{Performance Comparison on VLN-CE Benchmark.}
We compare ACE-Brain-0.5 with widely regarded open-source baselines on the Val-Unseen split of the R2R and RxR benchmarks.}
\label{tab:vlnce_open_source}
\begin{tabular}{lcccc|cccc}
\toprule
\multirow{2}{*}{\textbf{Method}} 
& \multicolumn{4}{c|}{\textbf{R2R Val-Unseen}} 
& \multicolumn{4}{c}{\textbf{RxR Val-Unseen}} \\
\cmidrule(lr){2-5} \cmidrule(lr){6-9}
& NE$\downarrow$ & OS$\uparrow$ & SR$\uparrow$ & SPL$\uparrow$ 
& NE$\downarrow$ & SR$\uparrow$ & SPL$\uparrow$ & nDTW$\uparrow$ \\
\midrule
NaVid~\cite{zhang2024navid}
& 5.7 & 49.2 & 41.9 & 36.5
& 5.7 & 45.7 & 38.2 & -- \\
Uni-NaVid~\cite{zhang2024uninavid}
& 5.6 & 53.3 & 47.0 & 42.7
& 6.2 & 48.7 & 40.9 & -- \\
NaVILA~\cite{cheng2024navila}
& 5.2 & 62.5 & 54.0 & 49.0
& 6.8 & 49.3 & 44.0 & 58.8 \\
StreamVLN~\cite{wei2025streamvln}
& 5.0 & 64.2 & 56.9 & 51.9
& 6.2 & 52.9 & 46.0 & 61.9 \\
NavFoM~\cite{zhang2025navfom}
& 5.0 & 64.9 & 56.2 & 51.2
& 5.5 & 57.4 & 49.4 & 60.2 \\
RynnBrain-Nav~\cite{dang2026rynnbrain}
& 4.9 & {\bf 71.6} & 58.6 & 49.6
& 6.2 & 56.1 & 49.6 & 59.6 \\
Qwen-VLA-Instruct~\cite{wang2026qwen}
& 5.1 & 69.0 & 57.5 & 51.2
& 5.8 & 59.6 & 47.8 & 57.1 \\
\midrule
\textbf{ACE-Brain-0.5}
& 4.8 & 63.7 & 57.4 & 51.7
& {\bf 4.3} &{\bf 63.8} & 47.9 & 64.6 \\
\textbf{ACE-Brain-0.5-Specialist}
& {\bf 4.2} & 67.4 & {\bf 62.2} & {\bf 56.2}
& 4.5 & 63.6 & {\bf 51.1} & {\bf 67.1} \\
\bottomrule
\end{tabular}
\end{table}

%% file: tables/VLA_LIBERO.tex
\begin{table}[H]
\centering
\caption{\textbf{Performance on LIBERO Benchmark.} }
\label{tab:vla_libero}
\begin{tabular}{l|cccc|c}
\toprule
\textbf{Method} & \textbf{Spatial} & \textbf{Object} & \textbf{Goal} & \textbf{Long} & \textbf{Average} \\
\midrule
OpenVLA~\cite{kim2024openvla} & 84.7 & 88.4 & 79.2 & 53.7 & 76.5 \\
OpenVLA-OFT~\cite{kim2025fine} & 97.6 & 98.4 & 97.9 & 94.5 & 97.1 \\
StarVLA-OFT~\cite{community2026starvla} & 97.8 & 98.6 & 96.2 & 93.8 & 96.6 \\
GR00T N1.6~\cite{nvidia2025gr00tn1} & 97.7 & 98.5 & 97.5 & 94.4 & 97.0 \\
$\pi_{0}$~\cite{black2024pi_0} & 96.8 & 98.8 & 95.8 & 85.2 & 94.1 \\
$\pi_{0.5}$~\cite{pi05_2025} & 98.8 & 98.2 & \textbf{98.0} & 92.4 & 96.9 \\
Mantis~\cite{yang2025mantis} & 98.8 & 99.2 & 94.4 & 94.2 & 96.7 \\
MemoryVLA~\cite{shi2025memoryvla} & 98.4 & 98.4 & 96.4 & 93.4 & 96.7 \\
VITA~\cite{vita} & 95.9 & 98.9 & 95.1 & 96.8 & 96.7 \\
Qwen-VLA-Instruct~\cite{wang2026qwen} & - & - & - & - & 97.9 \\
\midrule
\textbf{ACE-Brain-0.5} & \textbf{100.0} & \textbf{100.0} & 96.0 & \textbf{97.0}& \textbf{98.2} \\
\bottomrule
\end{tabular}
\end{table}

%% file: tables/Simplerenv_Bridge.tex
\begin{table}[t]
\centering
\caption{\textbf{Performance Comparison on Simpler-Env(Bridge) Benchmark.} All of the models are evaluated under a maximum inference horizon of 120 steps except Qwen-VLA-Instruct\cite{wang2026qwen}. The result of Qwen-VLA-Instruct is from \cite{wang2026qwen} while the results of other baseline methods are from \cite{ling2026guide}.}
\renewcommand{\arraystretch}{1.15}
\setlength{\tabcolsep}{6pt}

\begin{tabularx}{0.8\textwidth}{l|*{5}{>{\centering\arraybackslash}X}}
\toprule
Method
& \multicolumn{5}{c}{SIMPLER-Env (Bridge)} \\
\cmidrule(lr){2-6}
& Spoon & Carrot & Cube & Eggplant & Avg \\
\midrule
OpenVLA~\cite{kim2024openvla}
& 4.2 & 0.0 & 8.3 & 45.8 & 14.6 \\

$\pi_0$~\cite{black2024pi_0}
& 50.0 & 41.7 & 29.2 & 70.8 & 47.9 \\

GR00T N1~\cite{nvidia2025gr00tn1}
& 64.5 & 65.5 & 5.5 & 93.0 & 57.1 \\

X-VLA~\cite{zheng2025xvla}
& 95.8 & 75.0 & 62.5 & 70.8 & 76.0 \\

ThinkAct~\cite{huang2025thinkact}
& 37.5 & 8.7 & 58.3 & 70.8 & 43.8 \\

Uni-VLA~\cite{wang2025univla}
& 83.3 & 66.7 & 33.3 & 95.8 & 69.8 \\

GTA-VLA~\cite{ling2026guide}
& \textbf{95.8} & \textbf{87.5} & 66.7 & 75.0 & 81.2 \\

Qwen-VLA-Instruct~\cite{wang2026qwen}
& - & - & - & - & 73.7 \\

\midrule
ACE-Brain-0.5-VLA 
& 75.0 & 79.2 & \textbf{75.0} & \textbf{100.0} & \textbf{82.3} \\
\bottomrule
\end{tabularx}
\label{tab:simplerenv}
\end{table}

%% file: tables/reward_rbm_1m.tex

\begin{table}[H]
\centering
\caption{\textbf{Performance Comparison on RBM-EVAL Benchmarks.}
We compare ACE-Brain-0.5 with widely regarded open-source reward models and general VLM baselines on standard and refined RBM-EVAL ID/OOD benchmarks.}
\label{tab:reward_rbm_1m}
\begin{tabular}{lc|c|cc}
\toprule
\multirow{3}{*}{\textbf{Method}} 
& \multicolumn{4}{c}{\textbf{VOC $r \uparrow$}} \\
\cmidrule(lr){2-5}
& \multicolumn{2}{c|}{\textbf{RBM-EVAL-ID}} 
& \multicolumn{2}{c}{\textbf{RBM-EVAL-OOD}} \\
\cmidrule(lr){2-3} \cmidrule(lr){4-5}
& Standard & Refined & Standard & Refined \\
\midrule
VLAC-8B~\cite{zhai2025vision}
& 0.16 & 0.19 & 0.17 & 0.33 \\
RoboDopamine-8B~\cite{tan2026robodopamine}
& 0.79 & 0.55 & 0.80 & 0.65 \\
RoboReward-8B~\cite{lee2026roboreward}
& 0.82 & 0.51 & 0.88 & 0.60 \\
Robometer-4B~\cite{robometer2026}
& 0.92 & 0.78 & 0.94 & 0.81 \\
\midrule
Qwen3-VL-8B~\cite{bai2025qwen3vltechnicalreport}
& 0.73 & 0.28 & 0.87 & 0.30 \\
\midrule
\textbf{ACE-Brain-0.5}
& \textbf{0.94} & \textbf{0.80} & \textbf{0.96} & \textbf{0.88} \\
\bottomrule
\end{tabular}
\end{table}

%% file: tables/nav_improve.tex
\begin{table}[!htb]
\centering
\caption{\textbf{Ablation of Navigation Self Improvement.}
We evaluate whether the proposed \textit{Navigation Evolving} improves over static trajectory imitation by adding corrective supervision from model rollouts and oracle-recovered trajectories.}
\label{tab:self_improvement}
\begin{tabular}{lcccc|cccc}
\toprule
\multirow{2}{*}{\textbf{Setting}} 
& \multicolumn{4}{c|}{\textbf{R2R Val-Unseen}} 
& \multicolumn{4}{c}{\textbf{RxR Val-Unseen}} \\
\cmidrule(lr){2-5} \cmidrule(lr){6-9}
& NE$\downarrow$ & OS$\uparrow$ & SR$\uparrow$ & SPL$\uparrow$ 
& NE$\downarrow$ & SR$\uparrow$ & SPL$\uparrow$ & nDTW$\uparrow$ \\
\midrule
Static Trajectory Imitation
& 5.7 & 57.8 & 48.6 & 43.2
& 6.0 & 53.3 & 43.2 & 61.6 \\
Navigation Evolving
& 4.8 & 63.7 & 57.4 & 51.7
& 4.3 &63.8 & 47.9 & 64.6 \\
\bottomrule
\end{tabular}
\end{table}

%% file: contexts/5_data.tex
\section{Data and Benchmark}

\subsection{Training Data}

\begin{figure}[!tbp]
    \centering
    \begin{subfigure}{\linewidth}
        \centering
        \includegraphics[width=\linewidth]{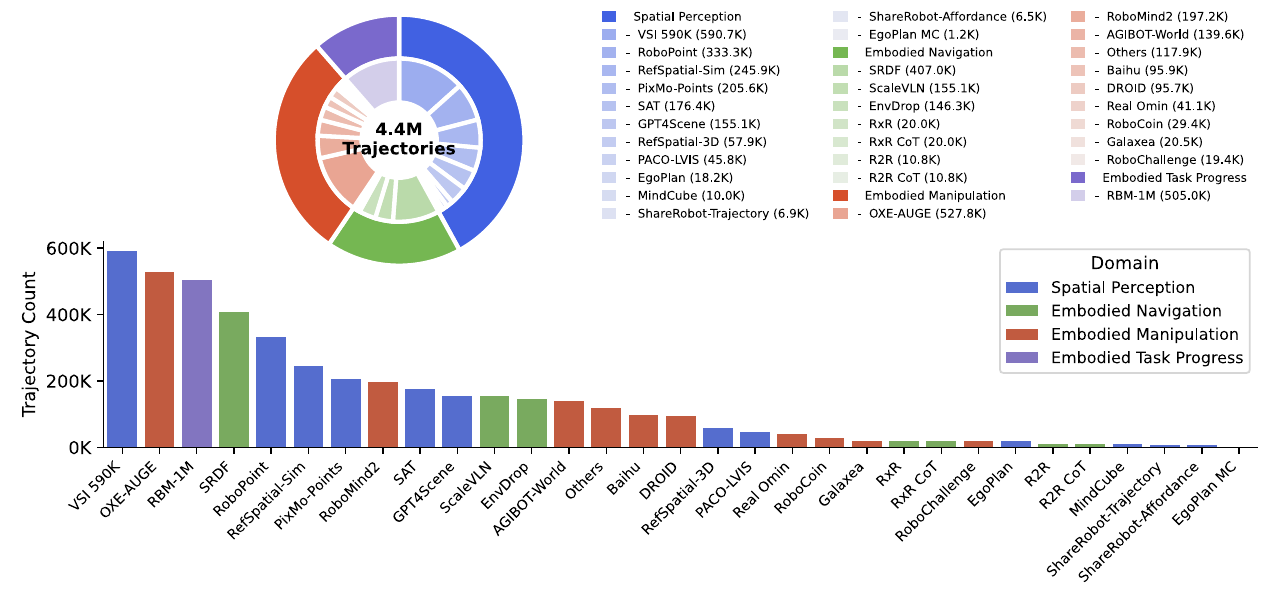}
        \vspace{-0.8cm}
        \caption{Domain distribution and trajectory statistics.}
        \label{fig:data_statistics}
    \end{subfigure}
    \vspace{0.6cm}
    \begin{subfigure}{\linewidth}
        \centering
        \includegraphics[width=\linewidth]{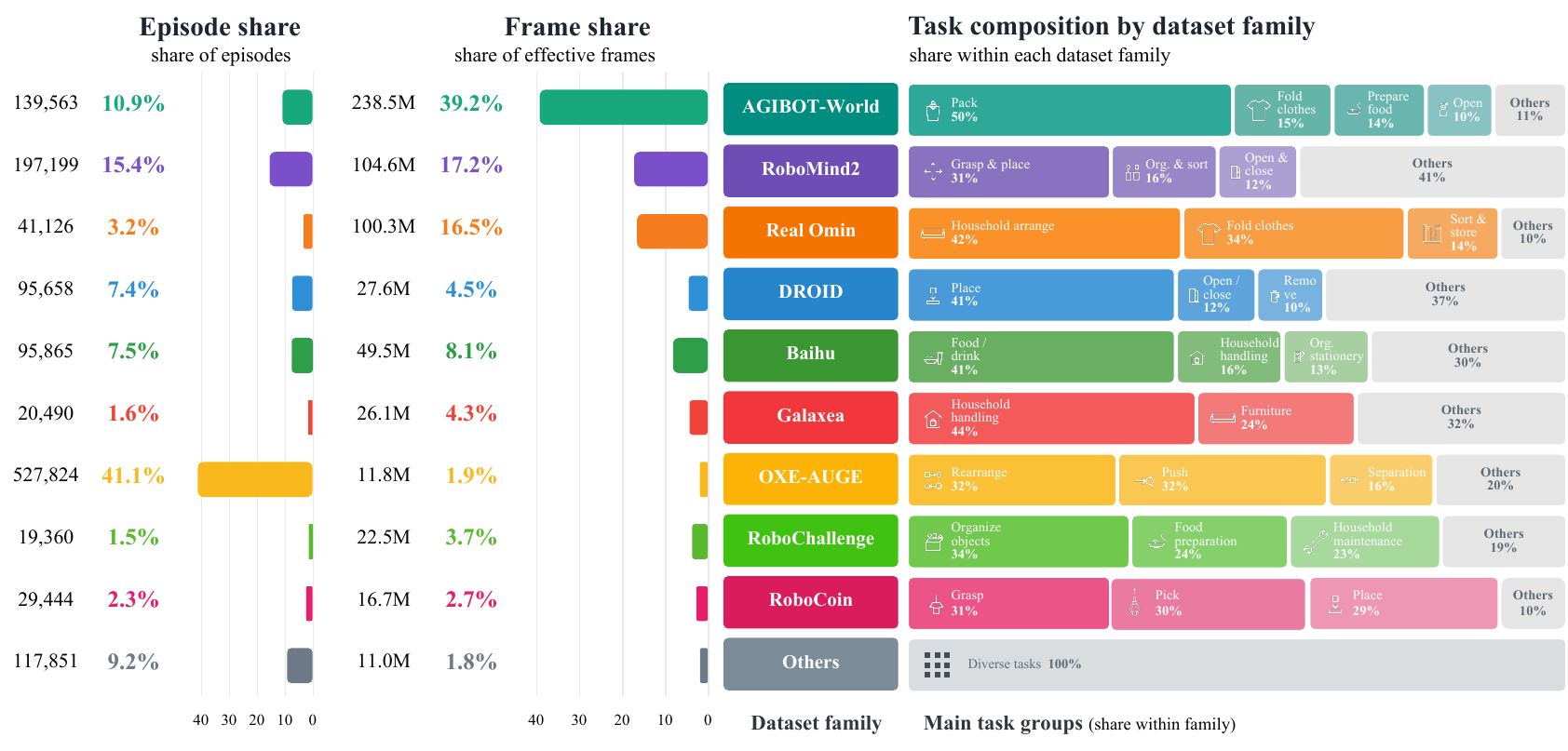}
        \caption{Overview of our pretraining dataset on embodied manipulation tasks.}
        \label{fig:pretrain_data}
    \end{subfigure}
    \vspace{-1cm}
    \caption{Statistics of the pretraining datasets.
    (a) Domain distribution and trajectory statistics.
    This nested pie chart illustrates the proportion of different domains in our dataset and the trajectory statistics for each domain.
    The distribution exhibits a long-tailed characteristic, with a small number of large-scale datasets (e.g., VSI 590K, RBM-1M, SRDF) accounting for the majority of trajectories, while many domain-specific datasets contribute comparatively small counts. 
    (b) Overview of the manipulation pretraining dataset, including dataset family shares and task composition.
    }
    \label{fig:data_overview}
\end{figure}
\paragraph{Spatial Perception Data.} We integrate diverse large-scale QA-oriented datasets spanning image-based, video-based, and multi-view 3D spatial understanding. Specifically, VSI-590K~\cite{vsi} and VLM-3R~\cite{vlm_3r} provide large-scale supervision for directional reasoning, distance estimation, counting, temporal ordering, camera-object interaction, and spatio-temporal understanding across static scenes and monocular videos. GPT4Scene~\cite{gpt4scene} further introduces geometry-aware 3D scene understanding through reconstructed point clouds, BEV representations, and object-consistent multi-view observations, enabling models to learn spatial grounding, dense captioning, and cross-view reasoning in embodied environments. In addition, MindCube~\citep{mindcube} emphasizes reasoning over unobservable space, requiring the model to infer occluded spatial relations, maintain cross-view consistency, and construct implicit spatial mental models under complex camera trajectories. Collectively, these datasets equip the model with comprehensive spatial capabilities including relative direction understanding, metric reasoning, multi-view spatial consistency, camera motion understanding, object-centric 3D reasoning, temporal spatial reasoning, and latent spatial mental model construction.

Beyond public datasets, we further construct an additional set of high-quality spatial reasoning data using a novel Spatial Harness Annotation Pipeline built upon geometrically constrained reasoning. The proposed pipeline formulates spatial reasoning as structured task decomposition over persistent object-centric spatial memory and reusable spatial skills, enabling iterative verification and evidence-grounded annotation for complex embodied spatial understanding scenarios. These additional data further strengthen the model's instruction-following ability, compositional spatial reasoning, long-horizon relation understanding, and robustness under complex embodied environments.

We combine multiple large-scale grounding-oriented datasets covering 2D visual grounding, embodied spatial localization, and action-oriented affordance prediction. RefSpatial~\cite{zhou2025roborefer} provides large-scale supervision for spatial referring and multi-step spatial reasoning across web images, embodied videos, and simulated 3D scenes, enabling models to localize objects and regions under complex relational instructions. PixMo-Points~\cite{deitke2024molmopixmoopenweights} further strengthens point-based grounding ability through exhaustive object pointing annotations, supporting fine-grained localization, counting-by-pointing, and grounding-aware visual explanation. RoboPoint~\cite{yuan2024robopoint} complements these capabilities with action-centric affordance grounding supervision, enabling models to predict actionable regions for robotic manipulation, navigation, and placement under diverse spatial layouts and viewpoints. Together, these datasets endow the model with robust grounding capabilities including fine-grained object localization, spatial referring, point-based grounding, affordance prediction, free-space localization, and instruction-conditioned embodied action grounding.

\paragraph{Embodied Navigation.}
For embodied navigation training, we construct trajectory-level supervision following the data formulation of StreamVLN~\cite{wei2025streamvln}, which organizes navigation as a step-by-step instruction-following action prediction task. We use a mixture of public VLN trajectory datasets, including R2R~\cite{r2r}, RxR~\cite{rxr}, EnvDrop~\cite{envdrop}, and ScaleVLN~\cite{scalevln}. In addition, we incorporate SRDF-400K, an adapted navigation corpus derived from the instruction data of SRDF~\cite{wang2024srdf}. All navigation datasets are unified into the same VLN-CE action interface with executable low-level actions.

Given a natural-language instruction, the agent observes the environment through
egocentric visual inputs and predicts executable navigation actions. Each expert
trajectory is converted into a sequence of step-level training samples, where the
input consists of the instruction, the navigation history, and the current
observation, while the target is the next expert action:
\begin{equation}
    (q, h_t, o_t) \rightarrow a_t,
\end{equation}
where \(q\) denotes the instruction, \(o_t\) is the current observation, and
\(a_t\) is the expert action. The navigation history \(h_t\) contains previous
observations and actions:
\[
    h_t = \{(o_i, a_i)\}_{i=1}^{t-1}.
\]
The action space follows the standard VLN-CE setting, including forward, left, right, and stop. At each step, the model receives the current forward-facing observation together with the accumulated navigation history, allowing it to integrate past visual evidence and previous actions for sequential decision making under partial observability. In addition to action-only supervision, we include progress-aware CoT navigation samples constructed from selected R2R and RxR states. These samples augment navigation supervision with textual reasoning traces grounded in the current observation, including instruction progress, local sub-goals, visible landmarks, and action rationales. Details of discrete-to-continuous trajectory conversion are provided in Appendix~\ref{app:de2ce}.



\paragraph{Embodied Manipulation.}
For manipulation training, we collect a large-scale cross-embodiment trajectory corpus from both public and in-house robot datasets. The corpus covers diverse manipulation scenarios, robot morphologies, camera configurations, and control interfaces, including single-arm, dual-arm, mobile manipulation, and table-top manipulation tasks. Each trajectory is converted into language-conditioned action prediction samples, where the model observes multi-view RGB inputs, task instruction, and proprioceptive states, and predicts a future chunk of executable robot actions. To support heterogeneous embodiments, we canonicalize robot states and actions into a shared schema containing joint states, gripper states, and end-effector pose and rotation, while padding unavailable channels with invalid masks. This unified representation allows trajectories from different robots to supervise a common Action Expert. During pretraining, the action pathway learns chunk-level continuous control from this mixture, providing an action-centric initialization that is later transferred to downstream manipulation benchmarks such as LIBERO and SimplerEnv.

\paragraph{Embodied Task Progress.}
We train progress modeling on RBM-1M~\cite{robometer2026}, a large-scale robotic manipulation dataset containing task instructions, trajectory videos, quality labels, and trajectory identifiers. Following the sampling strategy of Robometer, we convert the original trajectories into 505K multi-image instruction-following samples for Vision-Language Model training.

Let \(x\) denote a task instruction, and let \(\tau=\{I_i\}_{i=1}^{N}\) denote a robot trajectory with \(N\) video frames. For frame-wise progress prediction, we sample an ordered visual sequence \(V=\{v_t\}_{t=1}^{T}\) from \(\tau\), where \(T=8\). Specifically, we randomly sample the start frame, end frame, and part of the intermediate frames, while the remaining frames are selected at approximately uniform temporal intervals within the sampled span. The model takes \((x,V)\) as input and outputs a scaled progress sequence \(\hat{\mathbf{s}}=(\hat{s}_1,\hat{s}_2,\ldots,\hat{s}_T)\), where \(\hat{s}_t \in [0,1000]\). Each value \(\hat{s}_t\) is obtained by scaling the normalized progress score by 1000. For example, normalized labels such as \(0.124, 0.245, 0.387\) are represented as \(124, 245, 387\) during training. This representation avoids generating many repeated decimal tokens, such as "\texttt{0}" and "\texttt{.}", and provides more effective token-level supervision for autoregressive training. Larger values still indicate states closer to task completion. At evaluation time, we manually divide the generated values by 1000 to recover normalized progress scores in \([0,1]\). The progress sequence is generated through the autoregressive text interface of the VLM.

We also construct pairwise preference samples for relative progress comparison. For each sample, we select two trajectories \(\tau^A=\{I_i^A\}_{i=1}^{N_A}\) and \(\tau^B=\{I_i^B\}_{i=1}^{N_B}\) under the same task instruction \(x\). For the chosen trajectory, the first and last frames are always included, and the remaining frames are uniformly sampled in temporal order. For the rejected trajectory, we use the same stochastic sampling strategy as in frame-wise progress modeling. This produces two visual sequences \(V^A=\{v_t^A\}_{t=1}^{T}\) and \(V^B=\{v_t^B\}_{t=1}^{T}\). The model is then asked to output a textual label \(y \in \{\texttt{A},\texttt{B}\}\), indicating which trajectory shows greater progress toward the task goal. During training, the chosen trajectory is randomly assigned to position A or B to reduce positional bias.

This mixed training format enables the same model to learn two related abilities: estimating fine-grained progress from a single trajectory and comparing the relative progress of two trajectories under the same instruction. Both tasks are formulated through the same autoregressive VLM interface, with all supervision represented as text.

\subsection{Evaluation Benchmark}

\paragraph{Spatial Perception and Decision Making.}
We evaluate spatial perception and planning with benchmarks that cover egocentric spatial memory, multi-view reasoning, 3D scene understanding, language grounding, affordance localization, and trajectory prediction. \textbf{VSI}~\cite{vsi} contains visual spatial intelligence questions built from egocentric image and video observations. Inputs are visual observations with a natural-language question, and the model predicts the answer. It evaluates whether a model can perceive, remember, and recall spatial relations such as direction, distance, object arrangement, and temporal order. Evaluation follows the benchmark answer-matching protocol and reports accuracy. \textbf{MMSI}~\cite{mmsi} is a multi-image spatial intelligence benchmark. Inputs are multiple related images and a spatial question, requiring the model to integrate evidence across views. It evaluates cross-image spatial relation understanding, viewpoint reasoning, and spatial consistency, and reports answer accuracy. \textbf{MindCube}~\cite{mindcube} evaluates spatial mental modeling from limited views. Inputs are partial multi-view observations and questions whose answers often require reasoning about unobserved or occluded space. It tests whether the model can build an internal spatial representation beyond directly visible evidence, and reports question-answering accuracy.

\textbf{ScanQA}~\cite{scanqa} is a 3D question-answering benchmark for indoor scene understanding. Inputs are reconstructed 3D scans paired with natural-language questions, and the model predicts free-form answers about objects, attributes, counts, and spatial relations. It evaluates 3D scene comprehension and reports standard QA correctness and language-generation metrics. \textbf{SQA3D}~\cite{sqa3d} studies situated question answering in 3D scenes. Inputs include a 3D scene, an agent situation specified by position and orientation, and a question. It evaluates egocentric and situated spatial reasoning, and reports answer accuracy. \textbf{Scan2Cap}~\cite{scan2cap} is a dense captioning benchmark in RGB-D scans. Inputs are 3D scenes with target object regions, and the model generates object-centric captions grounded in the surrounding context. Considering the length of this benchmark, we set the maximum output length to 32 tokens for evaluating closed-source models. It evaluates context-aware 3D captioning using standard captioning metrics such as CIDEr, BLEU, METEOR, and ROUGE. \textbf{ScanRefer}~\cite{scanrefer} evaluates 3D object localization from natural language. Inputs are an RGB-D scan and a referring expression, and the model selects the referred 3D object. Evaluation reports localization accuracy under 3D IoU thresholds. \textbf{Multi3DRef}~\cite{multi3drefer} extends 3D visual grounding to descriptions that may refer to zero, one, or multiple objects. Inputs are 3D scans with textual descriptions, and the model predicts the target object set. It evaluates multi-object reference resolution using F1-style grounding metrics under 3D IoU matching.

\textbf{EmbSpatial}~\cite{embspatial} benchmarks spatial understanding for embodied tasks with large vision-language models. Inputs are embodied visual observations and natural-language questions about spatial relations, object states, and action-relevant scene structure. It evaluates whether visual spatial understanding can support embodied decision making, and reports answer accuracy. \textbf{ERQA}~\cite{gemini_robotics} evaluates embodied reasoning question answering in robot-relevant scenes. Inputs are visual observations and questions requiring physical-world reasoning about objects, relations, and task context. It evaluates embodied scene understanding and reports QA accuracy. \textbf{RoboSpatial}~\cite{robospatial} contains 2D and 3D spatial reasoning problems designed for robotics. Inputs are visual observations and language queries about robot-centric spatial relations, object configurations, and interaction-relevant geometry. It evaluates spatial understanding for robotic perception and planning, and reports accuracy. \textbf{RefSpatial}~\cite{zhou2025roborefer} is a spatial referring benchmark from RoboRefer. Inputs include visual observations and spatial referring instructions, and the model predicts object or point references. It evaluates multi-step spatial reference resolution and reports grounding accuracy. \textbf{PointArena}~\cite{pointarena} probes multimodal grounding through language-guided pointing. Inputs are images and natural-language target descriptions, and the model outputs point coordinates. It evaluates whether models can ground language to precise visual locations, and reports pointing accuracy under distance or region-matching criteria. \textbf{ShareRobot-Traj}~\cite{robobrain} evaluates trajectory prediction for robot-relevant spatial planning. Inputs are embodied scenes and task instructions, and the model predicts motion trajectories or waypoint-like action traces. It evaluates whether spatial reasoning can be converted into action-oriented paths, and reports trajectory prediction error. \textbf{RoboAfford}~\cite{roboafford} is an affordance benchmark for robot manipulation. Inputs are visual observations and manipulation-oriented language prompts, and the model predicts object or spatial affordance regions. It evaluates object affordance and spatial affordance understanding, and reports affordance localization accuracy.


\paragraph{Decision Making.} We evaluate autonomous-driving decision making and traffic-scene understanding on six benchmarks. \textbf{MAPLM}~\cite{cao2024maplm} contains 6,000 multiple-choice map/traffic questions and reports accuracy as the average of question-level accuracy and frame-level accuracy, where a frame is correct only if all questions from the same frame are answered correctly. \textbf{DriveAction}~\cite{hao2025driveaction} includes 16,185 multi-frame questions for action and behavior understanding, evaluated by exact-match accuracy. \textbf{LingoQA}~\cite{marcu2024lingoqa} evaluates open-ended language reasoning over driving video clips using LingoJudge accuracy. \textbf{NuScenes-QA}~\cite{qian2024nuscenes} contains 83,335 short-answer VQA questions over six-camera driving observations, evaluated by exact match. NuPlanQA~\cite{park2025nuplanqa} consists of 1,801 multiple-choice questions for traffic-element understanding and driving decision-making. \textbf{MME-RealWorld}~\cite{zhang2024mme} further evaluates real-world perception and reasoning in autonomous driving, using multiple-choice accuracy.

\paragraph{Embodied Interaction.}
\textbf{VLN-CE}~\cite{vlnce} evaluates language-guided navigation in continuous environments. We evaluate ACE-Brain-0.5 on R2R and RxR under the VLN-CE setting, where the agent follows natural-language instructions from egocentric visual observations and predicts executable actions in a closed-loop manner. This benchmark tests instruction grounding, visual history integration, spatial progress tracking, and sequential decision-making under partial observability. For R2R, we report NE, OS, SR, and SPL; for RxR, we report NE, SR, SPL, and nDTW.

\textbf{LIBERO}~\cite{liu2023libero} is a language-conditioned manipulation benchmark for lifelong robot learning. Inputs are robot observations and task instructions, and the policy outputs executable manipulation actions. It contains Spatial, Object, Goal, and Long task suites, which respectively test spatial relation changes, object generalization, goal-conditioned manipulation, and long-horizon task execution. Evaluation reports task success rate. \textbf{SimplerEnv-Bridge}~\cite{li2024simplerenv} evaluates real-world robot manipulation policies in simulation using tasks aligned with Bridge-style manipulation data. Inputs are visual observations and language instructions, and the policy is evaluated through task-specific closed-loop rollouts following the corresponding evaluation settings. It evaluates object-centric manipulation robustness across target objects, and reports success rate.

\paragraph{Self Monitoring.}
RBM-EVAL-ID and RBM-EVAL-OOD~\citep{robometer2026} are evaluation splits for robotic progress estimation. Each example consists of a language instruction and trajectory frames, and the model predicts progress scores along the trajectory. RBM-EVAL-ID evaluates held-out in-distribution trajectories, while RBM-EVAL-OOD tests generalization to unseen embodiments, camera viewpoints, and scenes. The standard evaluation reports VOC between the predicted progress scores and the ground-truth progress labels.

While the original RBM-EVAL is useful for measuring progress prediction, its reward-alignment setting mainly evaluates successful forward-execution trajectories. In these trajectories, progress usually increases as time advances. This creates a possible shortcut. A model may obtain high progress alignment by assigning larger scores to later frames, or by recognizing a single visual state close to task completion. Such behavior does not necessarily show that the model understands the temporal development of the task. This limitation is important for manipulation tasks whose success depends on action direction and state transition. For example, moving an object from left to right and moving it from right to left may contain similar visual states, but they correspond to opposite progress under a given instruction. In such cases, a single frame is often insufficient; the model must observe the full trajectory to infer whether the agent is moving toward or away from the goal.

To better evaluate this ability, we construct \textbf{RBM-EVAL-Refined} as a controlled complement to the original benchmark. It focuses on temporal understanding and action-direction reasoning in reward alignment. We keep the original evaluation on successful trajectories, and further select a subset of tasks whose progress cannot be reliably judged from a single frame. For each selected successful trajectory, we create a reversed version by playing the video frames in the opposite temporal order. The reversed trajectory contains nearly the same visual states as the original one, but the action direction is inverted. Therefore, instead of moving toward the task goal, the agent is moving away from it or undoing the completed task. Its progress labels are also reversed accordingly. This design controls for visual content and state distribution, while testing whether the model can use the full trajectory to reason about task progress. Appendix~\ref{app:rbm-eval-refined-details} shows the specific tasks selected for refinement.

%% file: contexts/6_conclusion.tex
\section{Conclusions}

We have presented \myname{}, a unified embodied foundation model that organizes robot intelligence into five tightly coupled cognitive functions: spatial perception, decision making, embodied interaction, self-monitoring, and self-improvement. Built on the spatial-intelligence scaffold of ACE-Brain-0, a single 8B backbone has directly instantiated the first four functions as a closed Perception--Planning--Action--Evaluation loop, and the fifth function, self-improvement, has been realized through a companion execution-state framework and a navigation evolving framework that turn the model's own rollouts into corrective training signal. To reconcile these heterogeneous capabilities without cross-task interference, we have introduced SSR+, extending Scaffold--Specialize--Reconcile with a lightweight Reactivate stage. Across more than fifteen benchmarks spanning spatial cognition, grounding, navigation, manipulation, and progress evaluation, \myname{} has improved over ACE-Brain-0 on the large majority of spatial and grounding benchmarks, has achieved competitive navigation and manipulation performance, and has provided strong progress-estimation ability under both in-distribution and out-of-distribution settings, showing that these five functions can reinforce one another within a single model rather than requiring separate specialized systems.
The current self-improvement mechanisms remain lightweight and domain-specific. Extending them into a general, model-level self-evolution mechanism, and scaling \myname{} to broader embodiments and longer-horizon tasks, are natural directions for future work.

%% file: contributions_author_list.tex
\section{Contributions and Author List}
\label{sec:contribution_author_list}

\vspace{0.8em}

\noindent
\begin{minipage}[t]{0.42\textwidth}
\textcolor{blue}{\textbf{Core Contributors}}

\vspace{0.8em}

\begin{itemize}
\setlength{\itemsep}{5pt}
\setlength{\parskip}{0pt}
\setlength{\parsep}{0pt}

    \item Ziyang Gong$^{}$
    \item Haoming Gu$^{}$
    \item Zehang Luo$^{}$
    \item Tianyi Zhang$^{}$
    \item Tao Tao$^{}$
    \item Yixiao Chi$^{}$
    \item Zhe Liu$^{}$
    \item Lingsi Zhu$^{}$
    \item Jingyuan Liu$^{}$
    \item Anke Tang$^{}$
    \item Zhi Hou$^{\dagger,\ddagger}$
    \item Xue Yang$^{\ddagger}$
    \item Dacheng Tao$^{\ddagger}$
    \item Xiaogang Wang$^{\ddagger}$

\end{itemize}
\end{minipage}
\hfill
\begin{minipage}[t]{0.48\textwidth}
\textcolor{blue}{\textbf{Contributors}}

\vspace{0.8em}

\begin{itemize}
\setlength{\itemsep}{5pt}
\setlength{\parskip}{0pt}
\setlength{\parsep}{0pt}

    \item Songze Li
    \item Yilun Kong
    \item Ningjing Liu
    \item Tianyu Zhu
    \item Yunpeng Qing
    \item Shuang Luo
    \item Xiang Liu
    \item Shi Fu
    \item Dawei Nie
    \item Sixiang Liu
    \item Zhexi Wen
    \item Feng Pan
    \item Xiaofeng Wang
    \item Chunxiao Liu
    \item Junchi Yan
    \item Hengshuang Zhao


\end{itemize}
\end{minipage}

\vfill

\noindent\rule{0.32\textwidth}{0.4pt}

\vspace{0.2em}

{\footnotesize
\noindent
$^{\dagger}$ Project Leader. \\
$^{\ddagger}$ Corresponding author.
}

%% file: appendix/appendix.tex
\subsection{Grounding-based Manipulation}

\begin{figure}
    \centering
    \includegraphics[width=0.9\linewidth]{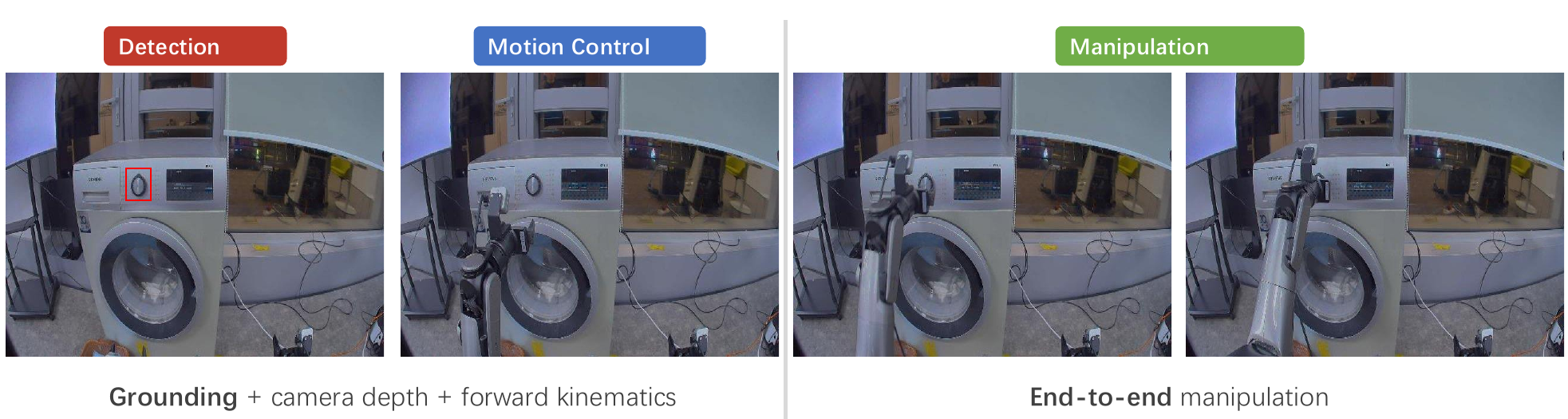}
    \caption{From Detection to End-to-End Manipulation}
    \label{fig:grounding_manipulation}
\end{figure}

Figure~\ref{fig:grounding_manipulation} presents a Grounding + end-to-end manipulation demonstration. The system first uses grounding to localize the target object and drives the robot arm to approach it, then switches to the end-to-end policy for fine-grained closed-loop manipulation. This coarse-to-fine pipeline couples reliable spatial grounding with precise low-level control to accomplish the task.

\subsection{Architecture Details}

\textbf{Manipulation.} Considering computational efficiency, we devise a 2B variant of \myname{}. Its vision-language branch is kept frozen and uses Qwen3-VL-2B-Instruct as the backbone, which consists of a 28-layer language transformer and a 24-block visual encoder. The frozen VLM, a trainable FastVision encoder, and an action expert are jointly organized through a mixture-of-transformer design, where each expert keeps its own weights but they interact within a shared attention. The FastVision encoder, based on DINOv3 ViT-L/16, processes three 224×224 camera views and produces $ 201 \times 3=603$ visual tokens that provide dense visual cues for control, which are linearly projected into the action-expert feature space. The action expert predicts a 10-step action sequence, and adaLN conditioning is used to inject the flow-matching timestep. Through the shared attention of the mixture-of-transformer, the VLM, FastVision, and action tokens are fused together, allowing low-level control to be conditioned on both high-level semantics and dense visual cues.

\begin{figure}[!htb]
    \centering
    \includegraphics[width=0.65\linewidth]{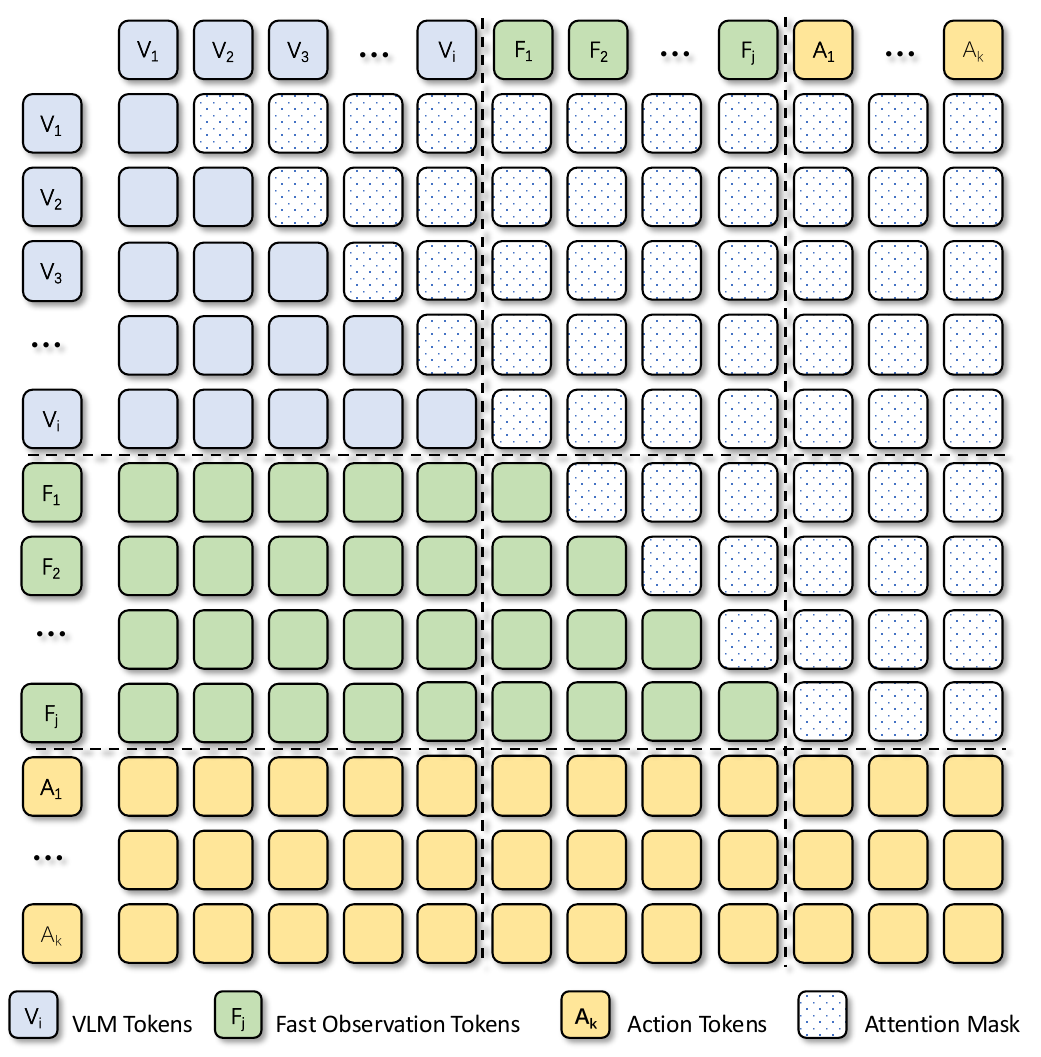}
    \caption{Attention mask of ACE-Brain-0.5. The input sequence consists of VLM tokens (blue), Fast Observation tokens (green), and Action tokens (yellow). VLM tokens employ causal attention to preserve the autoregressive perception and reasoning capability of the pretrained VLM, while Action tokens use full attention within their respective streams to enable efficient spatio-temporal feature aggregation and action generation. This hybrid attention design maintains the original VLM capability while facilitating efficient policy learning.}
    \label{fig:attention_mask}
\end{figure}

To preserve the perception capability of the VLM, we employ causal attention in the VLM, while using full attention for the action tokens as illustrated in Figure~\ref{fig:attention_mask}.

\subsection{Discrete-to-Continuous Trajectory Conversion.}
\label{app:de2ce}

Following the VLN-CE protocol, we convert discrete nav-graph trajectories into executable continuous trajectories in Habitat. Given a discrete trajectory
\begin{equation}
    \tau_i^{G}=[v_{i,1},\ldots,v_{i,T_i}]
    \quad \Rightarrow \quad
    \tau_i^{C}=[(w_{i,1},R_{i,1}),a_{i,1},\ldots,a_{i,L_i},w_{i,T_i}],
\end{equation}
where \(v_{i,t}\in\mathcal{V}_{G}\) denotes a Matterport navigation viewpoint and \(w_{i,t}\) is its corresponding continuous waypoint in Habitat. For each viewpoint, we first query MatterSim to obtain its panoramic 3D location \(p_{i,t}^{G}=(x_{i,t},y_{i,t},z_{i,t})\). The point is then transformed into the Habitat coordinate system as \(\tilde{w}_{i,t}=(x_{i,t},z_{i,t}-1.25,-y_{i,t})\), and projected to the nearest valid navigable point on the Habitat navmesh \(\mathcal{M}\). If the snapped point is valid, we keep the original horizontal location and use the snapped floor height; otherwise, the approximated point is used. For each consecutive waypoint pair \((w_{i,t},w_{i,t+1})\), a Habitat shortest-path follower generates low-level actions under physical simulation, where \(a_{i,l}\in\{\textsc{Forward}_{0.25m},\textsc{Left}_{15^\circ},\textsc{Right}_{15^\circ},\textsc{Stop}\}\). Trajectories whose endpoints are unreachable or whose action length exceeds the maximum step budget are discarded. Finally, we store the continuous training annotation as \(\mathcal{A}_i=[-1,a_{i,1},\ldots,a_{i,L_i}]\), where \(-1\) denotes the dummy start action.

\subsection{Construction of RBM-EVAL-Refined}
\label{app:rbm-eval-refined-details}

RBM-EVAL-ID and RBM-EVAL-OOD are built from the Robometer~\citep{robometer2026} evaluation splits. RBM-EVAL-ID contains held-out in-distribution trajectories from datasets such as OXE Eval Suite, RACER, MetaWorld, and LIBERO, while RBM-EVAL-OOD contains trajectories from unseen embodiments including Franka, Koch bimanual, Trossen, xArm, and SO101 robots. To construct RBM-EVAL-Refined, we select tasks whose progress cannot be reliably inferred from a single static frame, but instead depends on temporal ordering, motion direction, or state change across the trajectory. For each selected task, we keep the original forward trajectory and add its reversed version as a negative temporal control. The selected tasks and their temporal ambiguity are summarized in Table~\ref{tab:rbm_eval_refined_tasks}.

\input{tables/rbm_eval_refined_tasks}

\subsection{Theoretical Analysis of SSR+}
\label{app:theory}

We state two results that underpin the SSR+ design, adapted from ACE-Brain-0~\citep{acebrain0} to the task-interface setting of ACE-Brain-0.5.
Proofs follow those in Appendix~A of~\citet{acebrain0} with morphologies replaced by task interfaces.

\paragraph{Gradient Interference and the Necessity of Isolation (Stage~2).}
Let $\mathcal{T}=\{t_1,\ldots,t_K\}$ denote the $K$ tasks (QA, grounding, navigation, manipulation, progress estimation).
Each task $t_i$ induces a risk $R_i(\theta)$ and gradient $g_i(\theta):=\nabla_\theta R_i(\theta)$.
Consider a joint gradient step with weights $w\in\Delta_K$:
$\theta^{+}=\theta-\eta\sum_{j=1}^K w_j g_j(\theta)$.

\begin{theorem}[One-step interference bound~\citep{acebrain0}]
Under $L$-smoothness, for any task $t_i\in\mathcal{T}$,
\begin{equation}
R_i(\theta^{+}) \leq R_i(\theta)
  - \eta\!\Bigl(w_i\|g_i(\theta)\|^2
    + \sum_{j\neq i} w_j\langle g_i(\theta), g_j(\theta)\rangle\Bigr)
  + \frac{L\eta^2}{2}\Bigl\|\sum_{j=1}^K w_j g_j(\theta)\Bigr\|^2.
\label{eq:interference_bound}
\end{equation}
\end{theorem}

The cross-terms $\sum_{j\neq i}w_j\langle g_i(\theta),g_j(\theta)\rangle$ are the \emph{interference} terms.
When gradients from heterogeneous output interfaces (e.g., bounding-box coordinates vs.\ navigation actions vs.\ progress scalars) are misaligned, these terms are persistently negative and joint training may \emph{increase} $R_i$ despite gradient descent.
Stage~2 of SSR+ removes these terms by construction: each expert is optimized on its own gradient alone, eliminating the dominant source of interference before reconciliation.

\paragraph{Spatial Scaffold as a Universal Bridge (Stage~1).}
Let $\theta_{\mathrm{spatial}}$ denote the ACE-Brain-0 checkpoint (Stage~1 scaffold).
Under a Lipschitz condition on the geometry-conditioned loss (sensitivity $L_g$), a scaffold recoverability error $\varepsilon_g$, and a geometric distribution shift $\delta_i$ between the scaffold and task $t_i$:

\begin{theorem}[Scaffold-to-task transfer bound~\citep{acebrain0}]
For any task $t_i\in\mathcal{T}$,
\begin{equation}
R_i(\theta_{\mathrm{spatial}})
  \leq R_{\mathrm{sp}}(\theta_{\mathrm{spatial}})
  + C_i\,\delta_i
  + 2L_g\,\varepsilon_g
  + \varepsilon_i,
\label{eq:transfer_bound}
\end{equation}
where $R_{\mathrm{sp}}$ is the risk under the spatial scaffold distribution, $\delta_i$ quantifies the geometric distribution shift, $\varepsilon_g$ is the scaffold recoverability error, and $\varepsilon_i$ aggregates task-specific residuals.
\end{theorem}

Eq.~\eqref{eq:transfer_bound} has two practical implications for SSR+.
First, stronger spatial pretraining (smaller $\varepsilon_g$) directly reduces the initialization risk for every downstream task, lowering the specialization cost in Stage~2.
Second, broader scaffold coverage (smaller $\delta_i$) improves cross-task transfer; ACE-Brain-0's training over spatial cognition, autonomous driving, low-altitude sensing, and embodied understanding is precisely designed to minimize $\delta_i$ across diverse embodied tasks.
Together, these results provide theoretical grounding for the SSR+ design choices: isolation before merging, and spatial scaffolding as the universal initialization.

\subsection{Visualization}

\begin{figure}[htb]
    \centering
    \includegraphics[width=\linewidth]{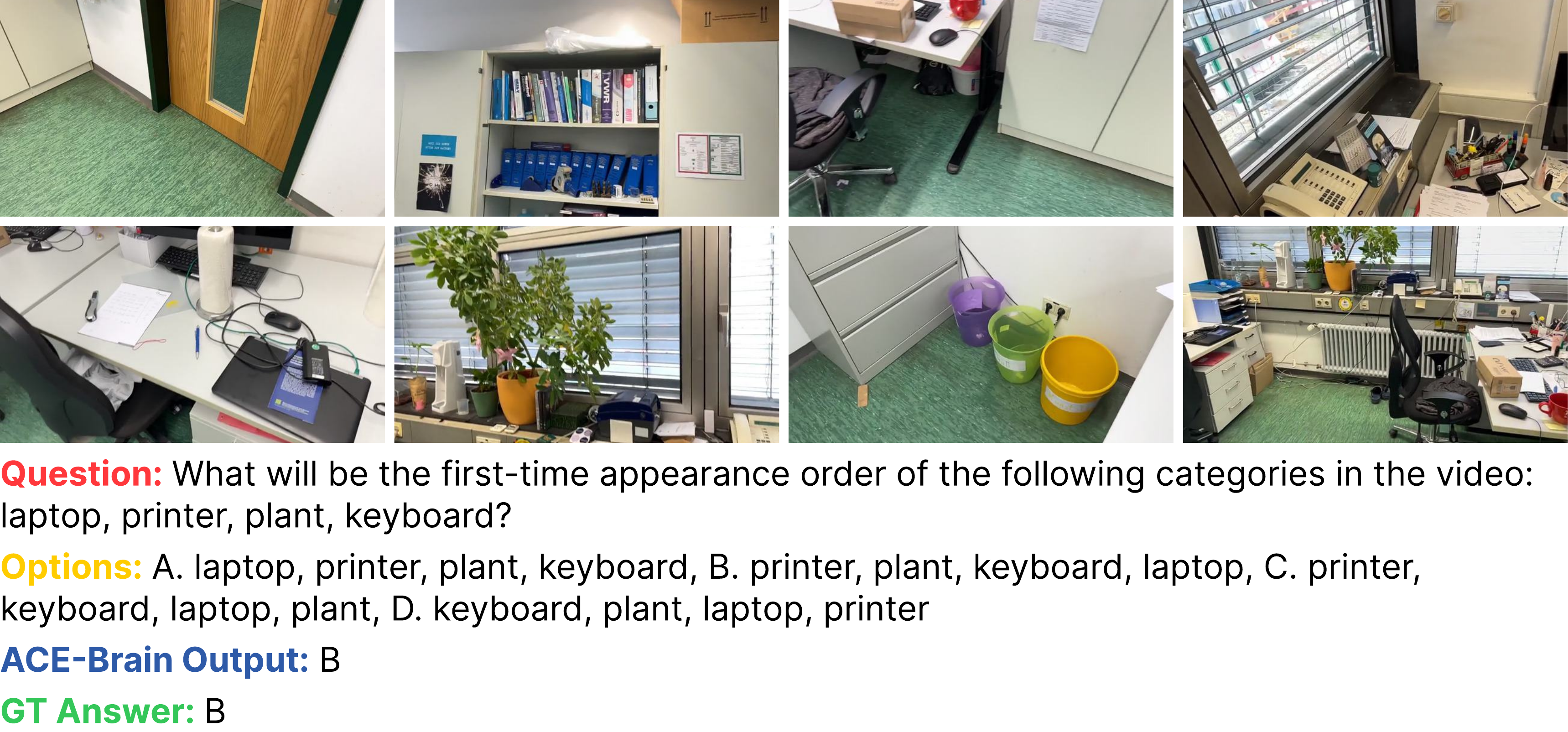}
    \caption{Example 1 of VSI Benchmark.}
    \label{fig:vsi-bench-1}
\end{figure}
\begin{figure}[htb]
    \centering
    \includegraphics[width=\linewidth]{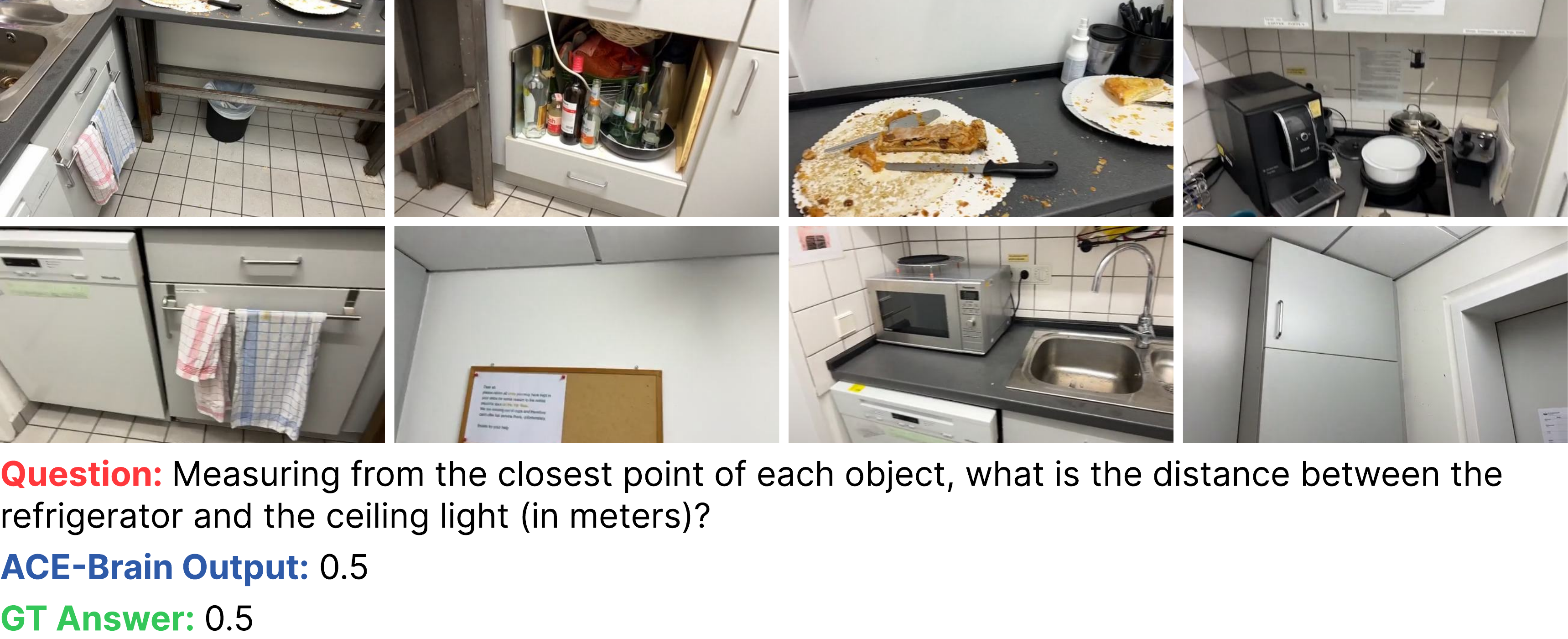}
    \caption{Example 2 of VSI Benchmark.}
    \label{fig:vsi-bench-2}
\end{figure}
\begin{figure}[htb]
    \centering
    \includegraphics[width=\linewidth]{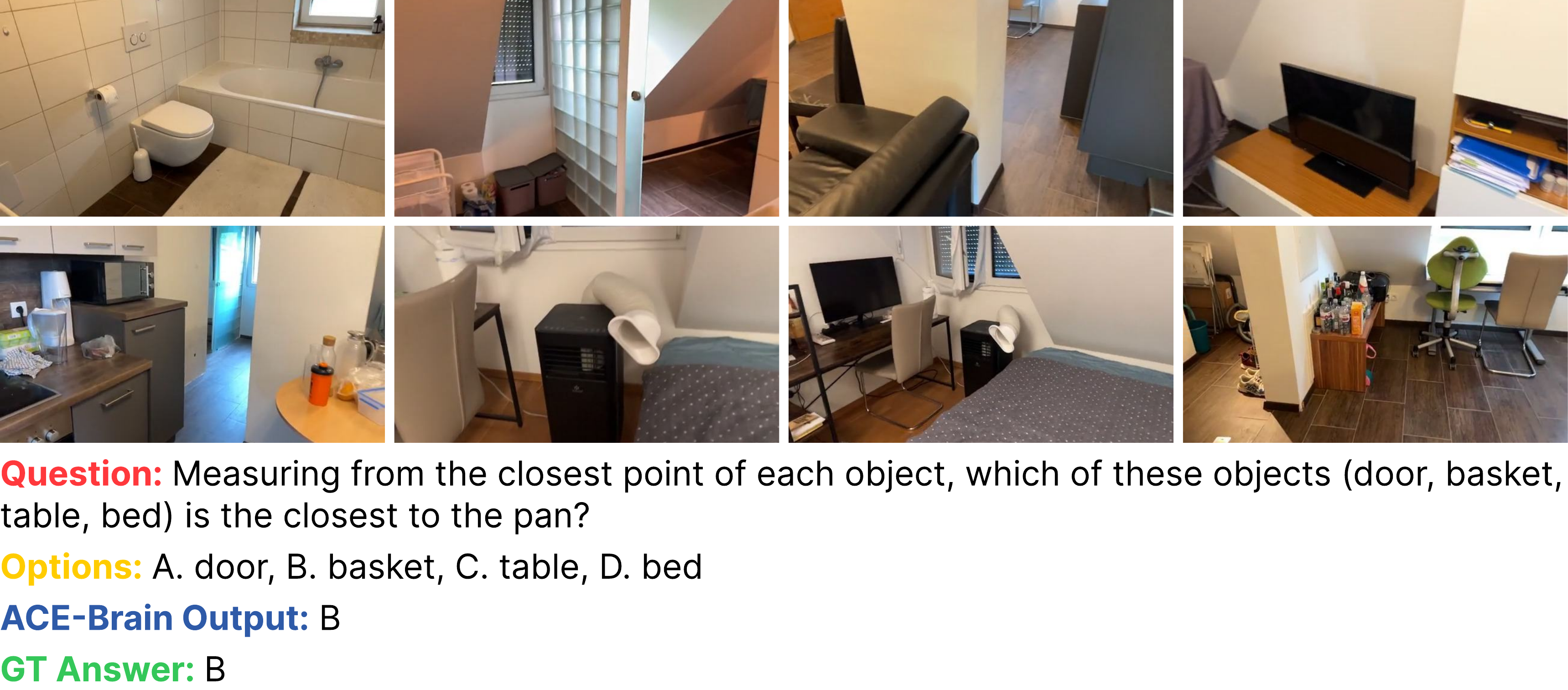}
    \caption{Example 3 of VSI Benchmark.}
    \label{fig:vsi-bench-3}
\end{figure}
\begin{figure}[htb]
    \centering
    \includegraphics[width=\linewidth]{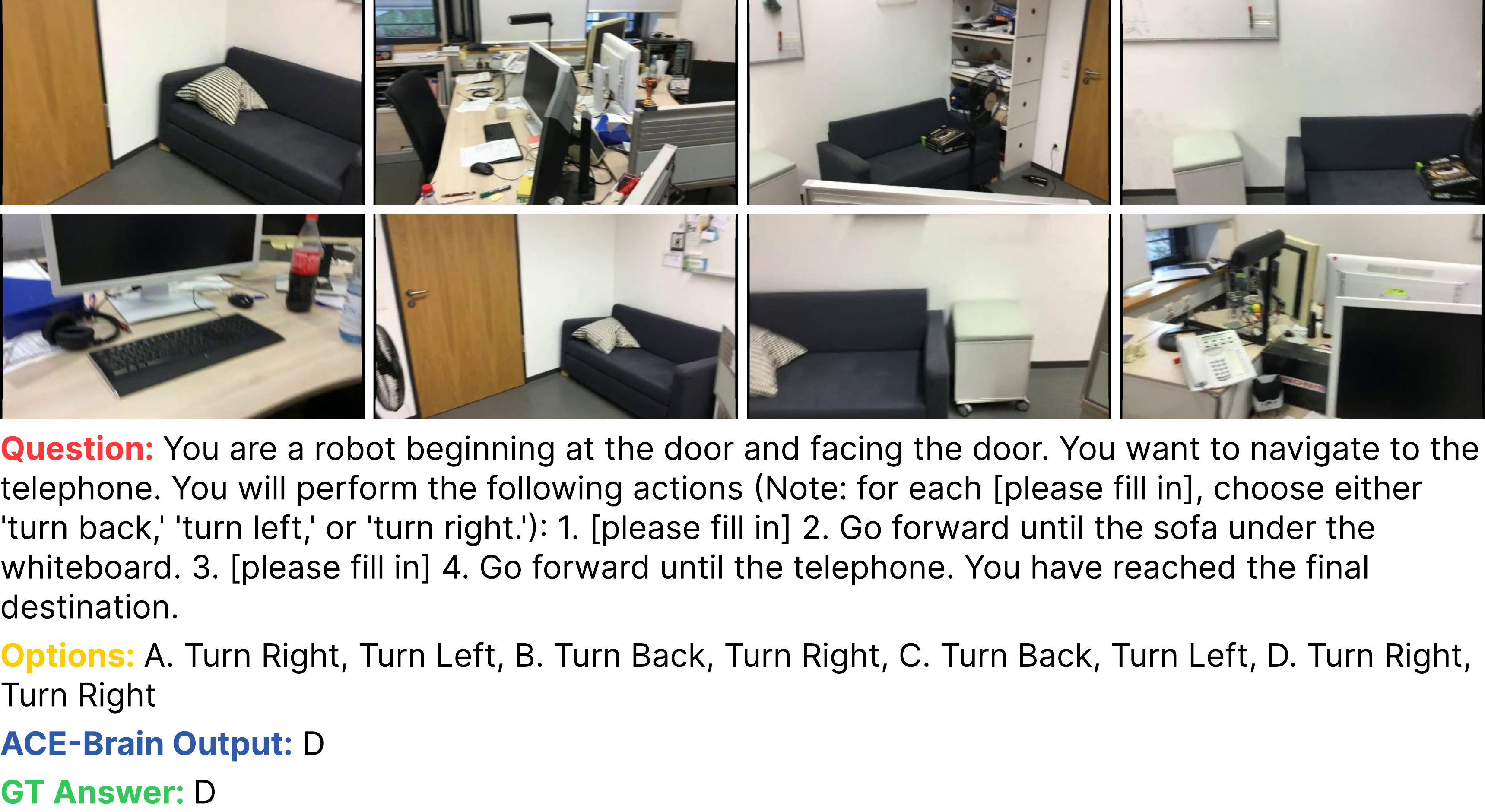}
    \caption{Example 4 of VSI Benchmark.}
    \label{fig:vsi-bench-4}
\end{figure}

\begin{figure}
    \centering
    \includegraphics[width=\linewidth]{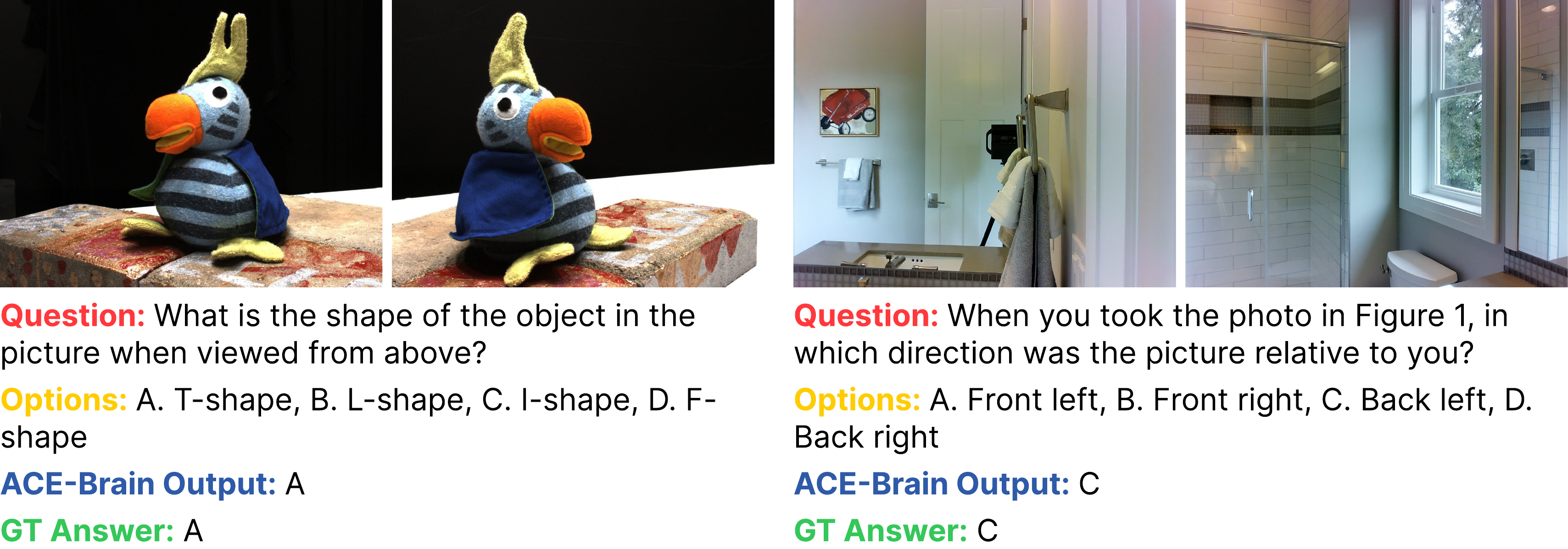}
    \caption{Examples 1, 2 of MMSI Benchmark.}
    \label{fig:mmsi-bench-12}
\end{figure}
\begin{figure}
    \centering
    \includegraphics[width=\linewidth]{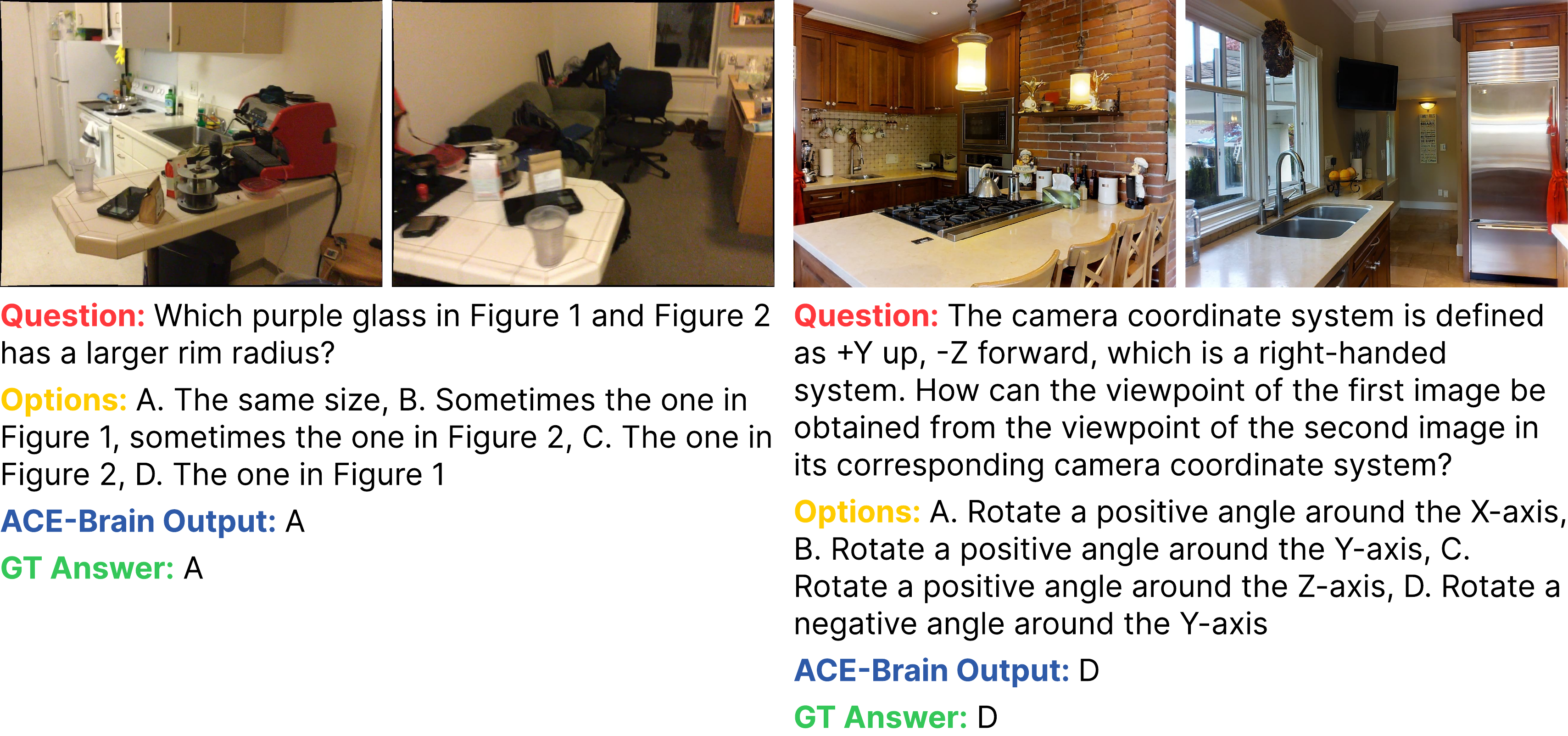}
    \caption{Examples 3, 4 of MMSI Benchmark.}
    \label{fig:mmsi-bench-34}
\end{figure}

\begin{figure}
    \centering
    \includegraphics[width=\linewidth]{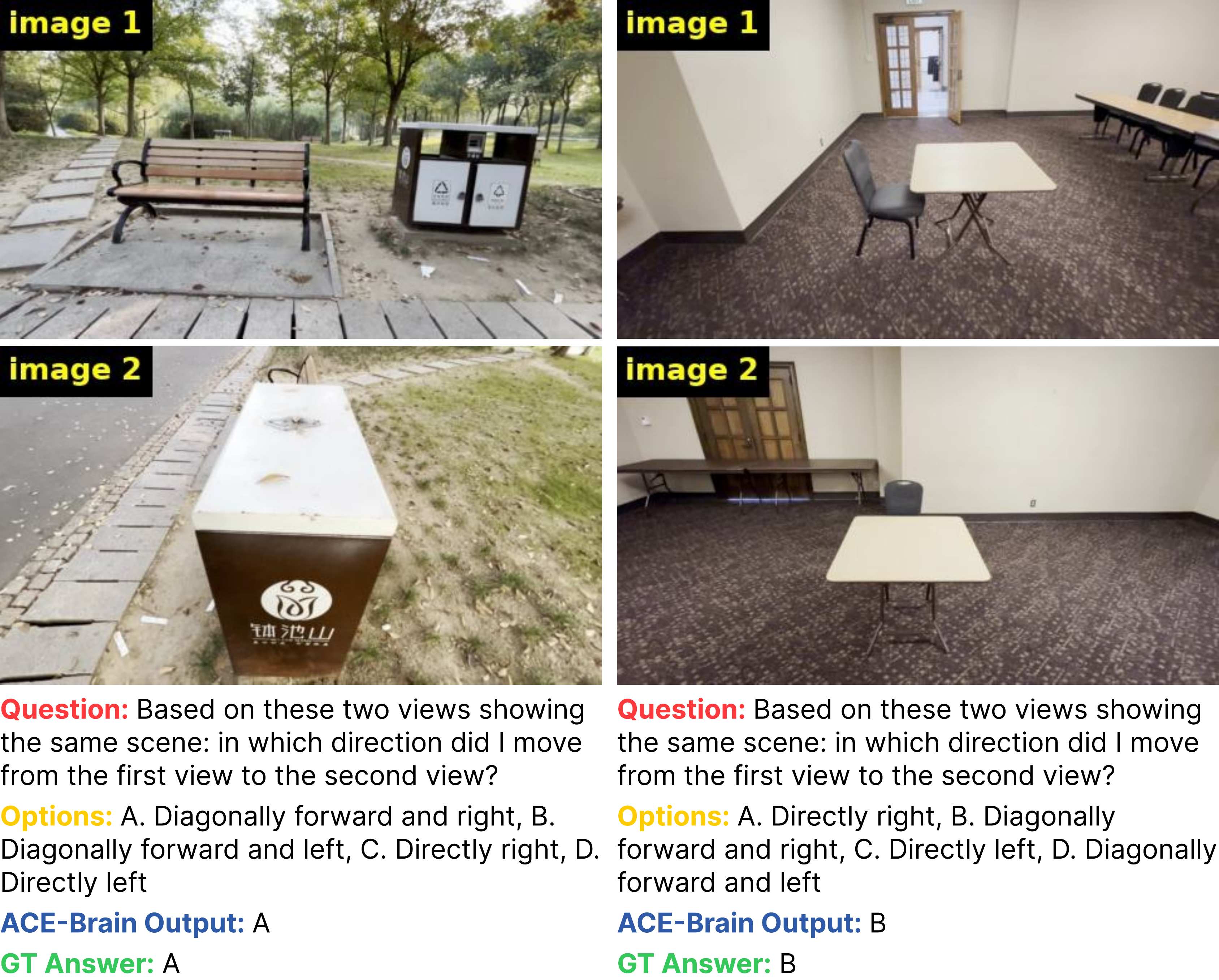}
    \caption{Examples 1, 2 of MindCube Benchmark.}
    \label{fig:minicube-12}
\end{figure}
\begin{figure}
    \centering
    \includegraphics[width=\linewidth]{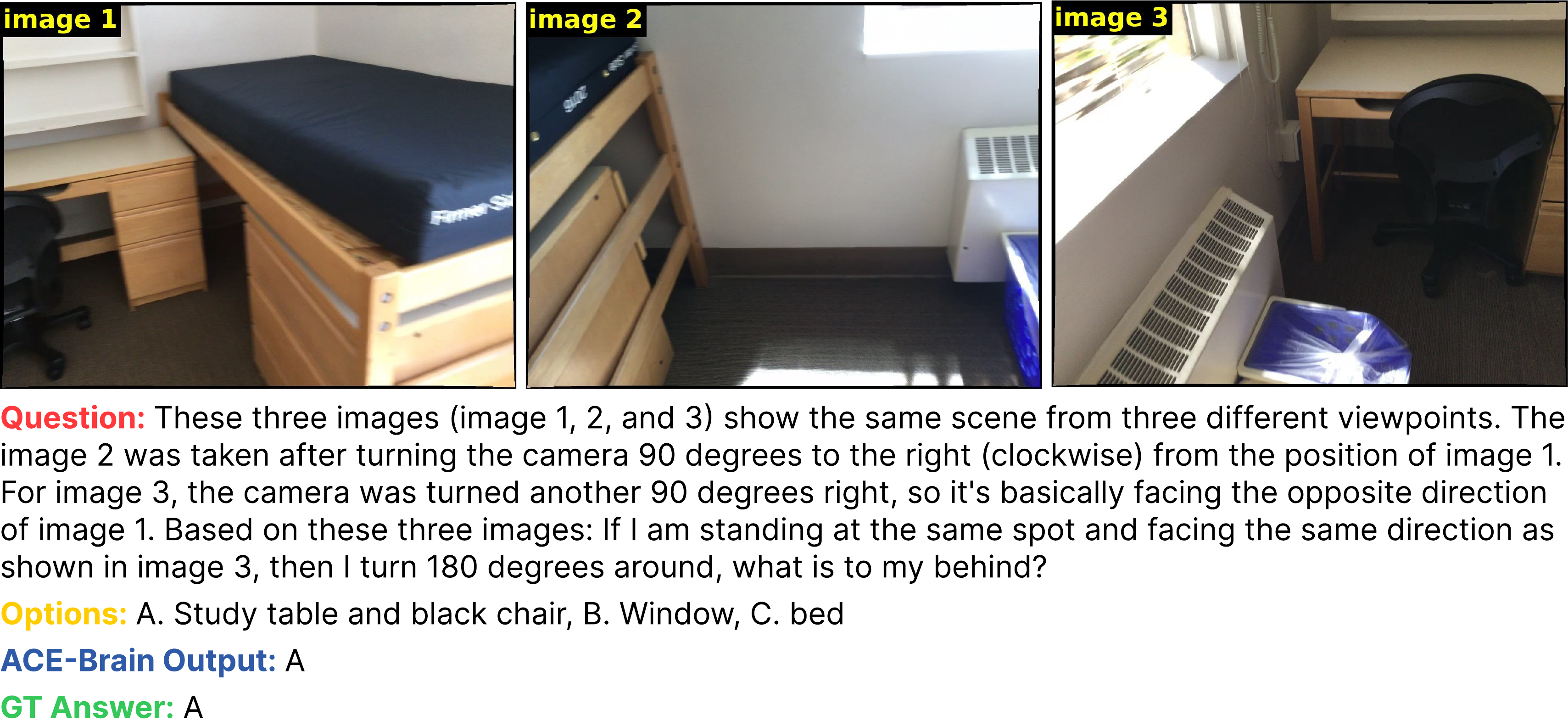}
    \caption{Example 3 of MindCube Benchmark.}
    \label{fig:minicube-3}
\end{figure}

\begin{figure}
    \centering
    \includegraphics[width=\linewidth]{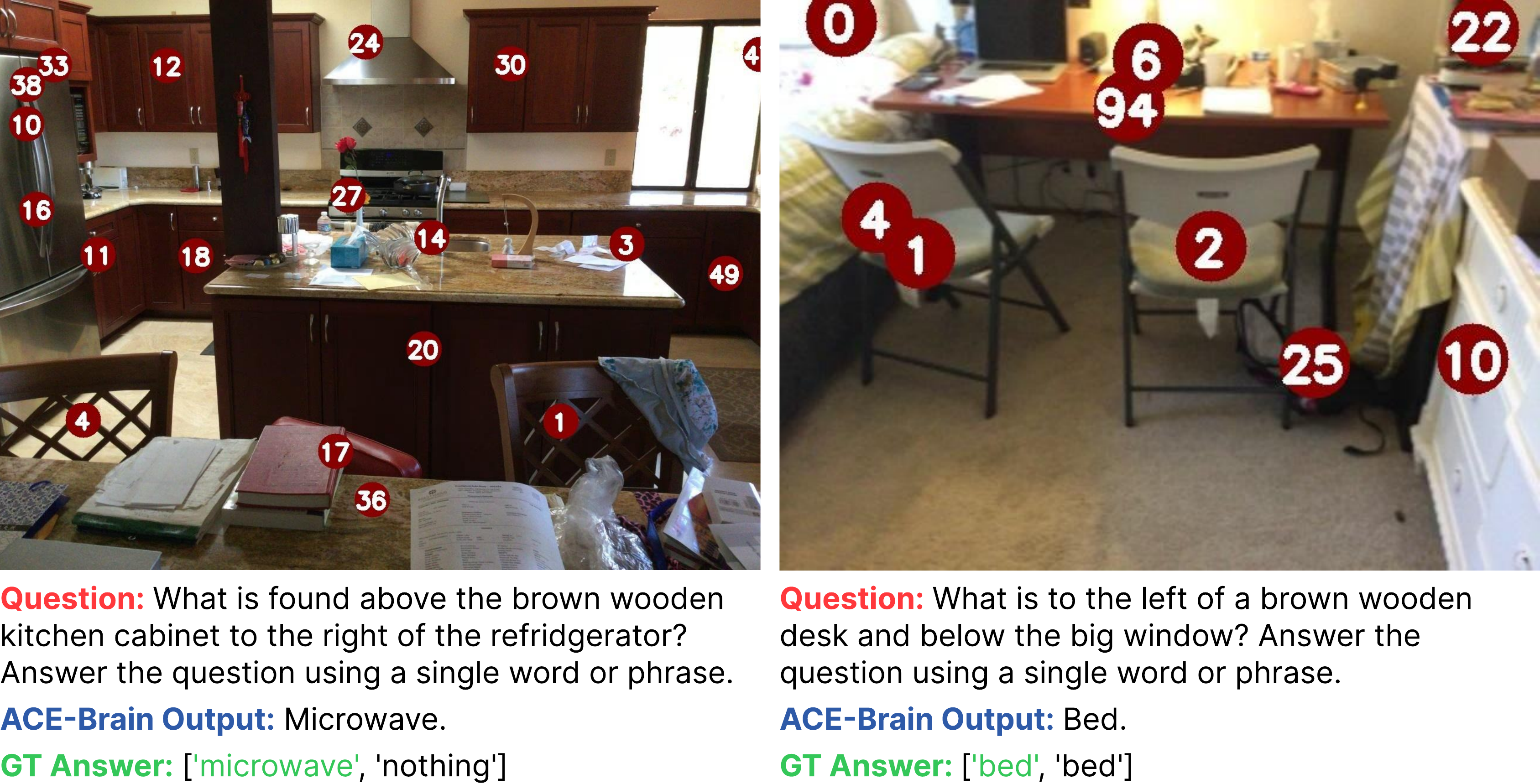}
    \caption{Examples 1, 2 of ScanQA Benchmark.}
    \label{fig:scanqa-12}
\end{figure}
\begin{figure}
    \centering
    \includegraphics[width=\linewidth]{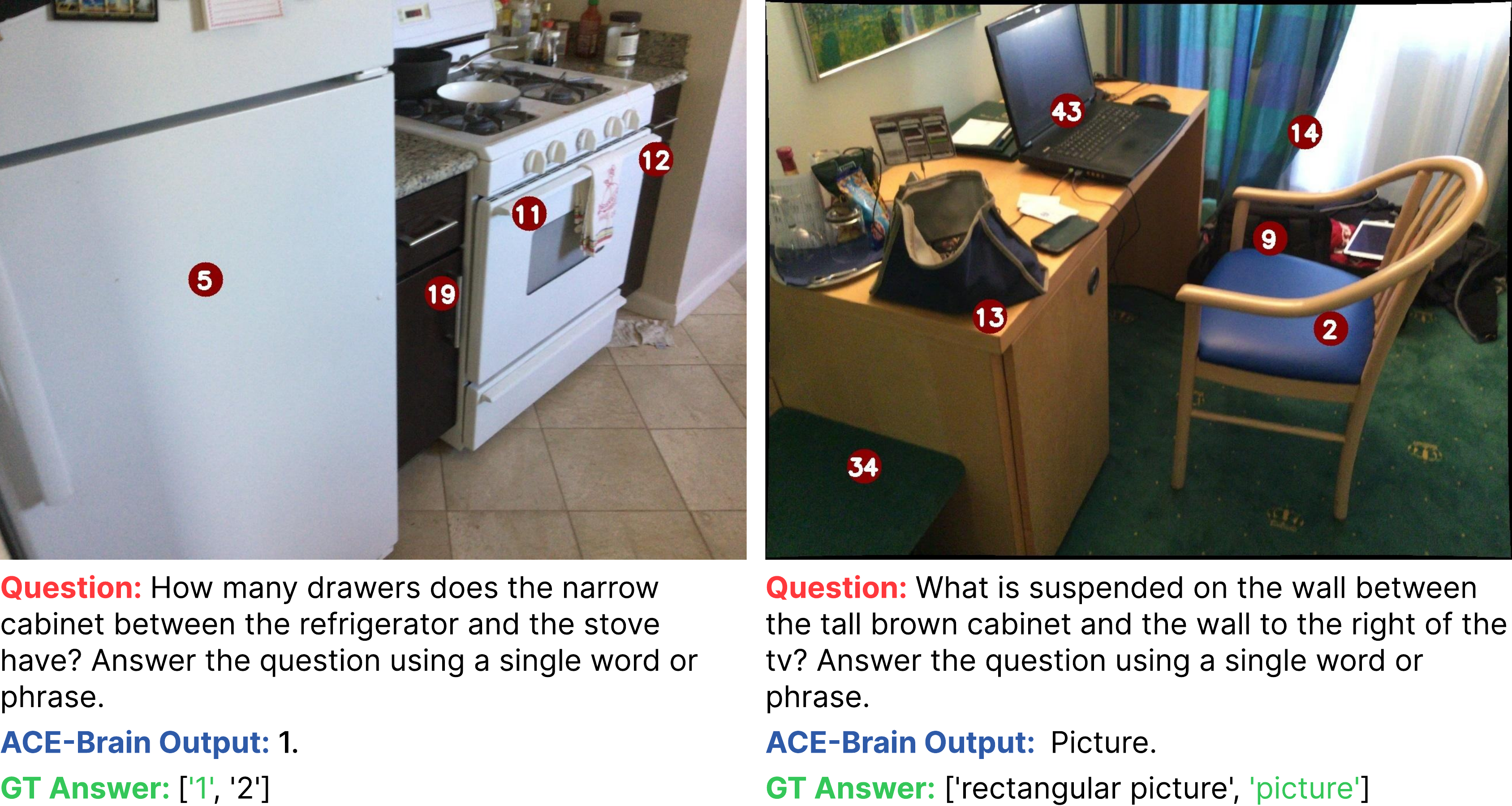}
    \caption{Examples 3, 4 of ScanQA Benchmark.}
    \label{fig:scanqa-34}
\end{figure}

\begin{figure}
    \centering
    \includegraphics[width=\linewidth]{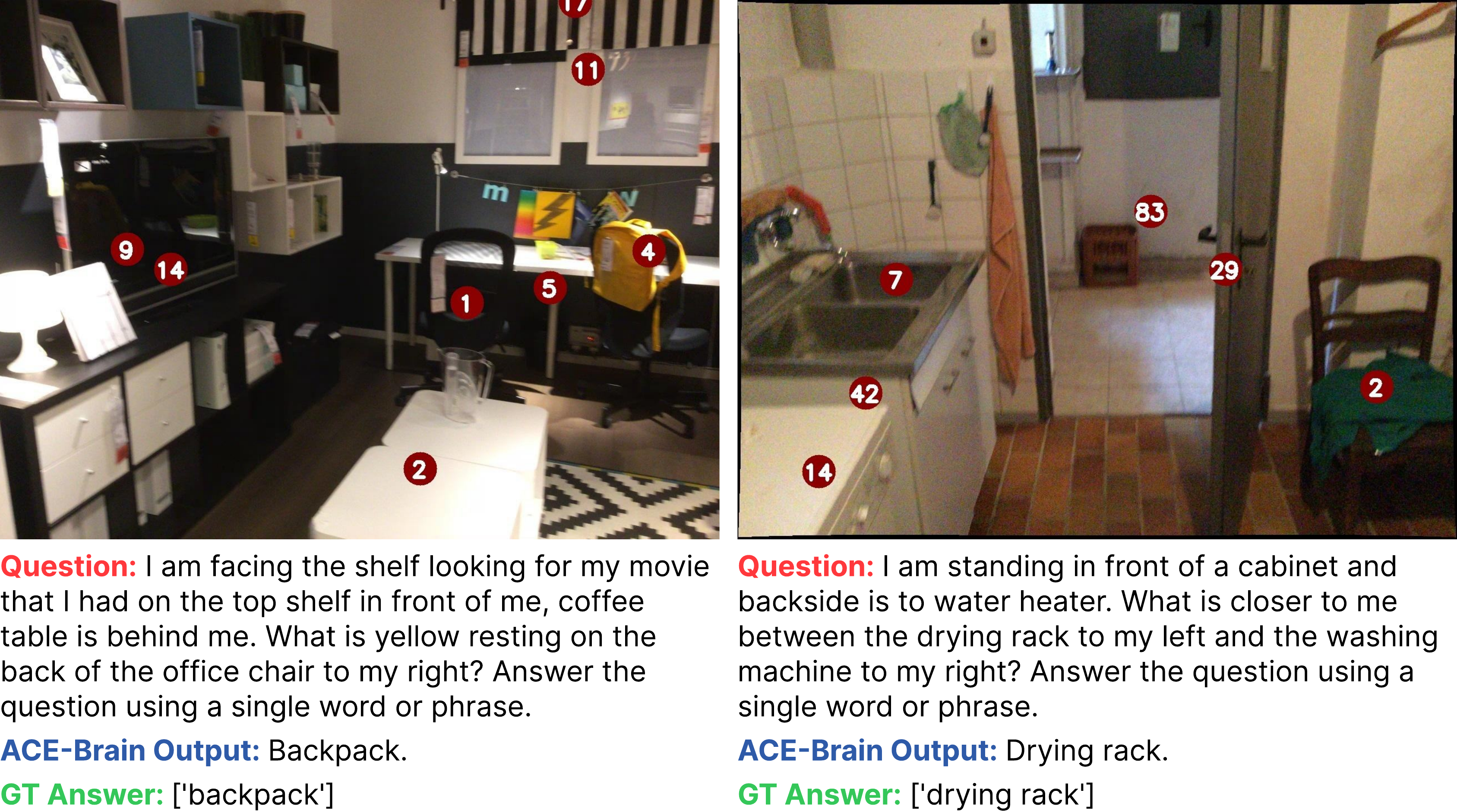}
    \caption{Examples 1, 2 of SQA3D Benchmark.}
    \label{fig:sqa3d-12}
\end{figure}
\begin{figure}
    \centering
    \includegraphics[width=\linewidth]{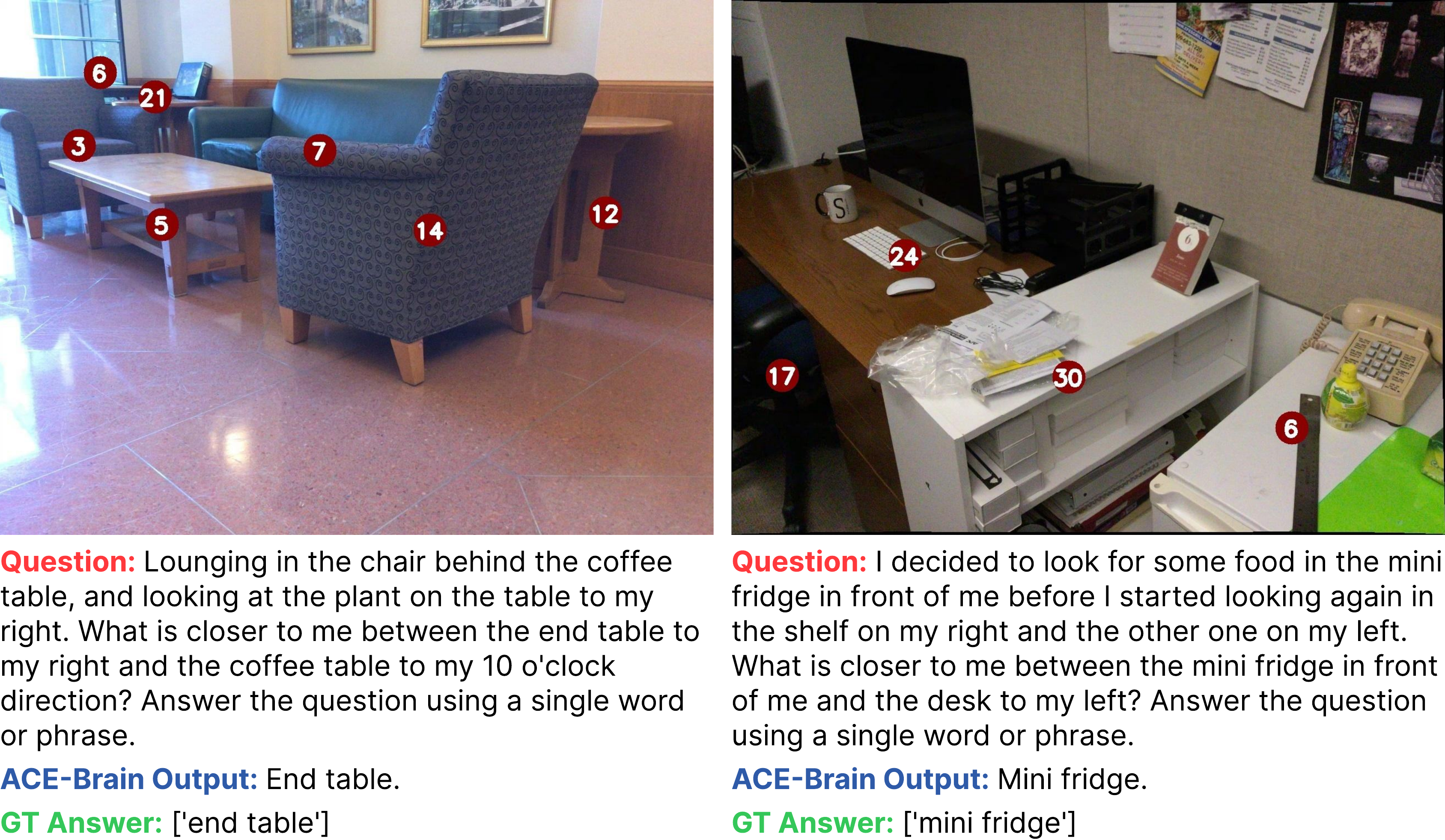}
    \caption{Examples 3, 4 of SQA3D Benchmark.}
    \label{fig:sqa3d-34}
\end{figure}

\begin{figure}
    \centering
    \includegraphics[width=\linewidth]{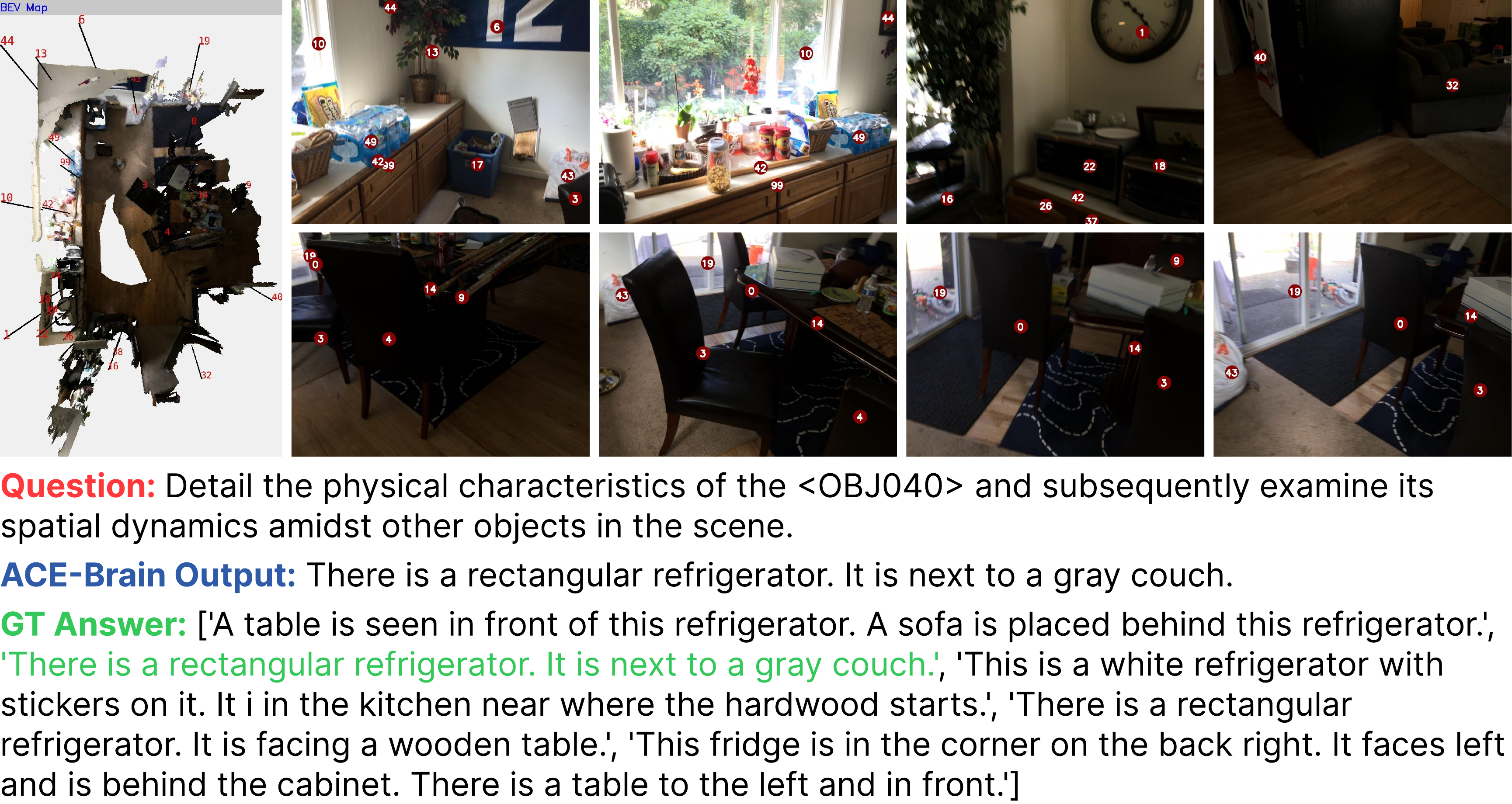}
    \caption{Example 1 of Scan2Cap Benchmark.}
    \label{fig:scan2cap1}
\end{figure}
\begin{figure}
    \centering
    \includegraphics[width=\linewidth]{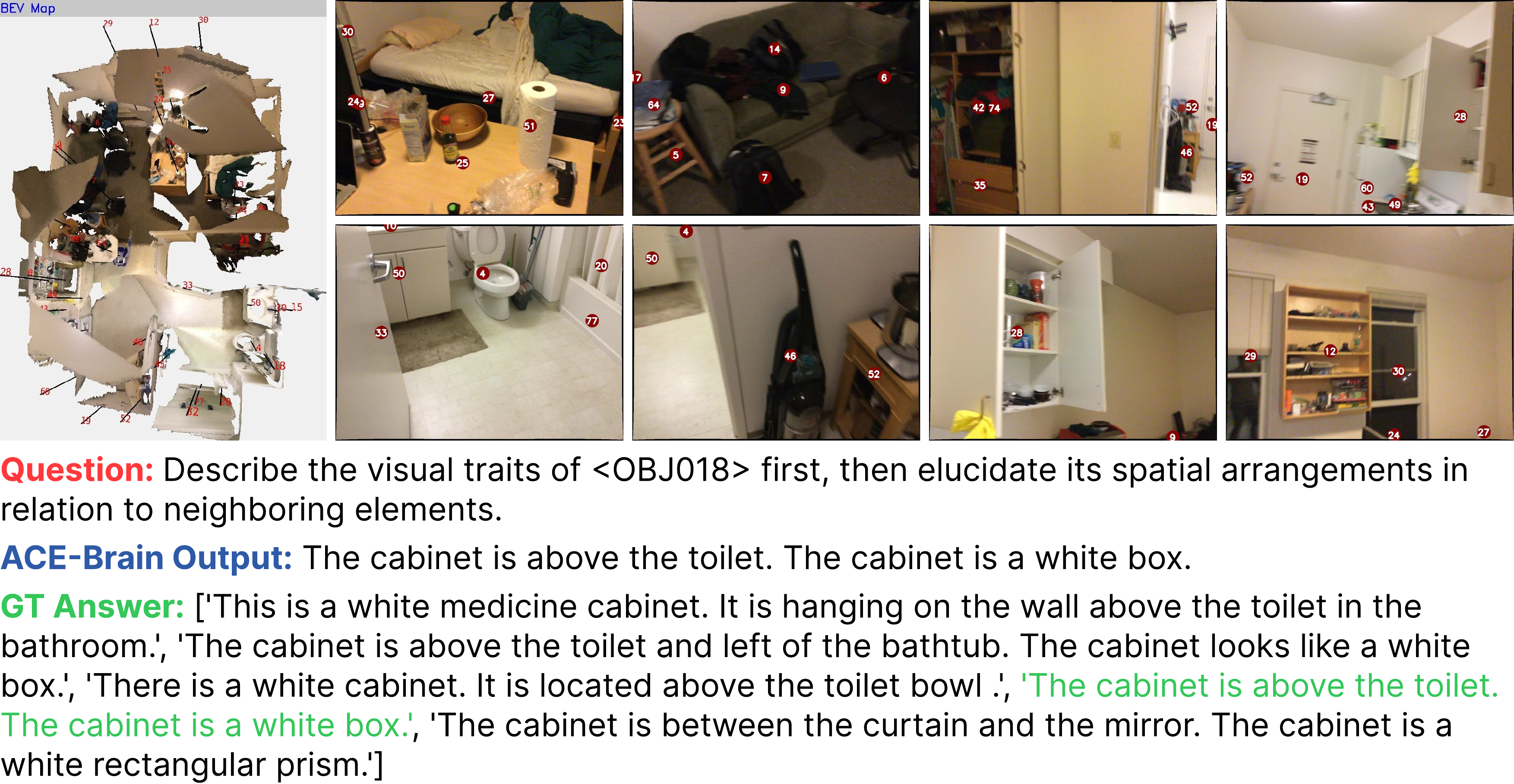}
    \caption{Example 2 of Scan2Cap Benchmark.}
    \label{fig:scan2cap2}
\end{figure}

\begin{figure}
    \centering
    \includegraphics[width=\linewidth]{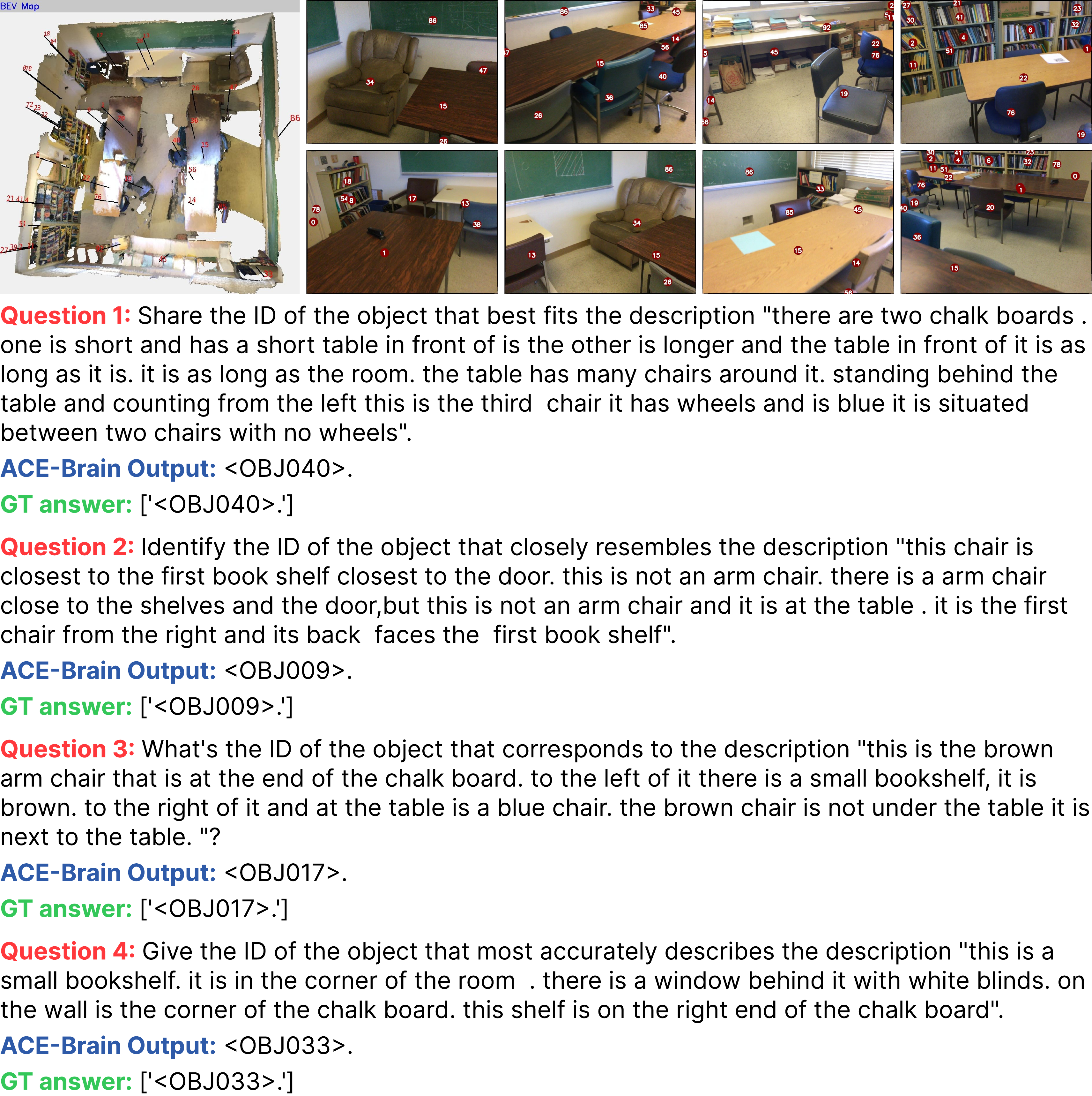}
    \caption{Example 1 of ScanRefer Benchmark.}
    \label{fig:scanrefer1}
\end{figure}
\begin{figure}
    \centering
    \includegraphics[width=\linewidth]{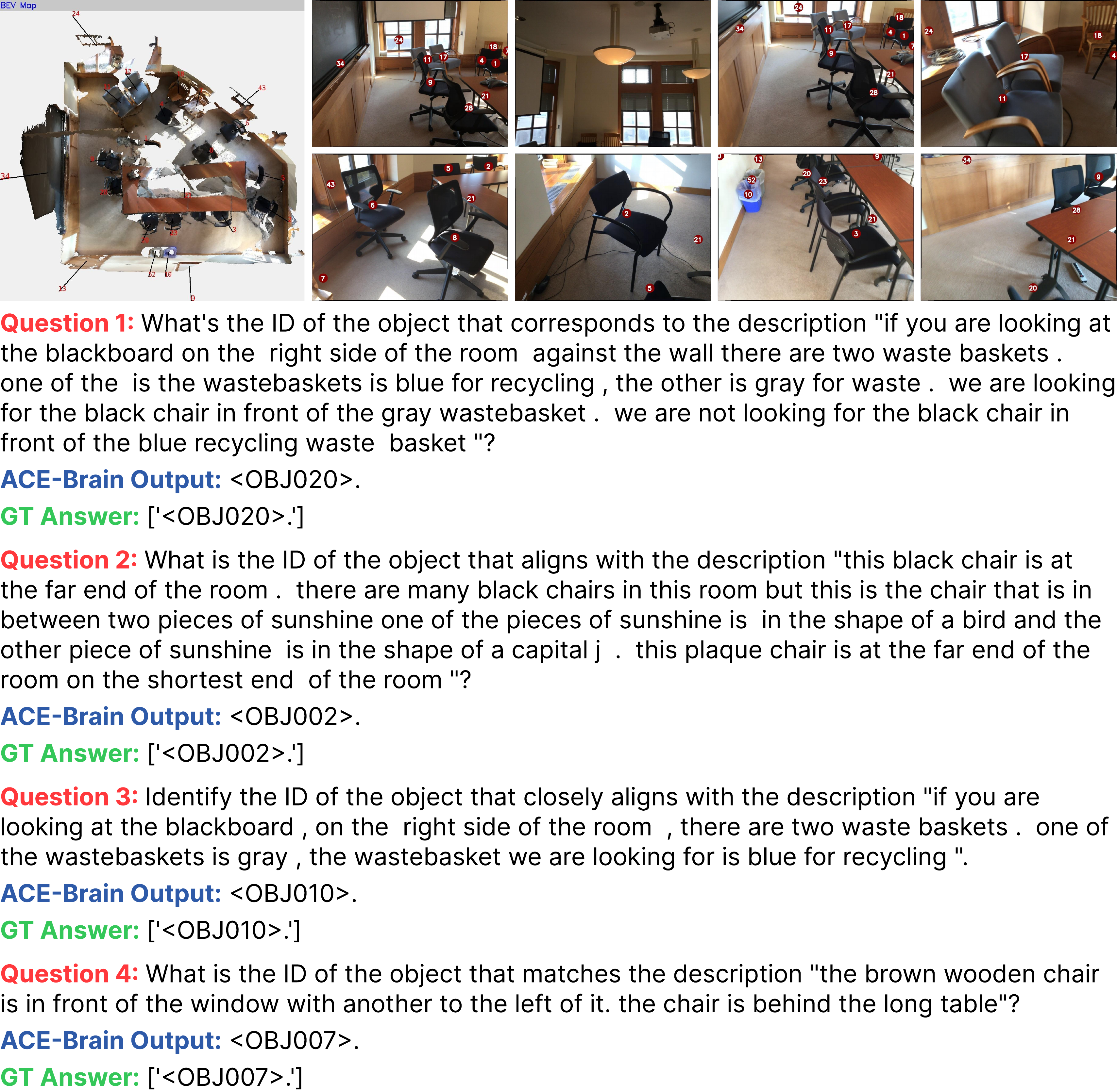}
    \caption{Example 2 of ScanRefer Benchmark.}
    \label{fig:scanrefer2}
\end{figure}

\begin{figure}
    \centering
    \includegraphics[width=\linewidth]{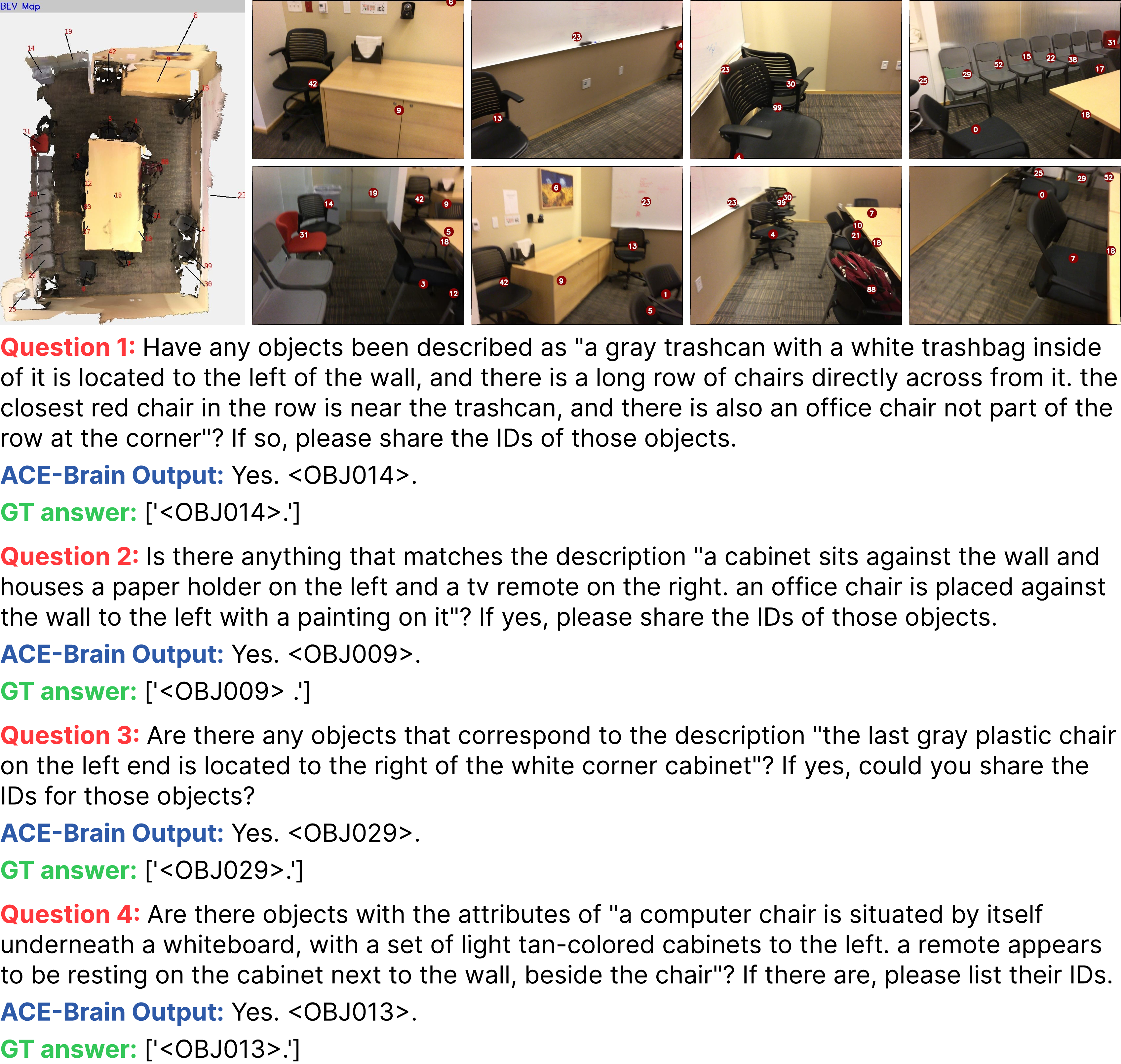}
    \caption{Example 1 of Multi3DRef Benchmark.}
    \label{fig:multi3dref1}
\end{figure}
\begin{figure}
    \centering
    \includegraphics[width=\linewidth]{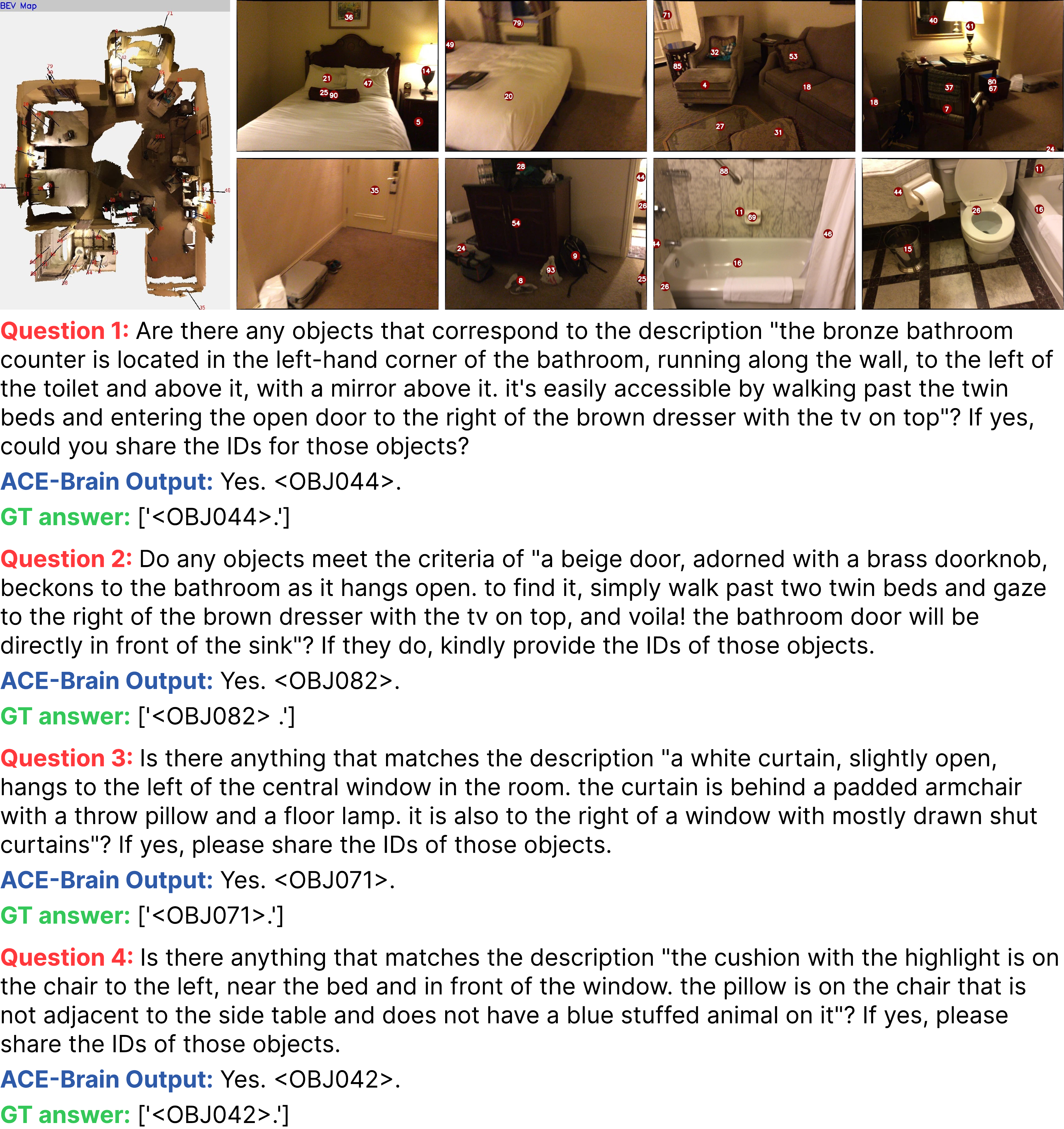}
    \caption{Example 2 of Multi3DRef Benchmark.}
    \label{fig:multi3dref2}
\end{figure}

\begin{figure}
    \centering
    \includegraphics[width=\linewidth]{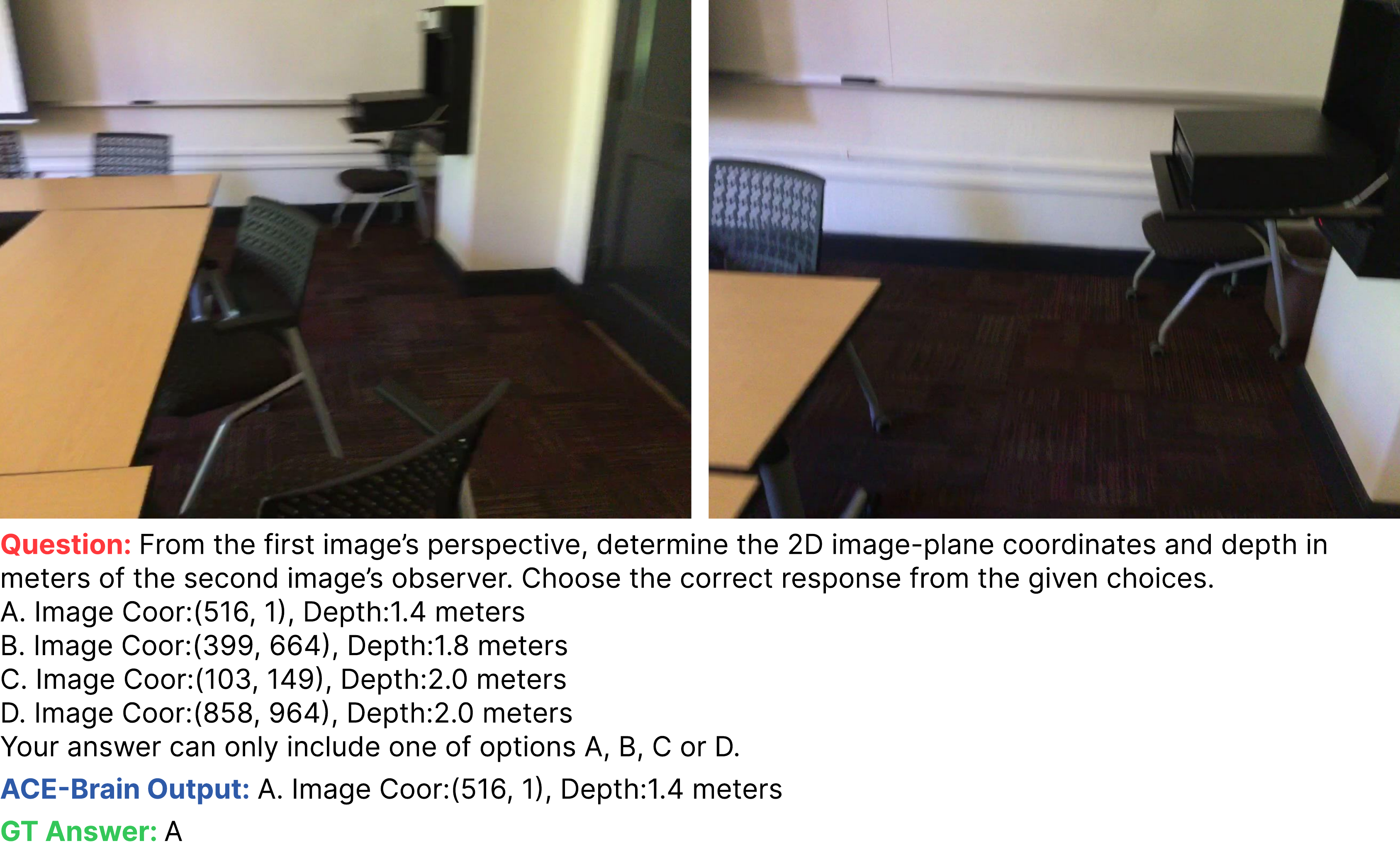}
    \caption{Example 1 of SparBench Benchmark.}
    \label{fig:sparbench-1}
\end{figure}
\begin{figure}
    \centering
    \includegraphics[width=\linewidth]{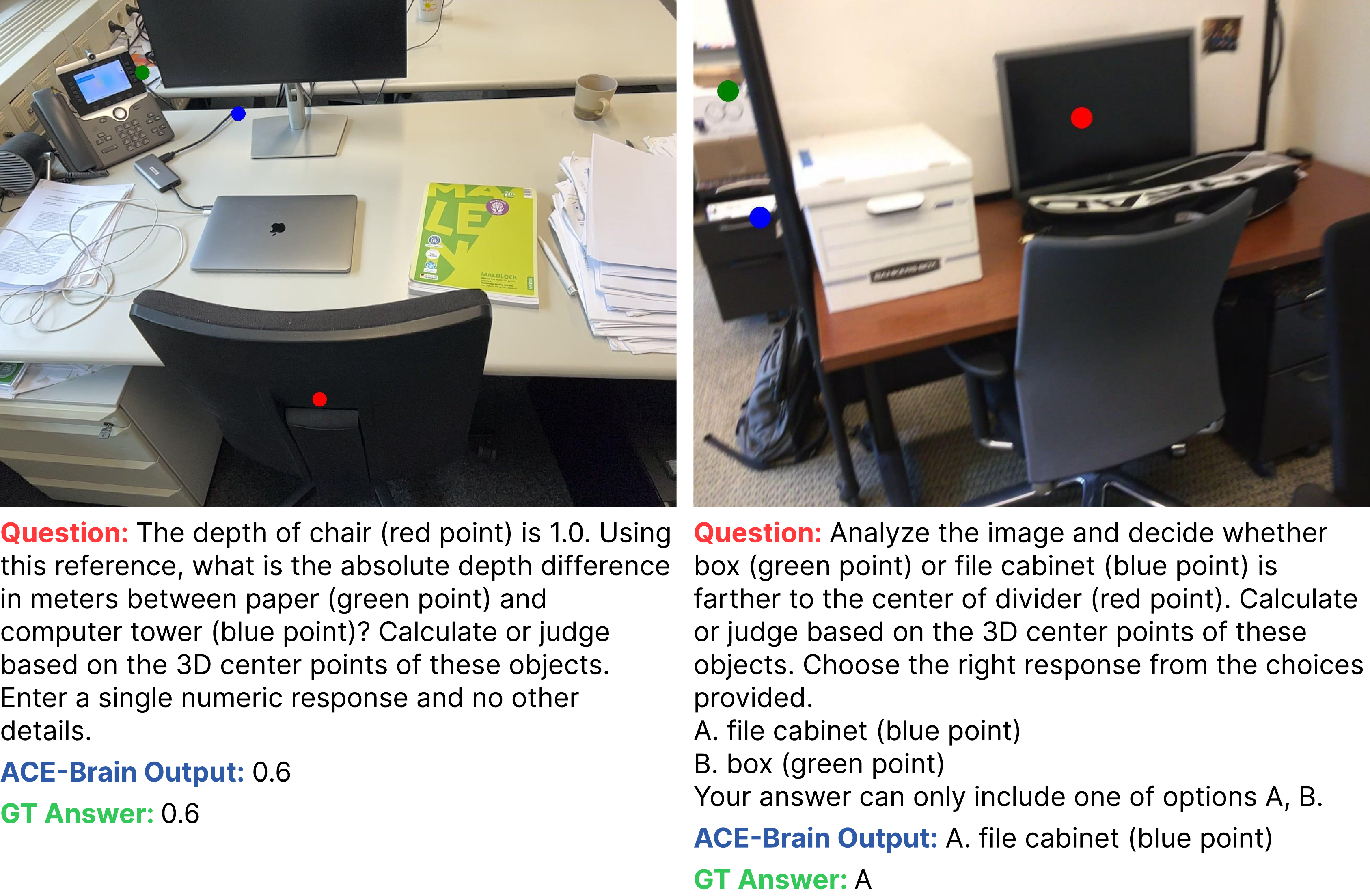}
    \caption{Example 2 of SparBench Benchmark.}
    \label{fig:sparbench-2}
\end{figure}
\begin{figure}
    \centering
    \includegraphics[width=\linewidth]{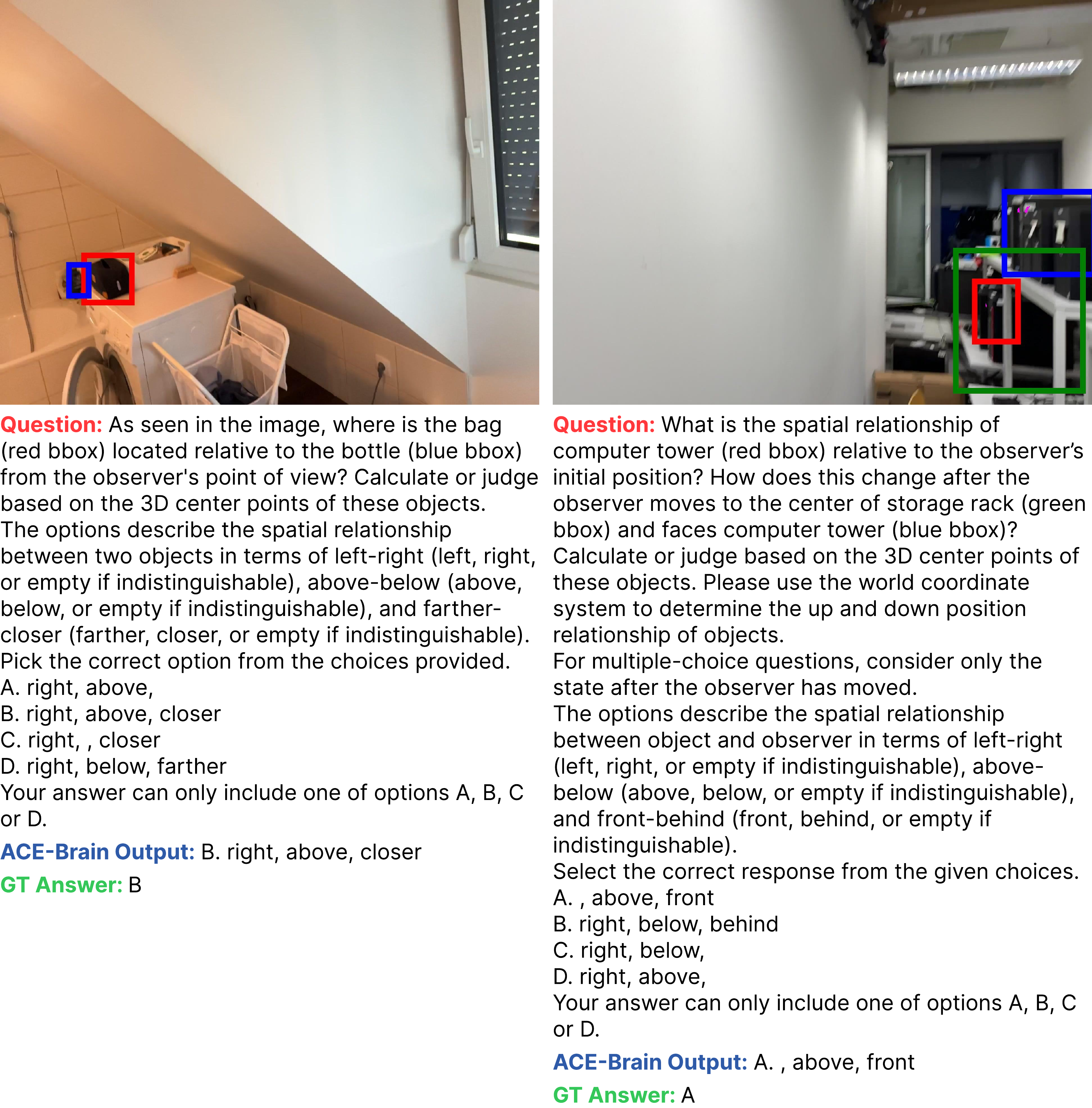}
    \caption{Example 3 of SparBench.}
    \label{fig:sparbench-3}
\end{figure}

\begin{figure}
    \centering
    \includegraphics[width=\linewidth]{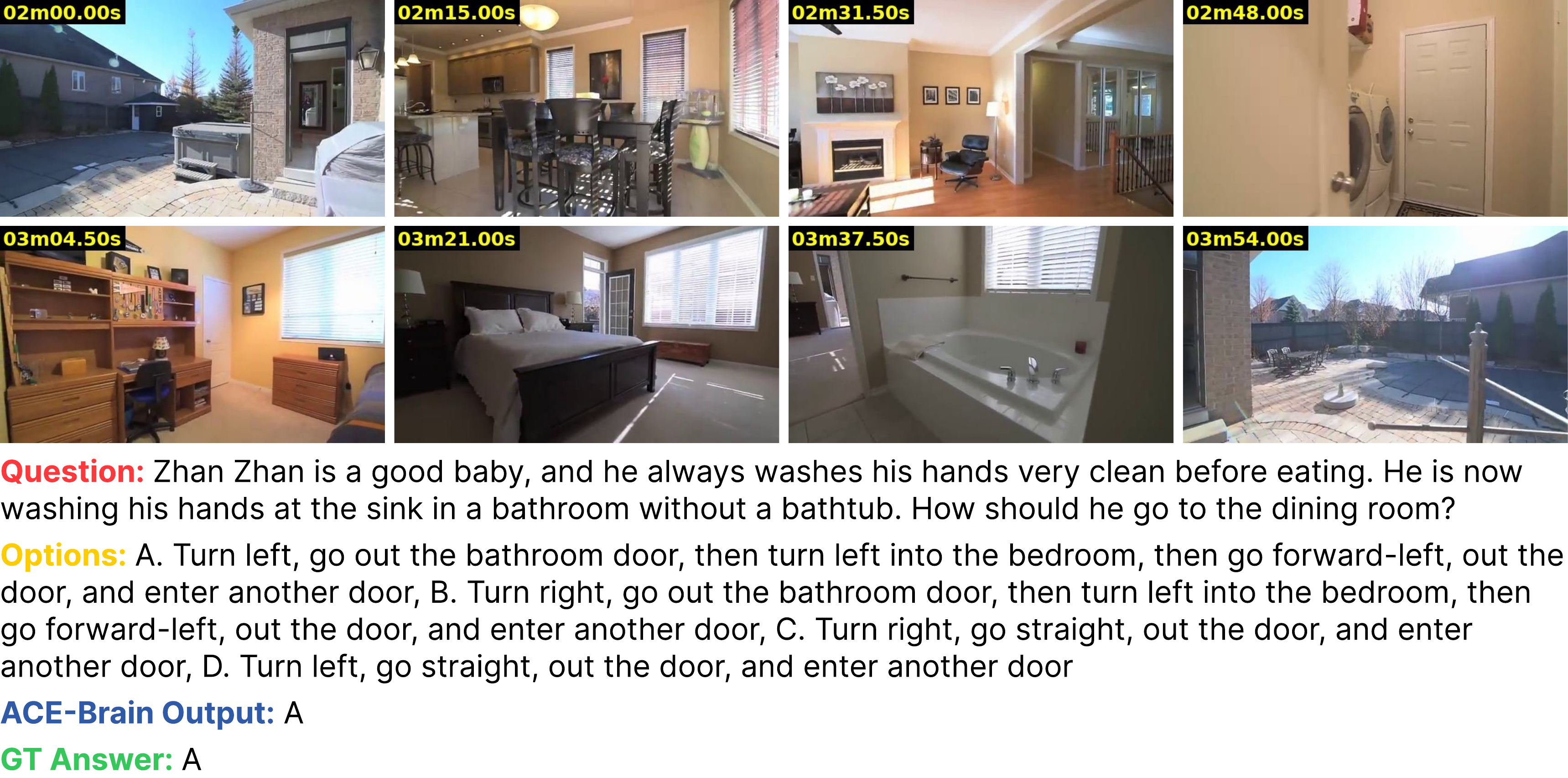}
    \caption{Example 1 of MMSIVideo Benchmark.}
    \label{fig:mmsivideo-1}
\end{figure}
\begin{figure}
    \centering
    \includegraphics[width=\linewidth]{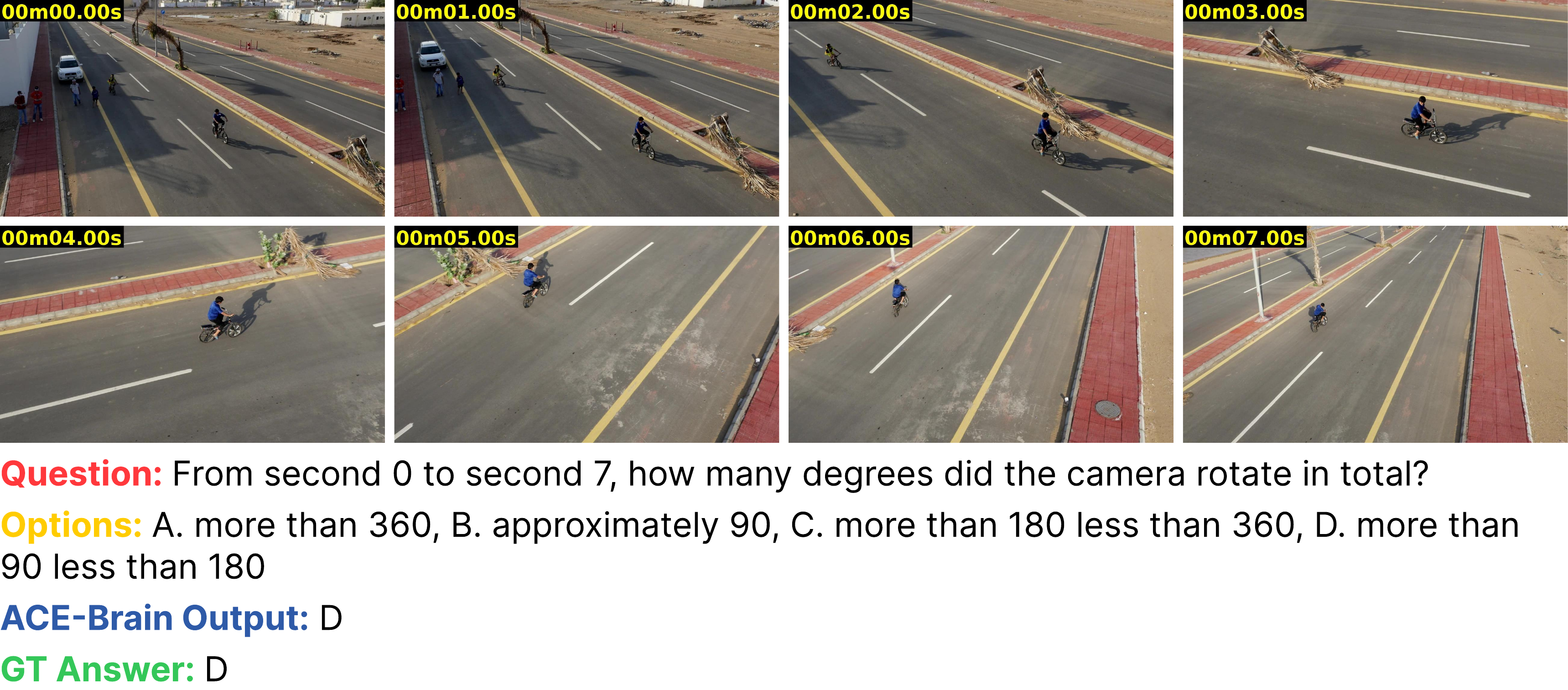}
    \caption{Example 2 of MMSIVideo Benchmark.}
    \label{fig:mmsivideo-2}
\end{figure}

\begin{figure}
    \centering
    \includegraphics[width=\linewidth]{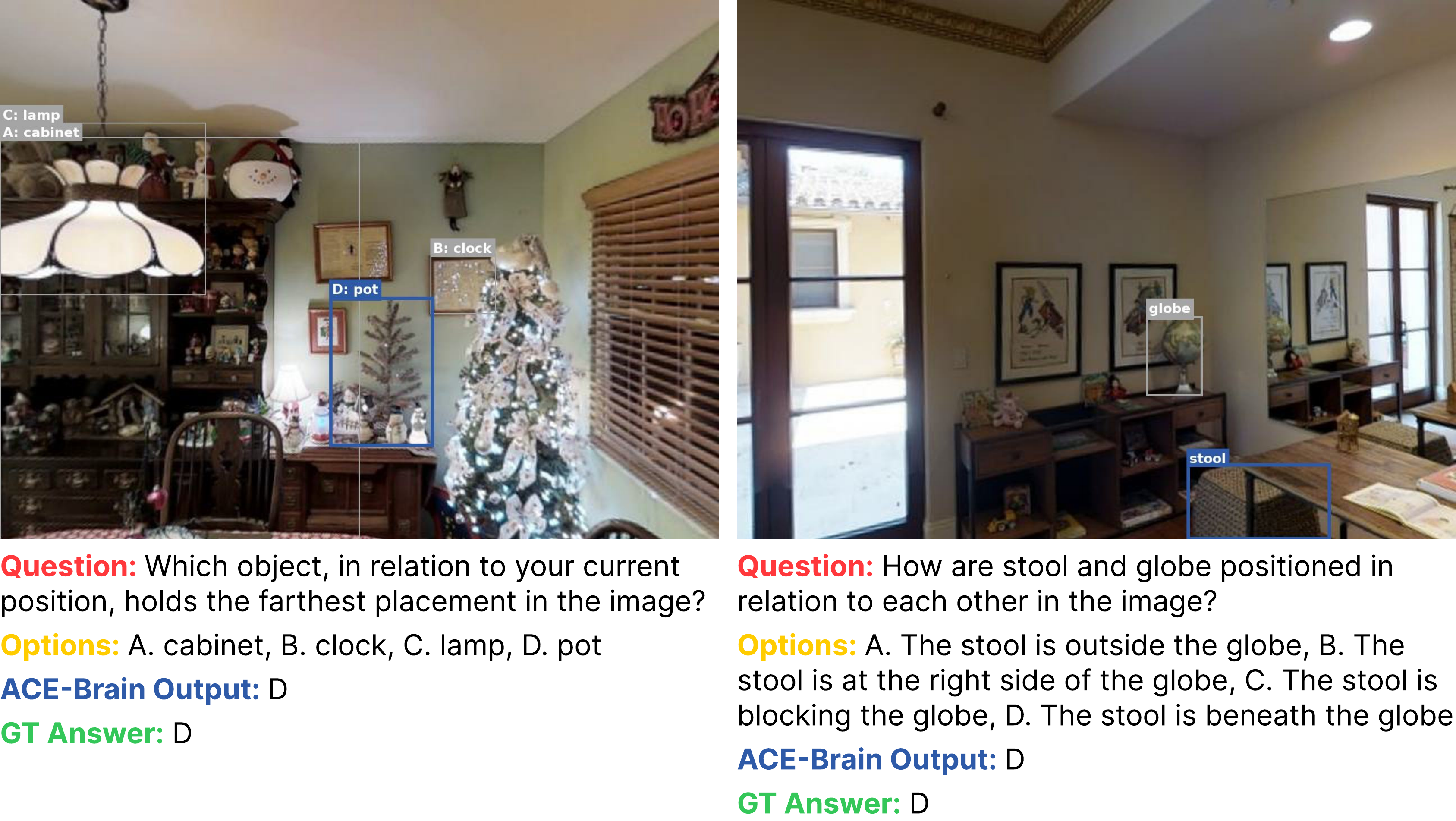}
    \caption{Examples 1, 2 of EmbSpatial Benchmark.}
    \label{fig:embspatial-12}
\end{figure}
\begin{figure}
    \centering
    \includegraphics[width=\linewidth]{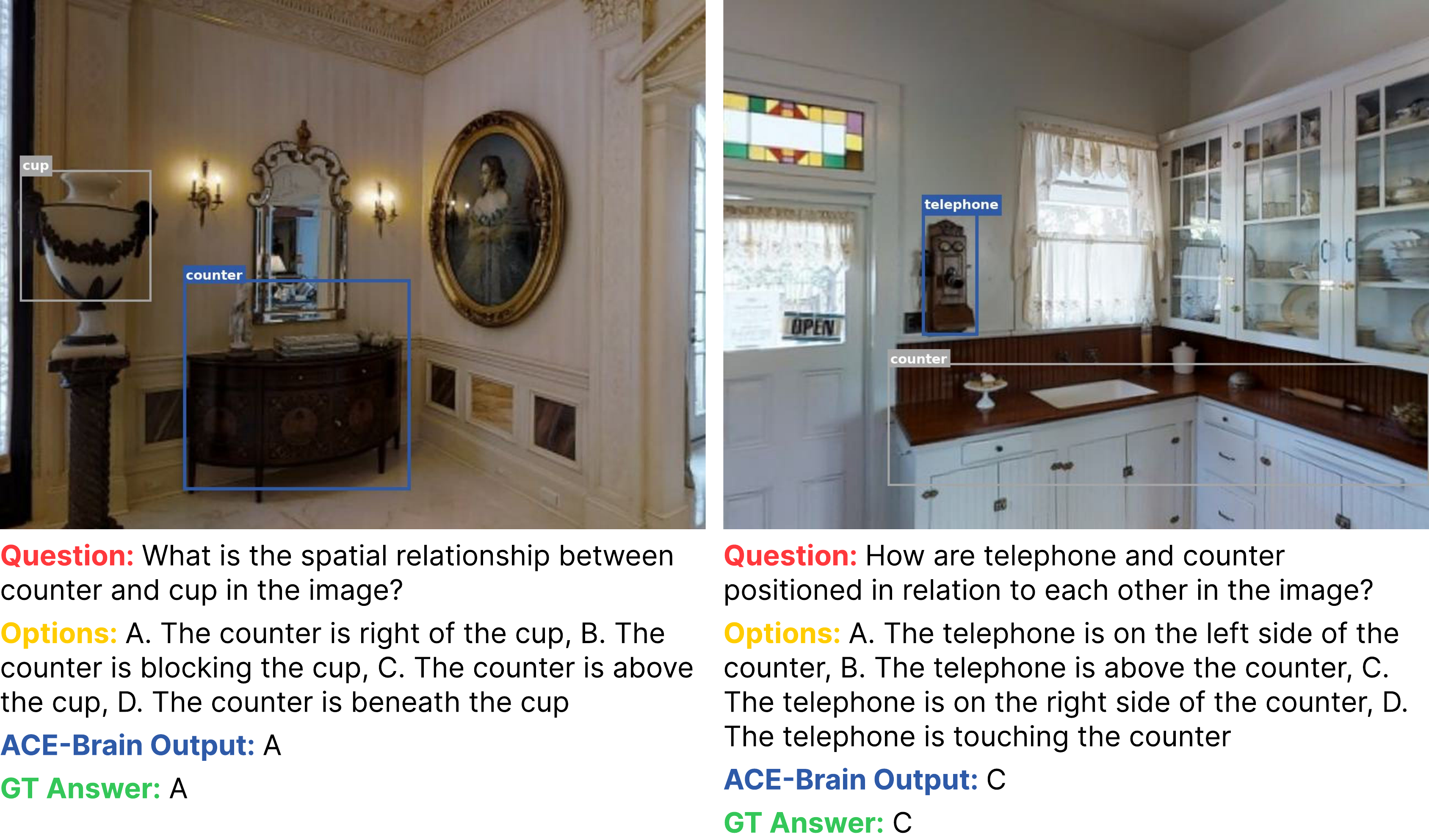}
    \caption{Examples 3, 4 of EmbSpatial Benchmark.}
    \label{fig:embspatial-34}
\end{figure}

\begin{figure}
    \centering
    \includegraphics[width=\linewidth]{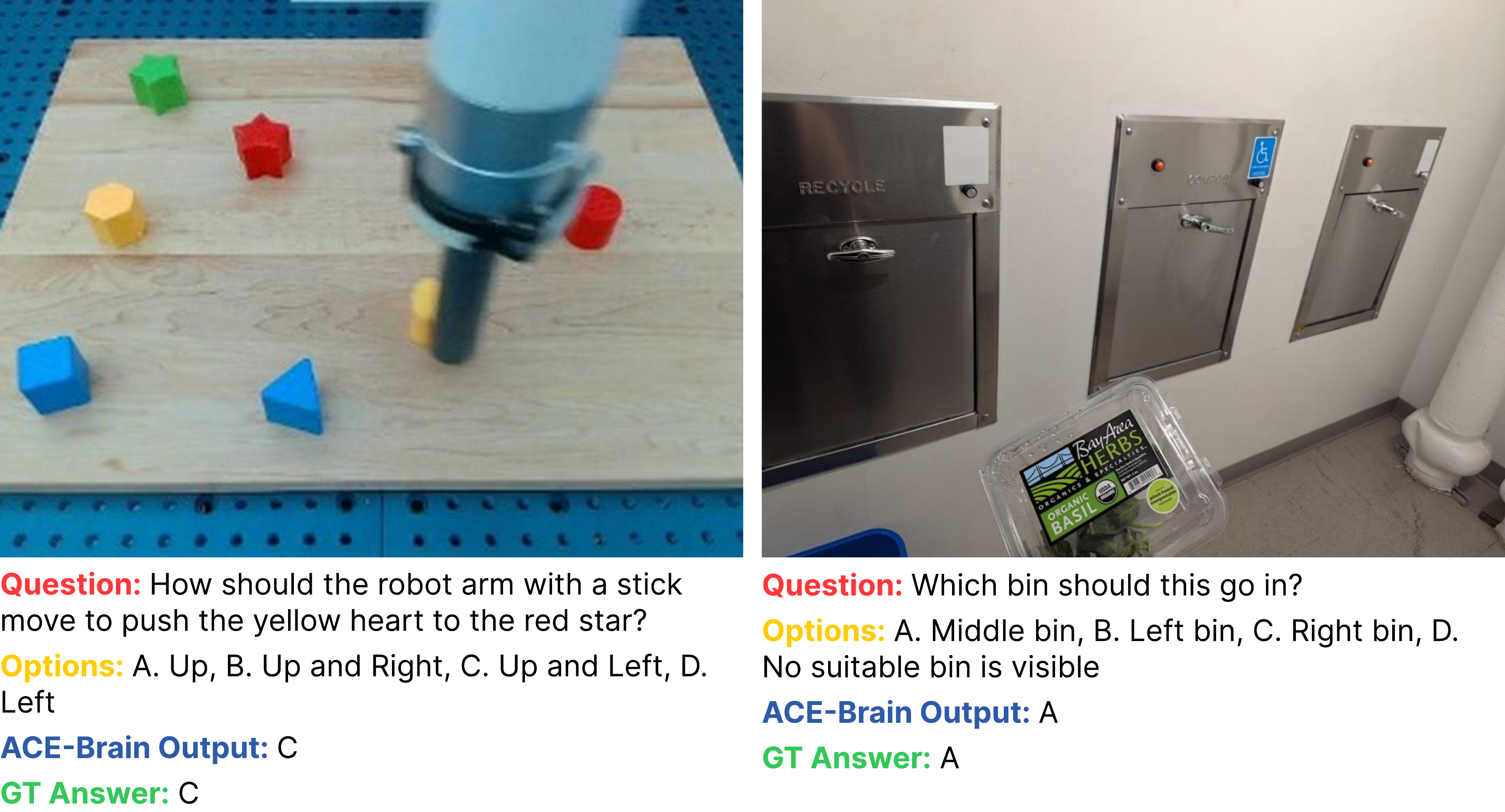}
    \caption{Examples 1, 2 of ERQA Benchmark.}
    \label{fig:erqa-12}
\end{figure}
\begin{figure}
    \centering
    \includegraphics[width=\linewidth]{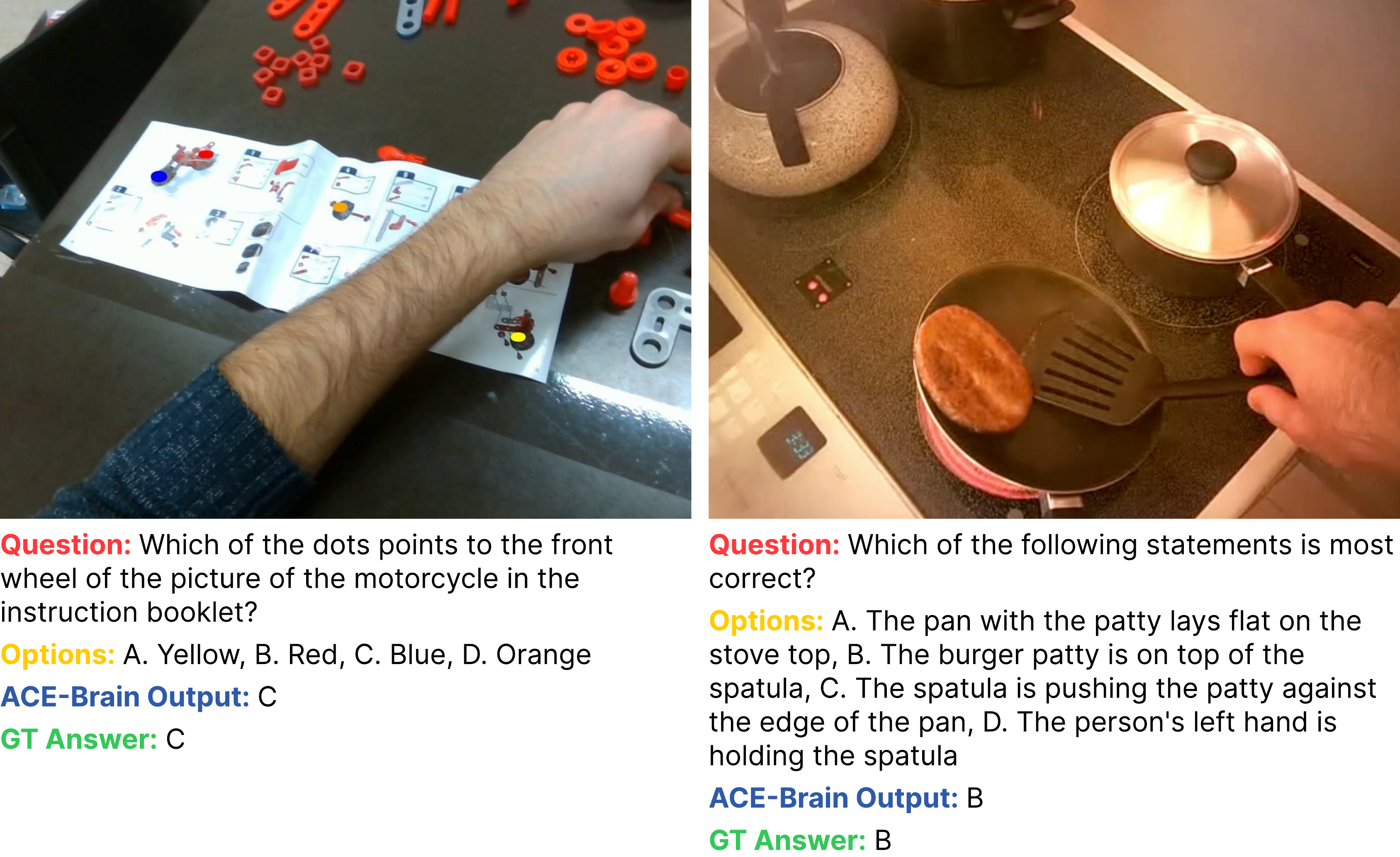}
    \caption{Examples 3, 4 of ERQA Benchmark.}
    \label{fig:erqa-34}
\end{figure}

\begin{figure}
    \centering
    \includegraphics[width=\linewidth]{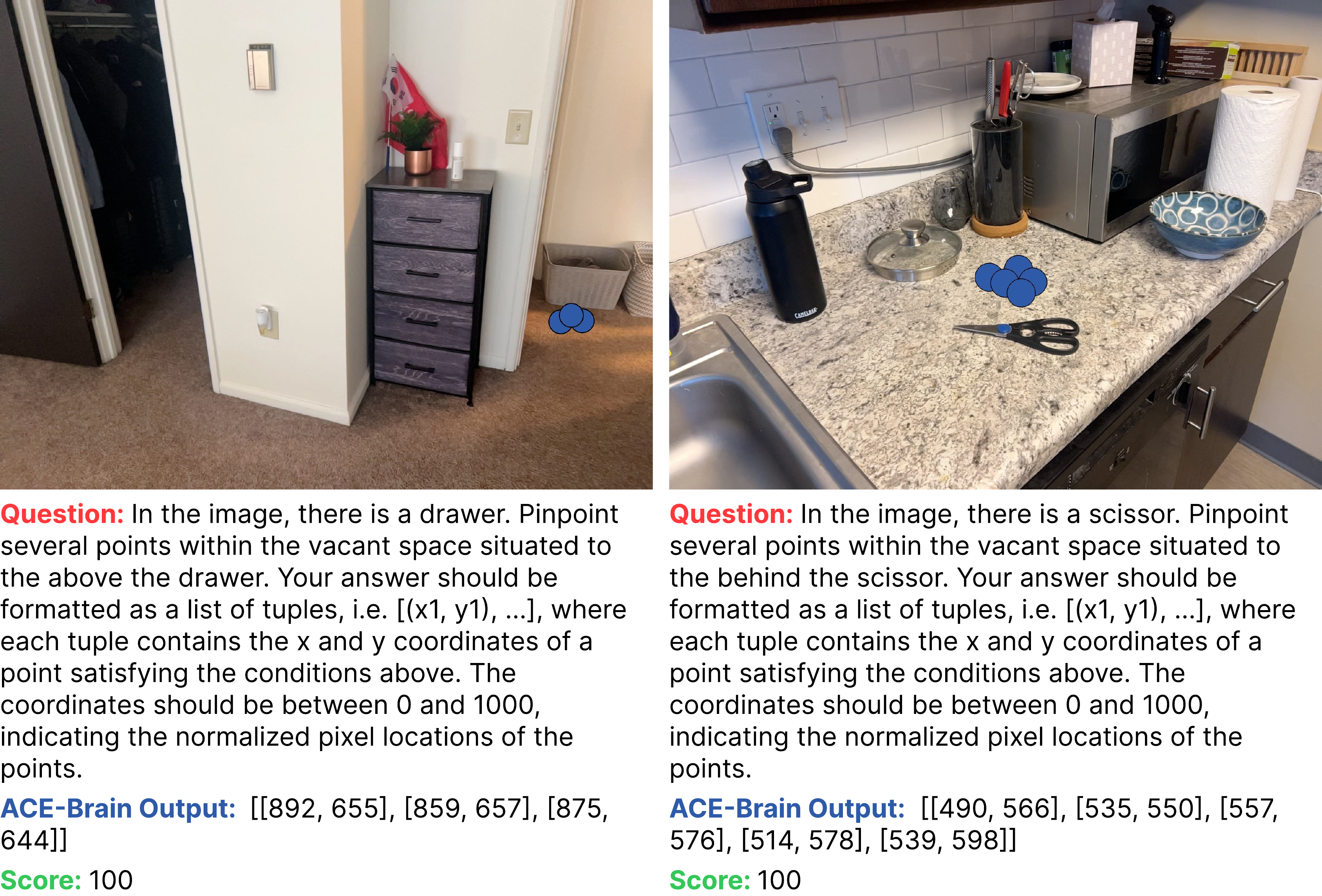}
    \caption{Examples 1, 2 of RoboSpatial Benchmark.}
    \label{fig:robospatial-12}
\end{figure}
\begin{figure}
    \centering
    \includegraphics[width=\linewidth]{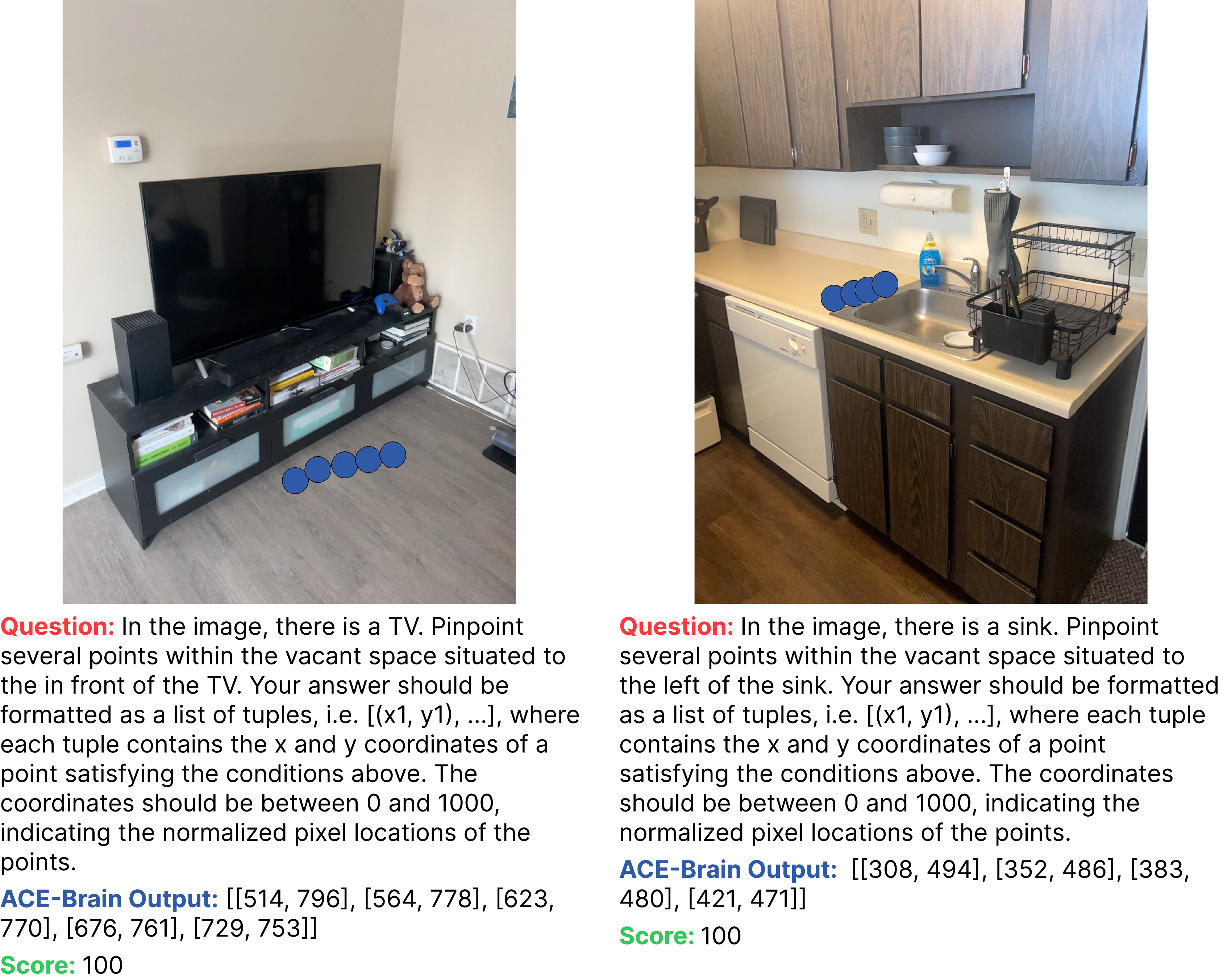}
    \caption{Examples 3, 4 of RoboSpatial Benchmark.}
    \label{fig:robospatial-34}
\end{figure}

\begin{figure}
    \centering
    \includegraphics[width=\linewidth]{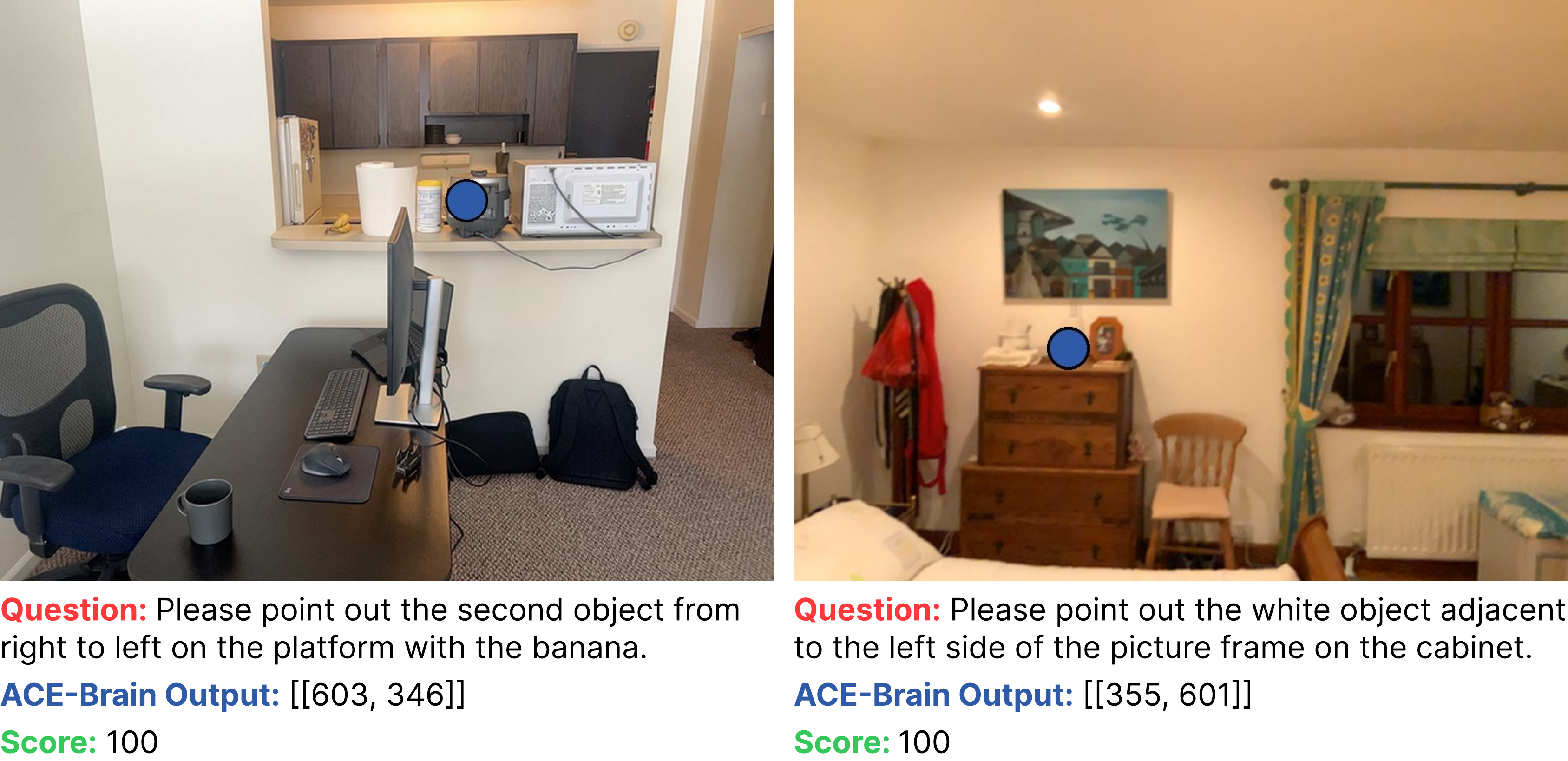}
    \caption{Examples 1, 2 of RefSpatial Benchmark.}
    \label{fig:refspatial-12}
\end{figure}
\begin{figure}
    \centering
    \includegraphics[width=\linewidth]{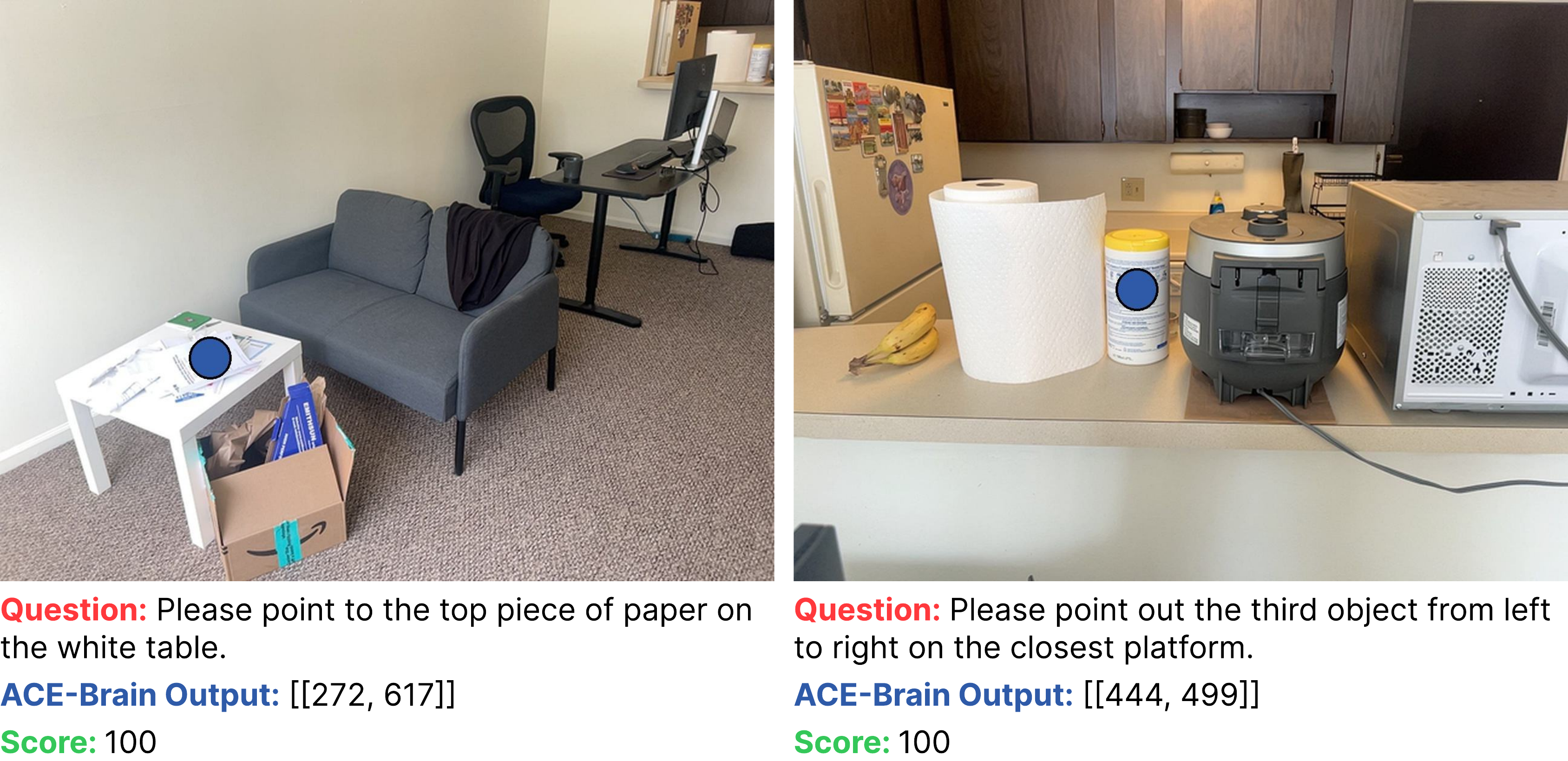}
    \caption{Examples 3, 4 of RefSpatial Benchmark.}
    \label{fig:refspatial-34}
\end{figure}

\begin{figure}
    \centering
    \includegraphics[width=\linewidth]{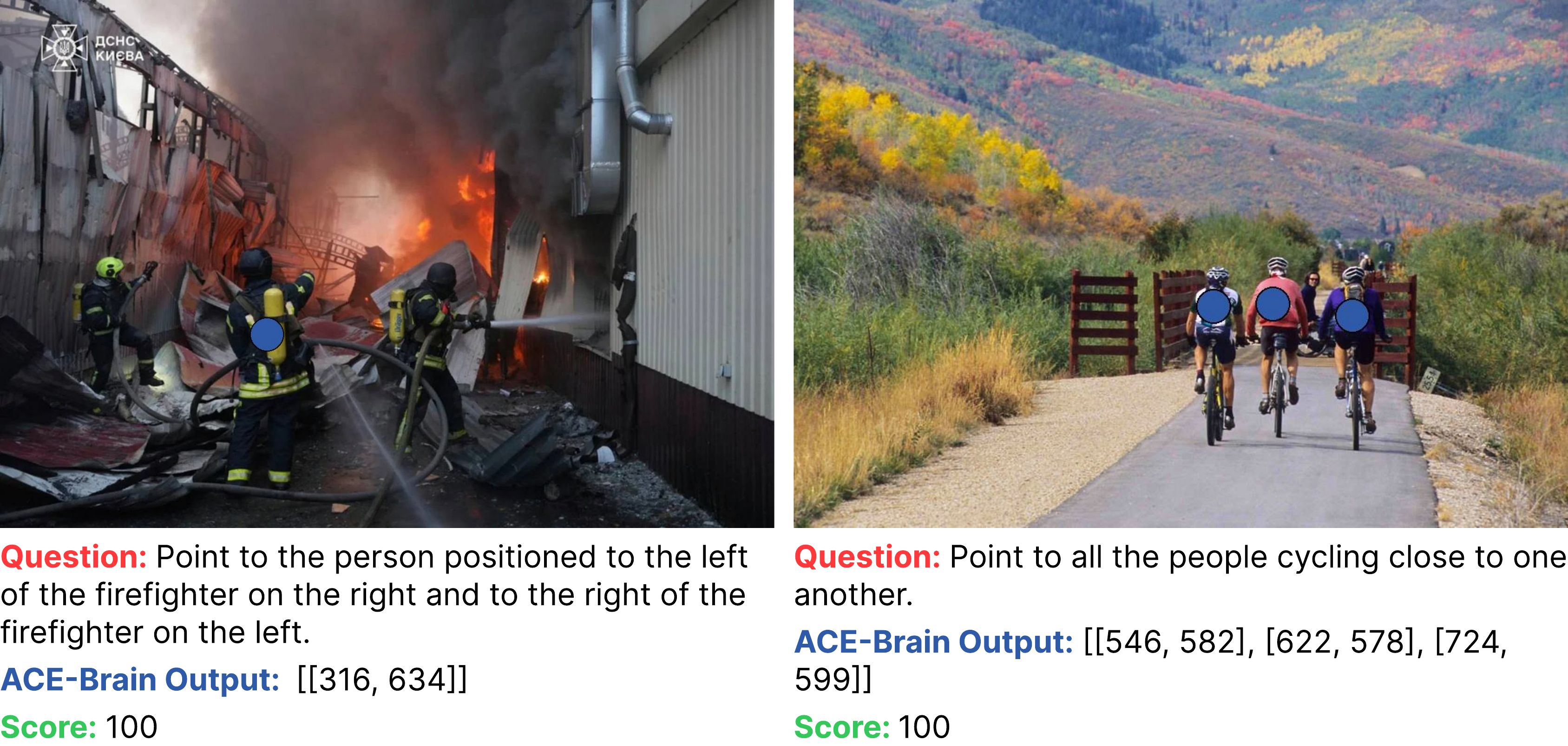}
    \caption{Examples 1, 2 of PointArena Benchmark.}
    \label{fig:pointarena-12}
\end{figure}
\begin{figure}
    \centering
    \includegraphics[width=\linewidth]{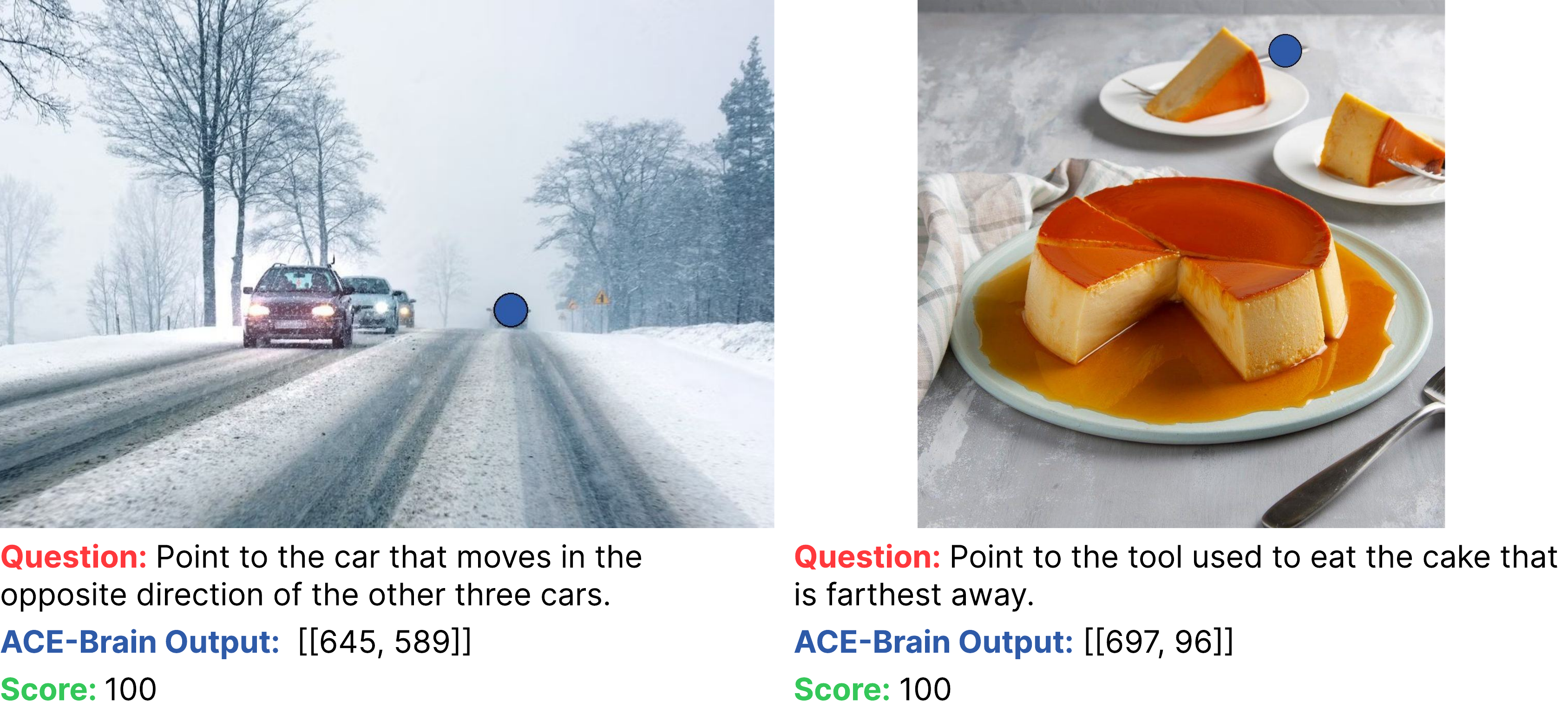}
    \caption{Examples 3, 4 of PointArena Benchmark.}
    \label{fig:pointarena-34}
\end{figure}

\begin{figure}
    \centering
    \includegraphics[width=\linewidth]{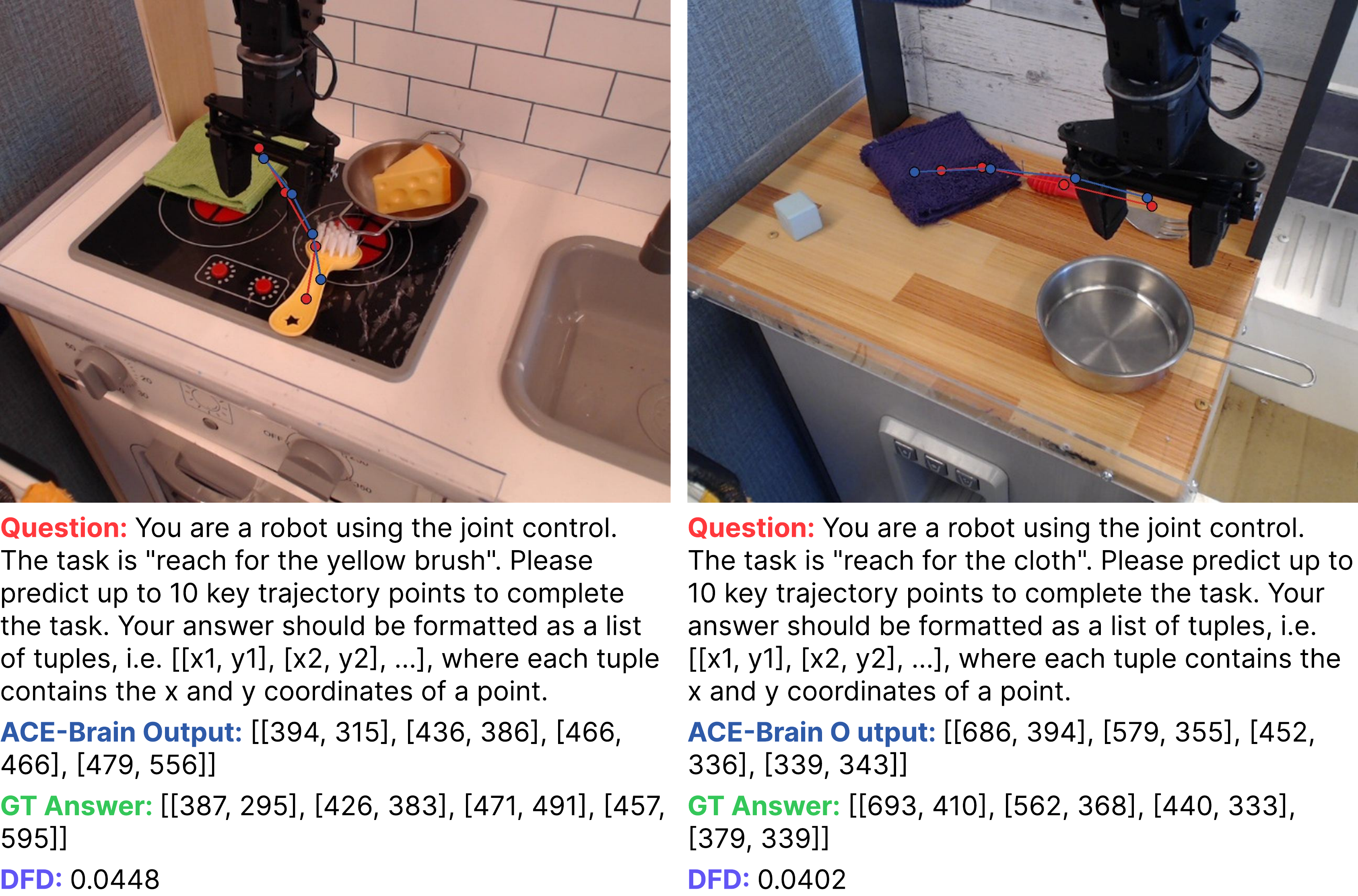}
    \caption{Examples 1, 2 of ShareRobot Benchmark.}
    \label{fig:sharerobot-12}
\end{figure}
\begin{figure}
    \centering
    \includegraphics[width=\linewidth]{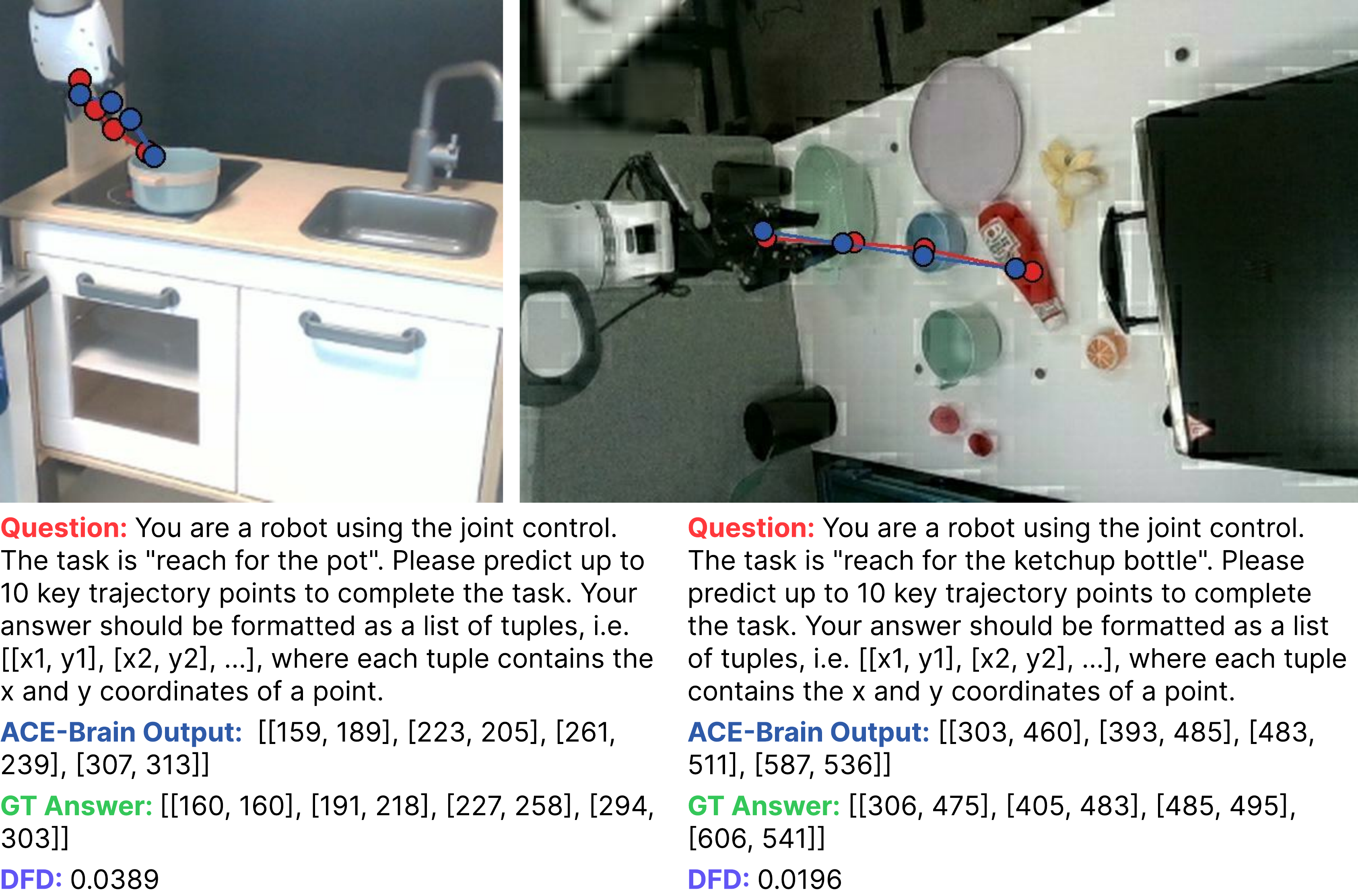}
    \caption{Examples 3, 4 of ShareRobot Benchmark.}
    \label{fig:sharerobot-34}
\end{figure}

\begin{figure}
    \centering
    \includegraphics[width=\linewidth]{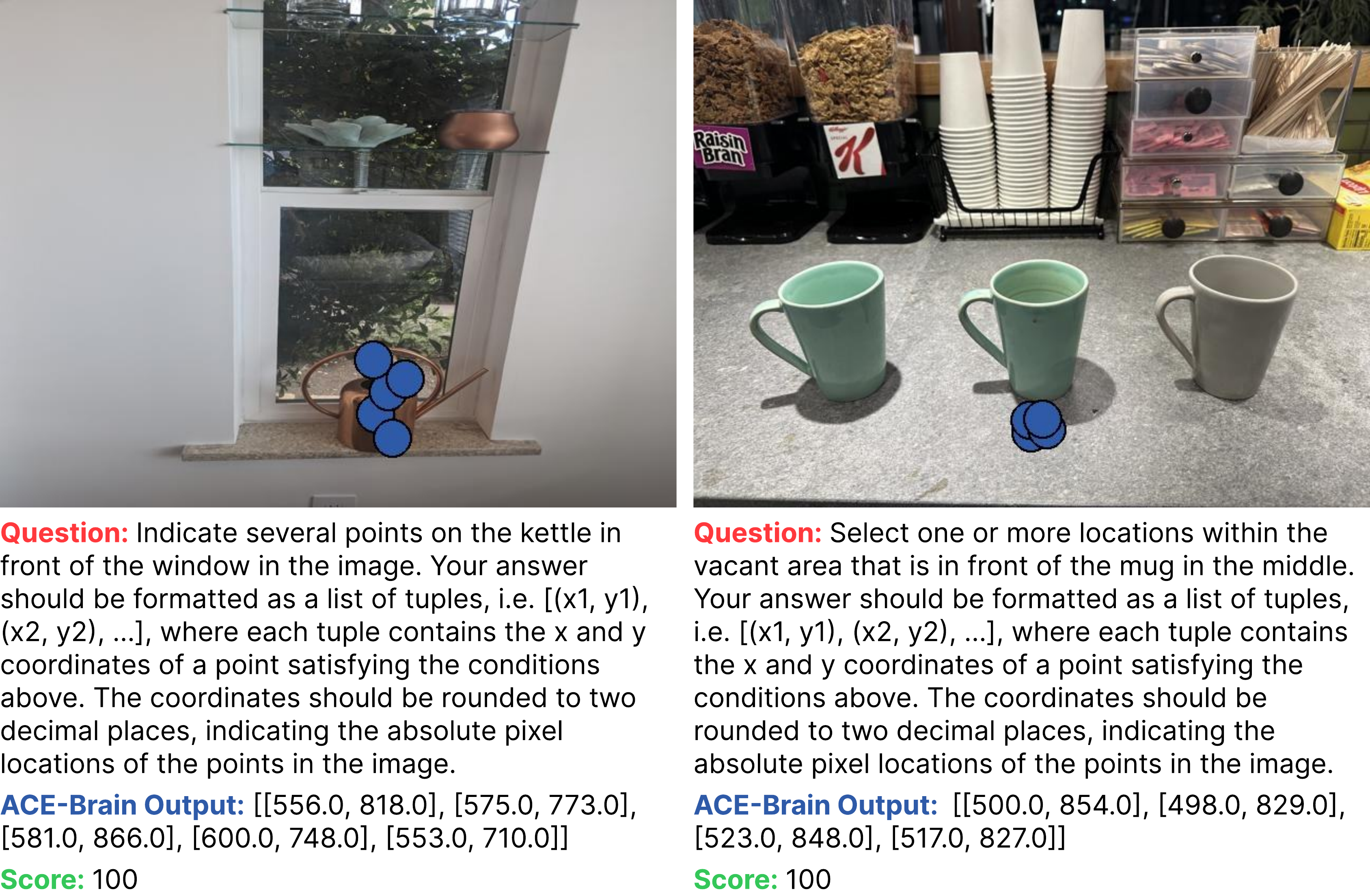}
    \caption{Examples 1, 2 of RoboAfford Benchmark.}
    \label{fig:roboafford-12}
\end{figure}
\begin{figure}
    \centering
    \includegraphics[width=\linewidth]{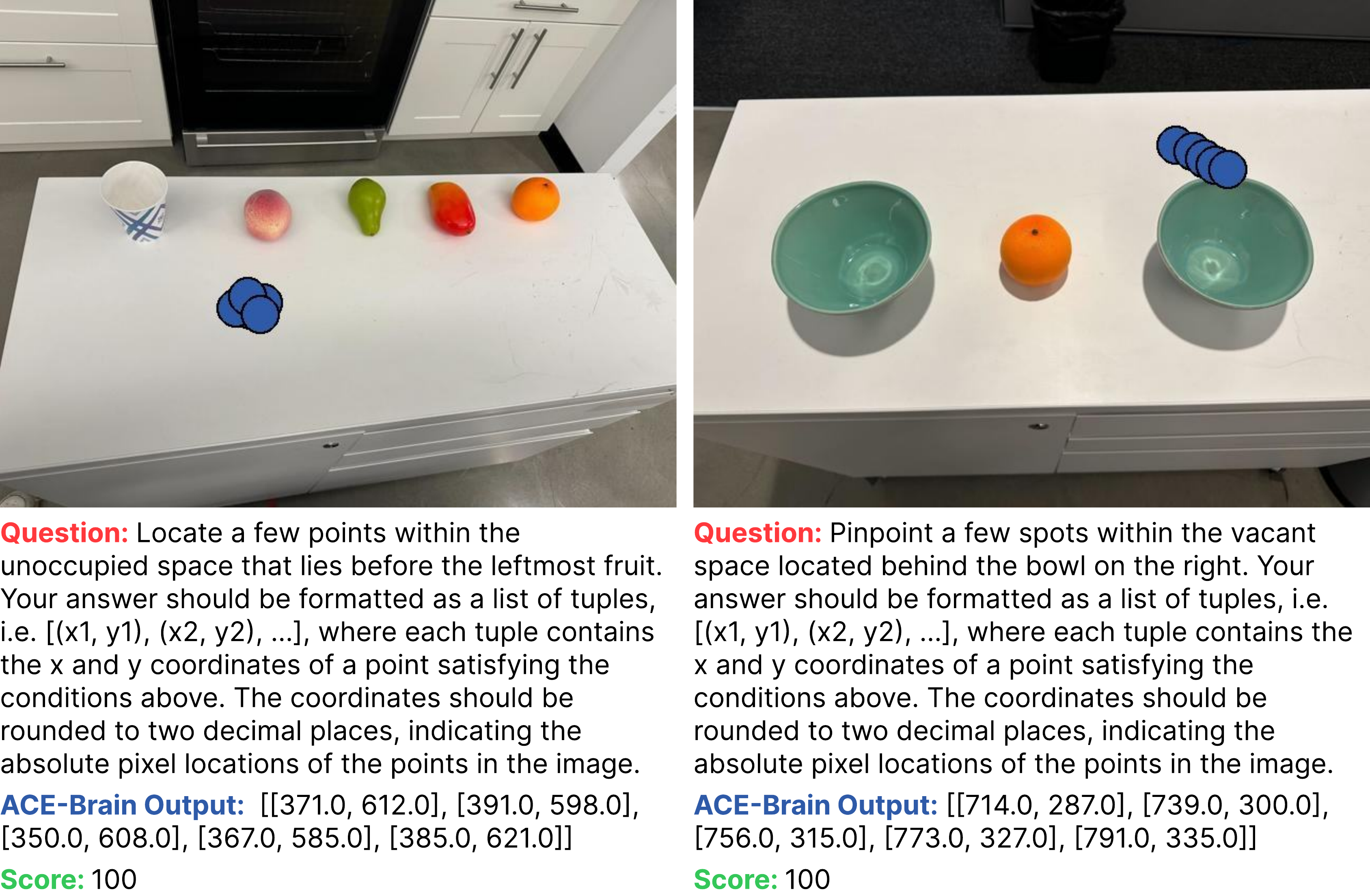}
    \caption{Examples 3, 4 of RoboAfford Benchmark.}
    \label{fig:roboafford-34}
\end{figure}

\begin{figure}
    \centering
    \includegraphics[width=\linewidth]{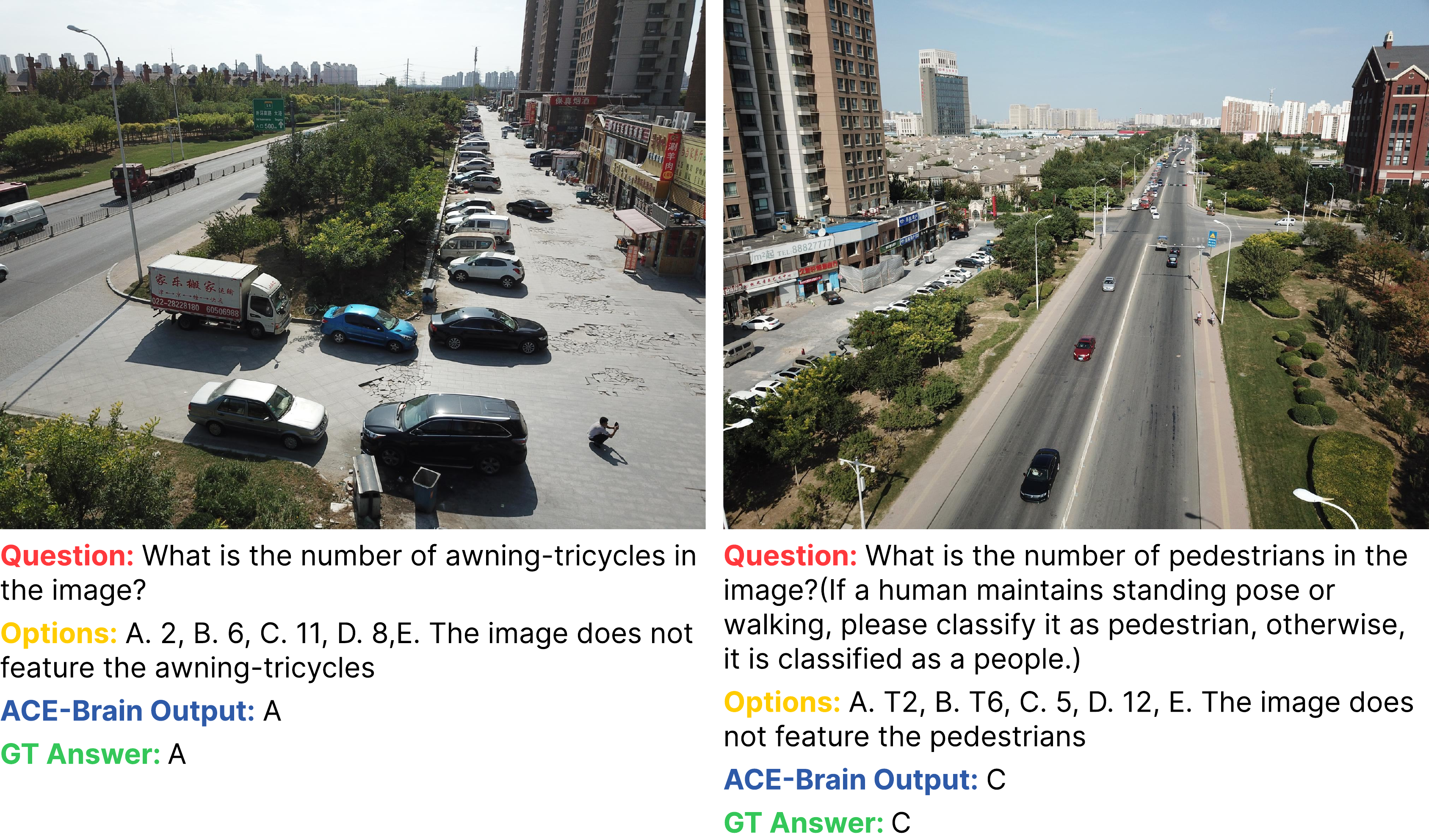}
    \caption{Examples 1, 2 of MME-RealWorld Benchmark.}
    \label{fig:mme-realworld-12}
\end{figure}
\begin{figure}
    \centering
    \includegraphics[width=\linewidth]{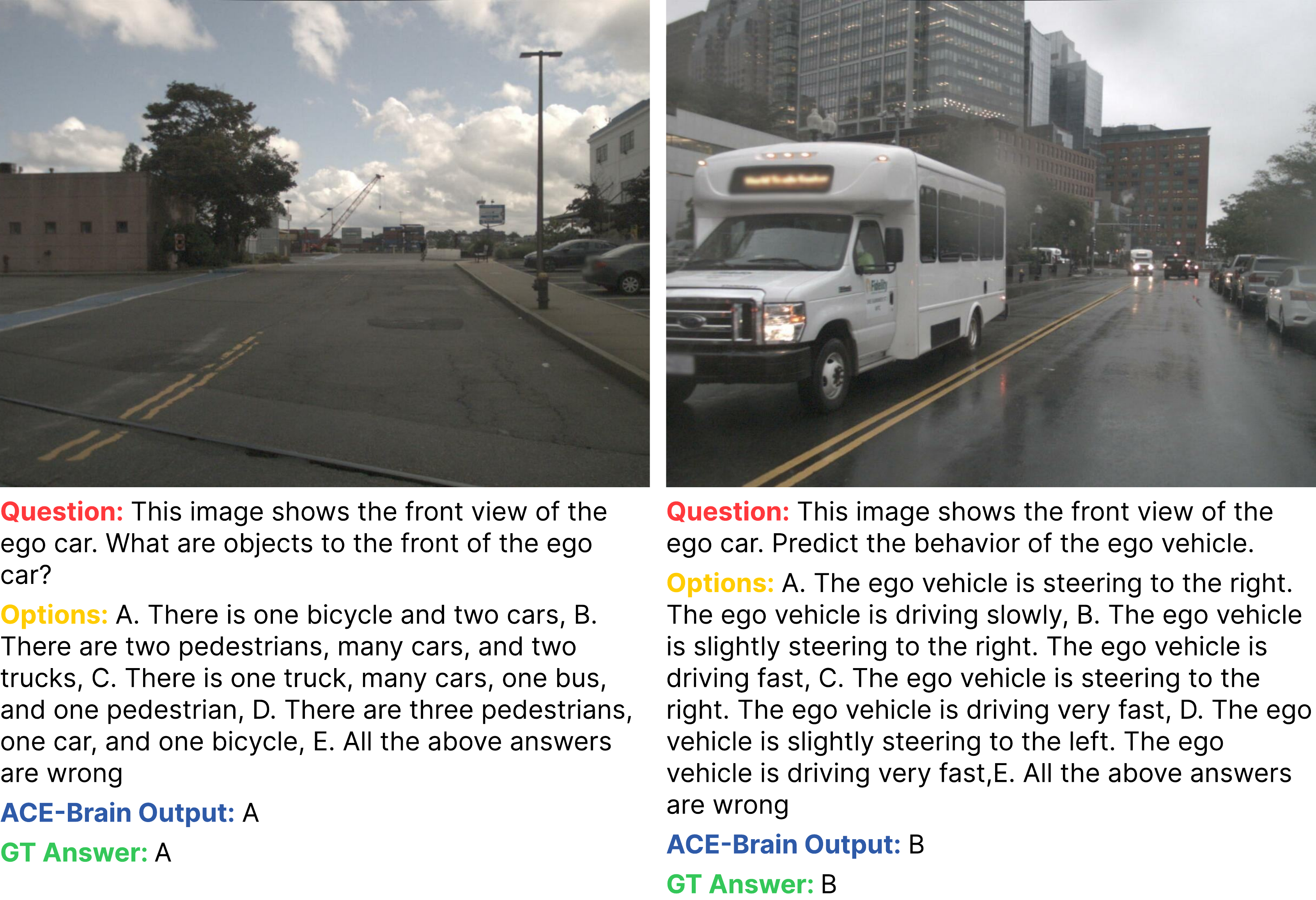}
    \caption{Examples 3, 4 of MME-RealWorld Benchmark.}
    \label{fig:mme-realworld-34}
\end{figure}

\begin{figure}
    \centering
    \includegraphics[width=\linewidth]{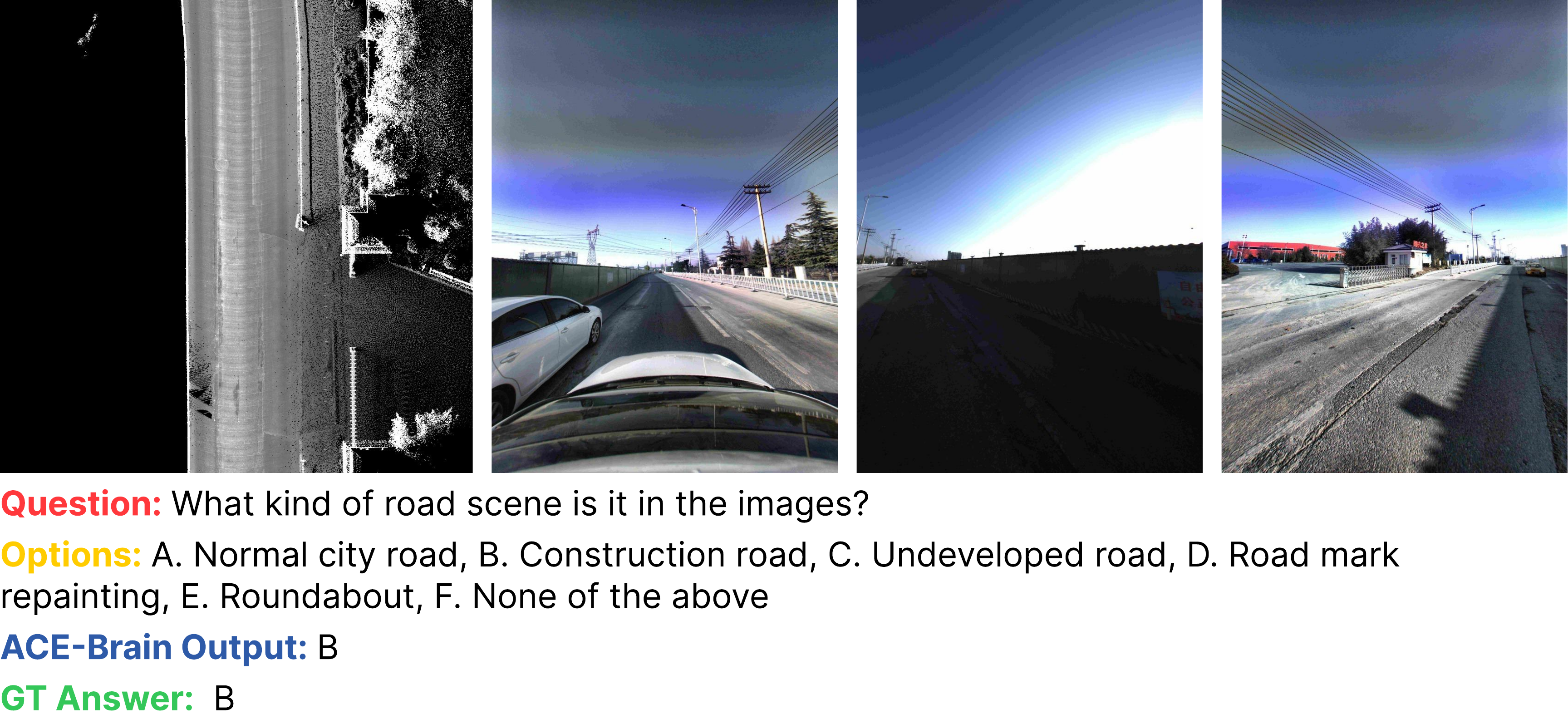}
    \caption{Example 1 of MAPLM Benchmark.}
    \label{fig:maplm-1}
\end{figure}
\begin{figure}
    \centering
    \includegraphics[width=\linewidth]{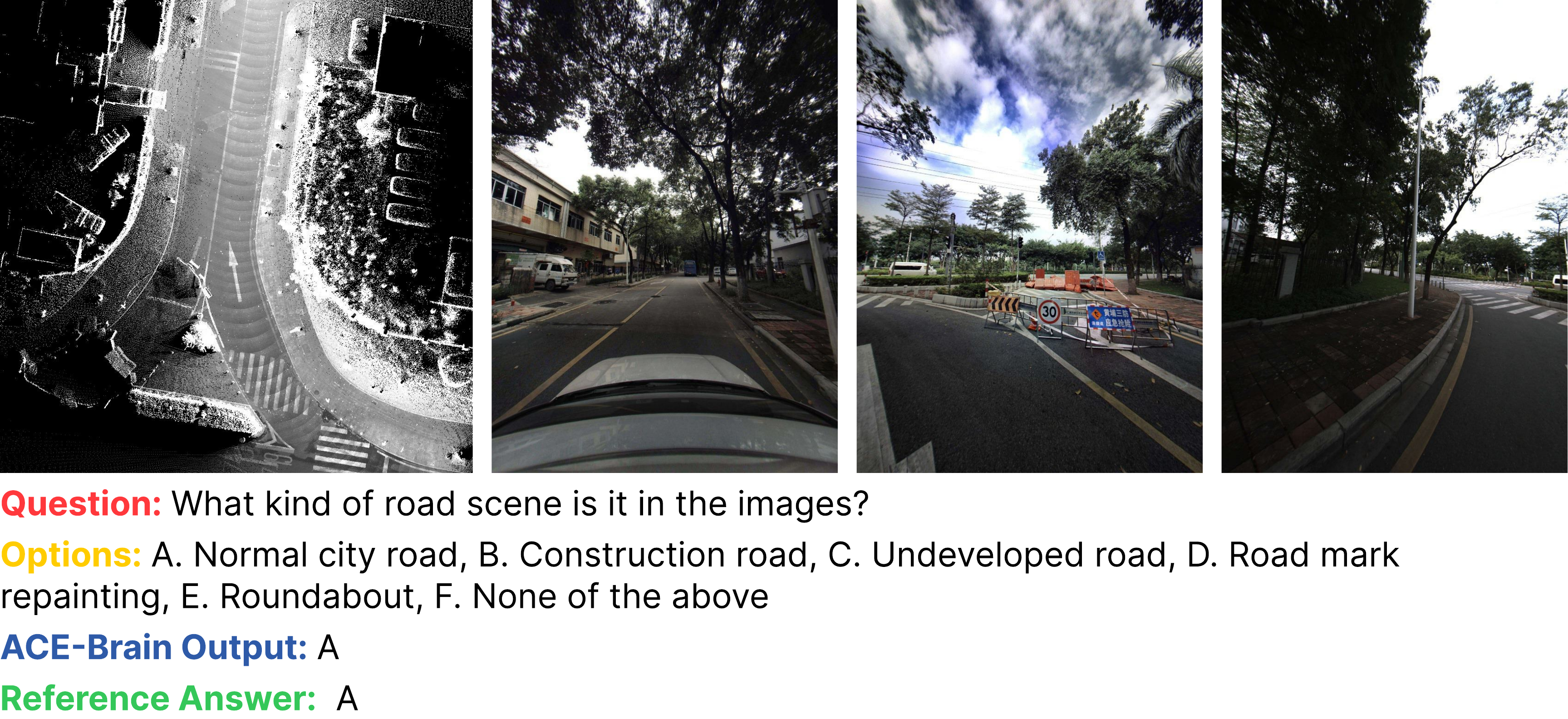}
    \caption{Example 2 of MAPLM Benchmark.}
    \label{fig:maplm-2}
\end{figure}

\begin{figure}
    \centering
    \includegraphics[width=\linewidth]{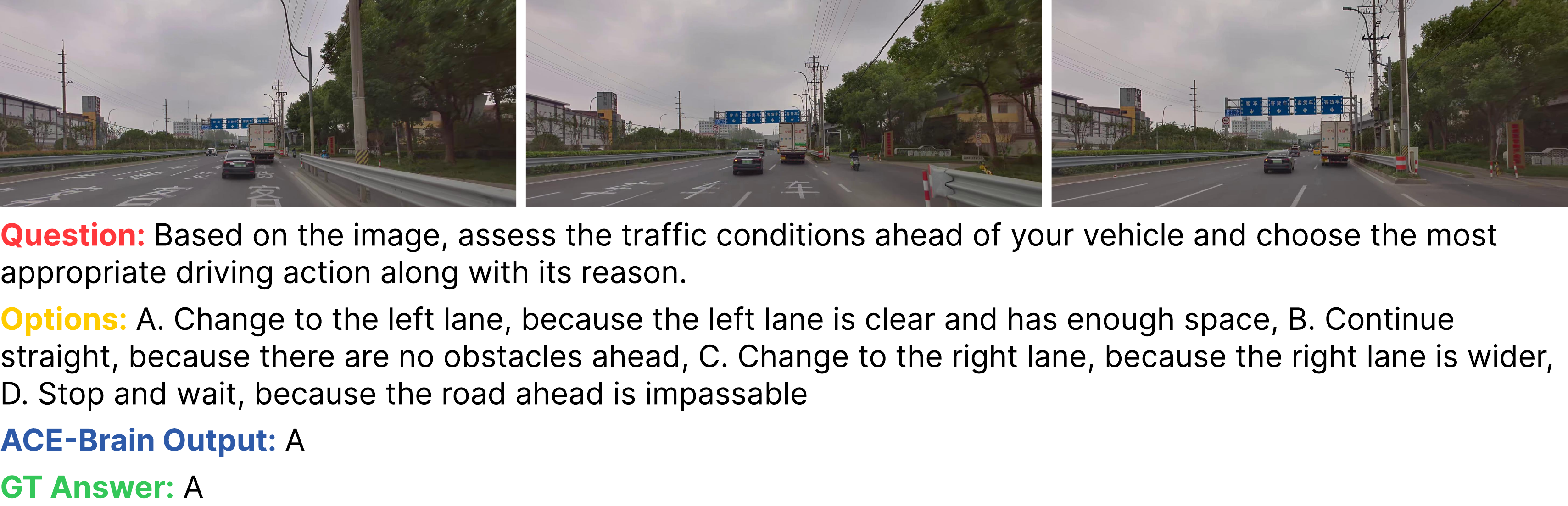}
    \caption{Example 1 of DriveAction Benchmark.}
    \label{fig:driveaction1}
\end{figure}
\begin{figure}
    \centering
    \includegraphics[width=\linewidth]{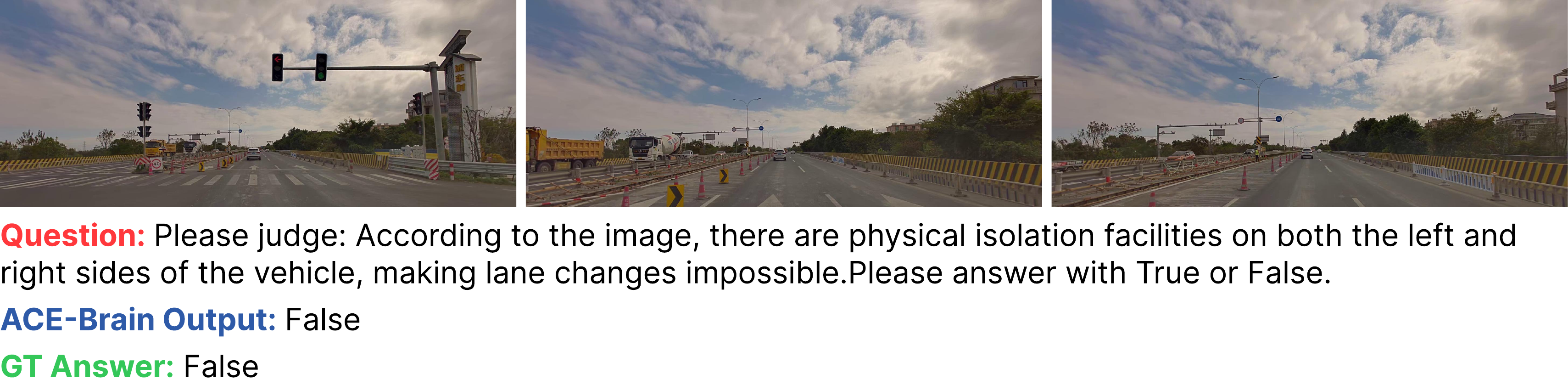}
    \caption{Example 2 of DriveAction Benchmark.}
    \label{fig:driveaction2}
\end{figure}

\begin{figure}
    \centering
    \includegraphics[width=\linewidth]{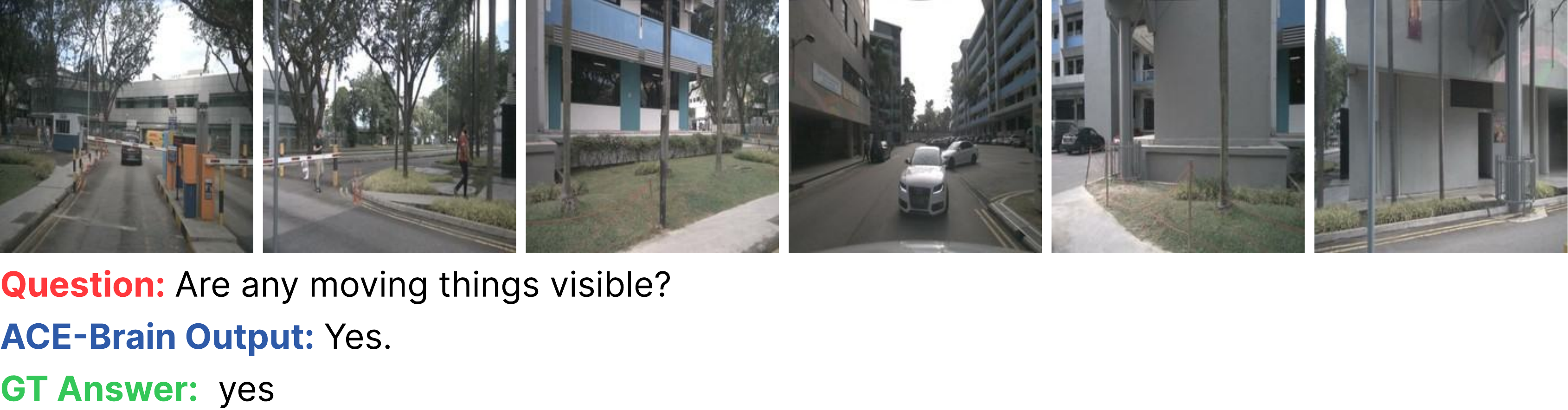}
    \caption{Example 1 of NuScenesQA Benchmark.}
    \label{fig:nuscenesqa-1}
\end{figure}
\begin{figure}
    \centering
    \includegraphics[width=\linewidth]{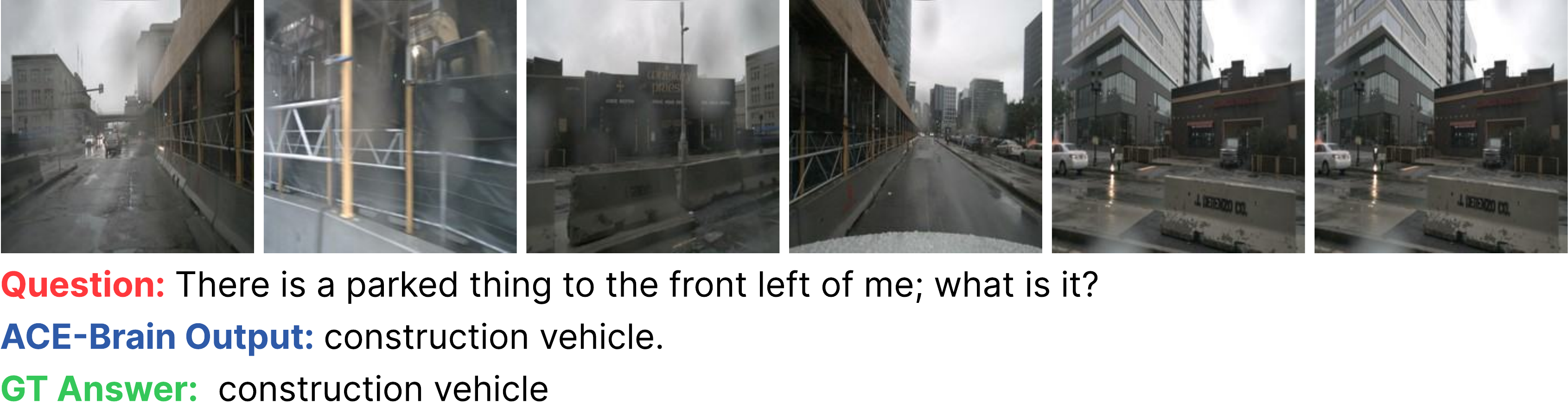}
    \caption{Example 2 of NuScenesQA Benchmark.}
    \label{fig:nuscenesqa-2}
\end{figure}

\begin{figure}
    \centering
    \includegraphics[width=\linewidth]{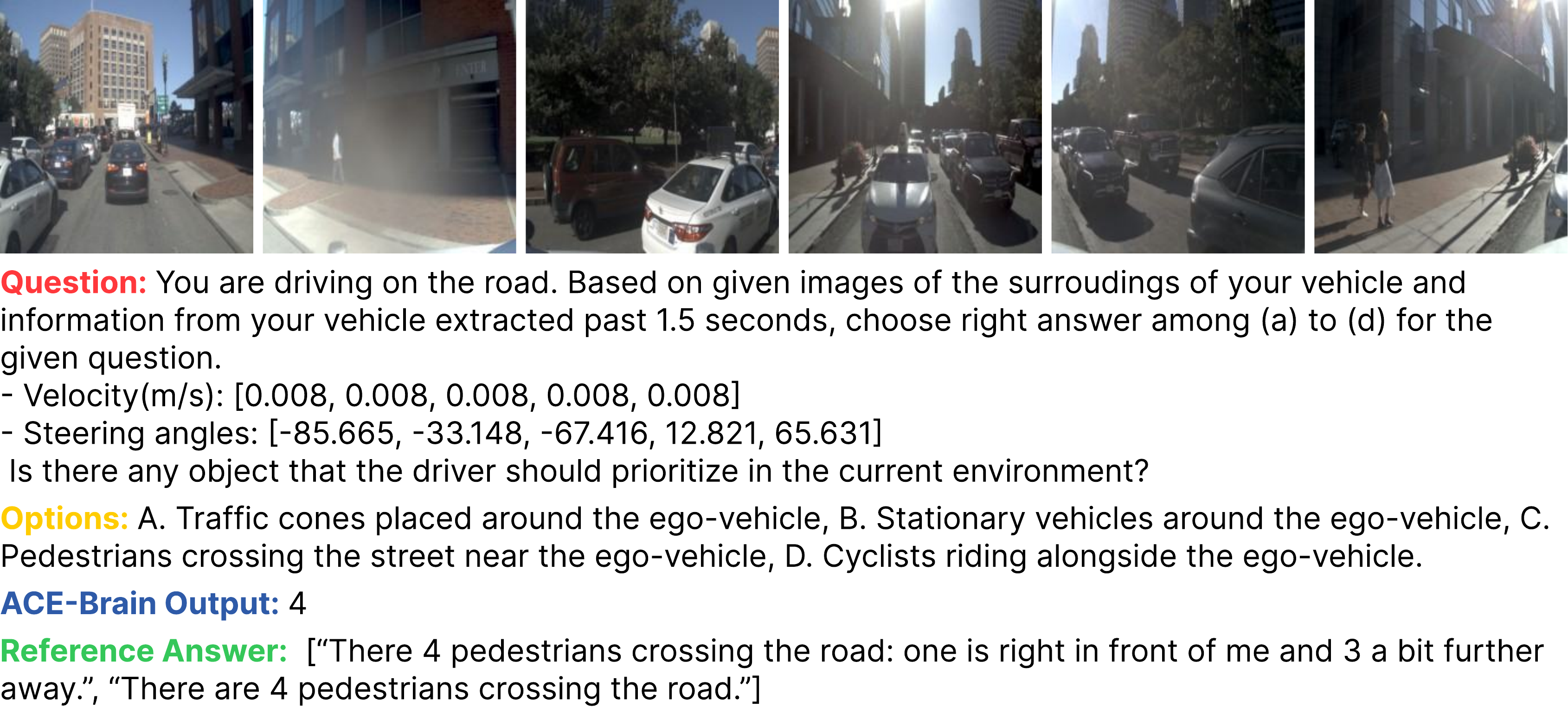}
    \caption{Example 1 of NuPlanQA Benchmark.}
    \label{fig:nuplanqa-1}
\end{figure}
\begin{figure}
    \centering
    \includegraphics[width=\linewidth]{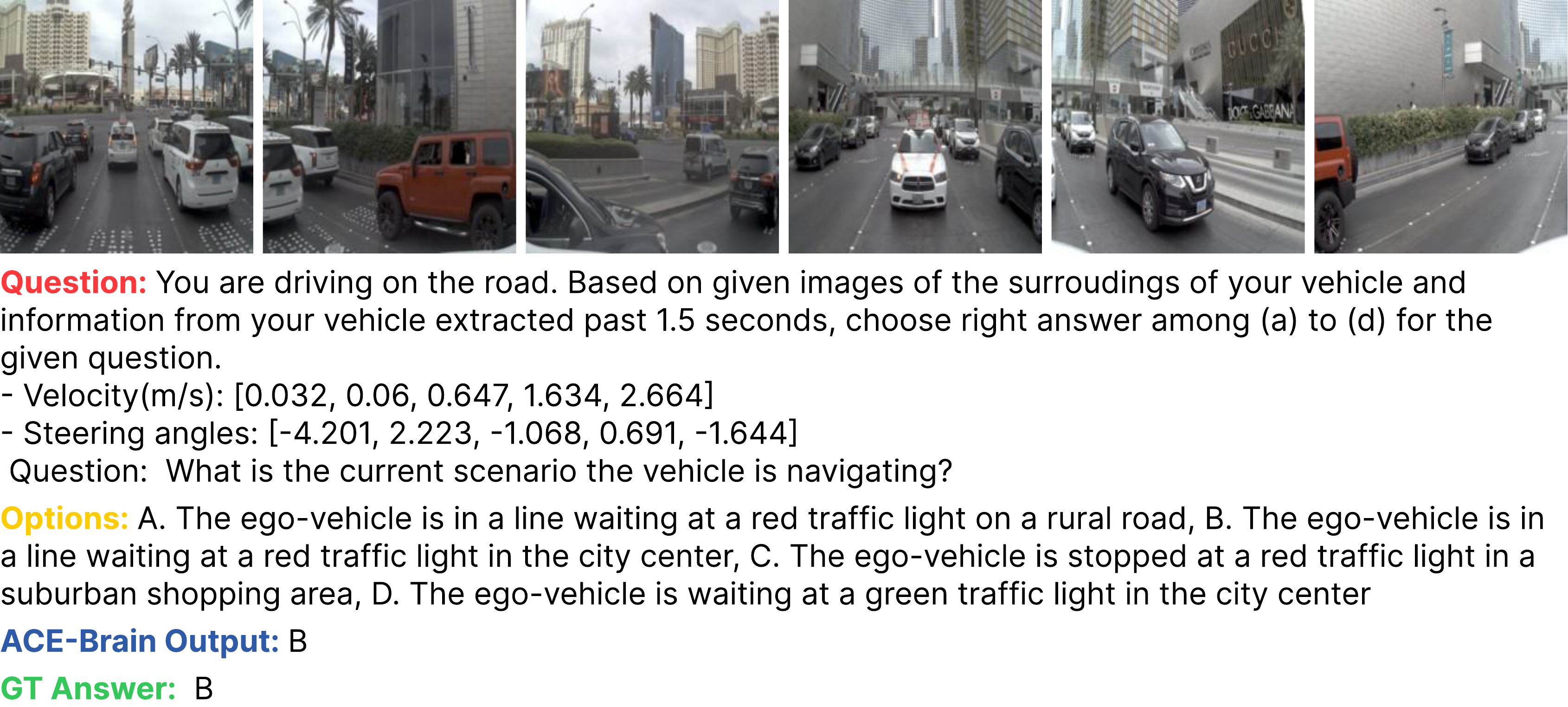}
    \caption{Example 2 of NuPlanQA Benchmark.}
    \label{fig:nuplanqa-2}
\end{figure}

\begin{figure}
    \centering
    \includegraphics[width=\linewidth]{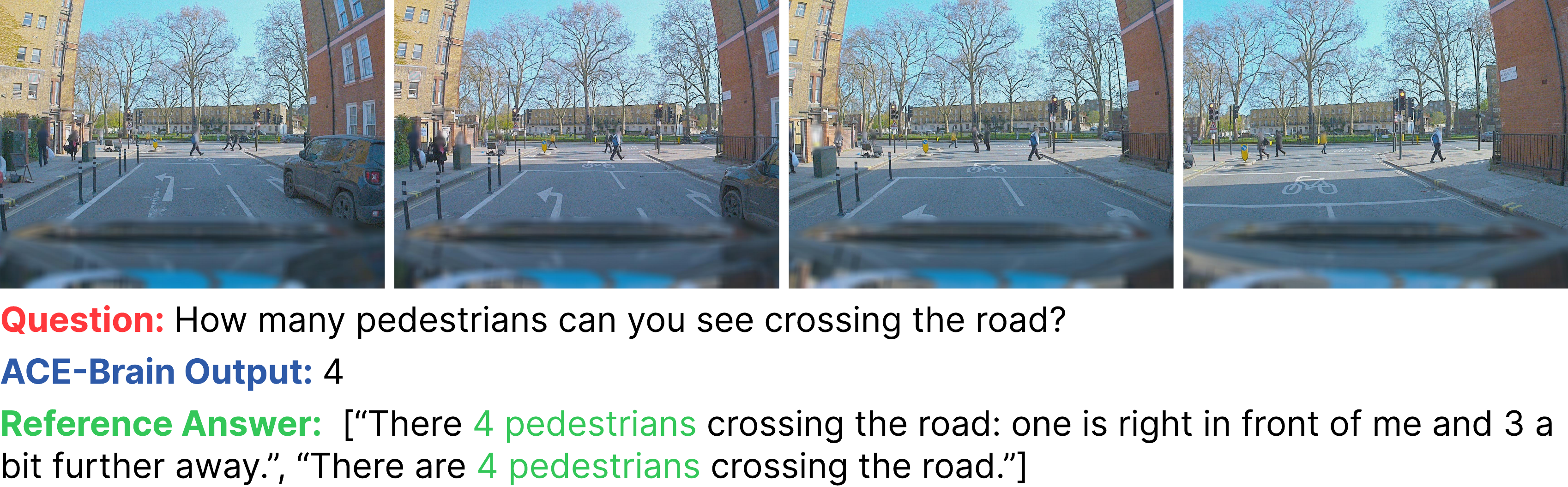}
    \caption{Example 1 of LingoQA Benchmark.}
    \label{fig:lingoqa-1}
\end{figure}
\begin{figure}
    \centering
    \includegraphics[width=\linewidth]{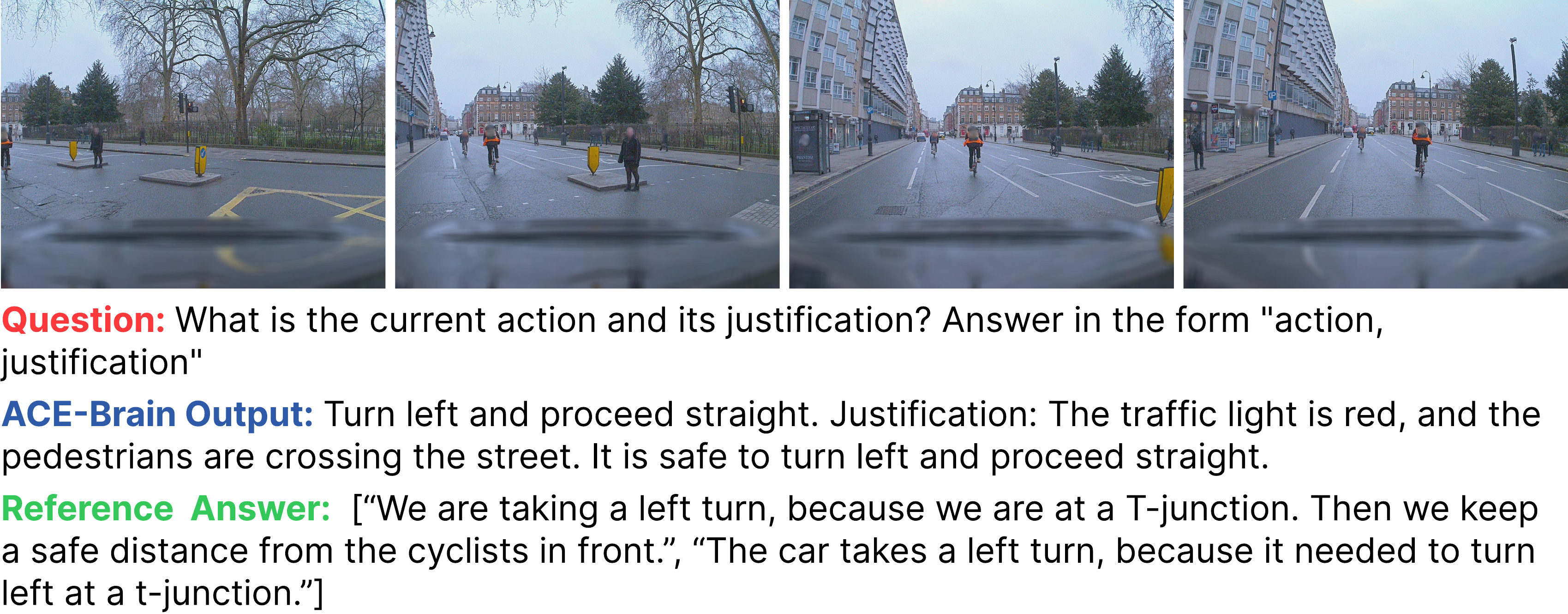}
    \caption{Example 2 of LingoQA Benchmark.}
    \label{fig:lingoqa-2}
\end{figure}

\clearpage

\begin{figure}[!p]
    \centering
    \includegraphics[width=\linewidth,]{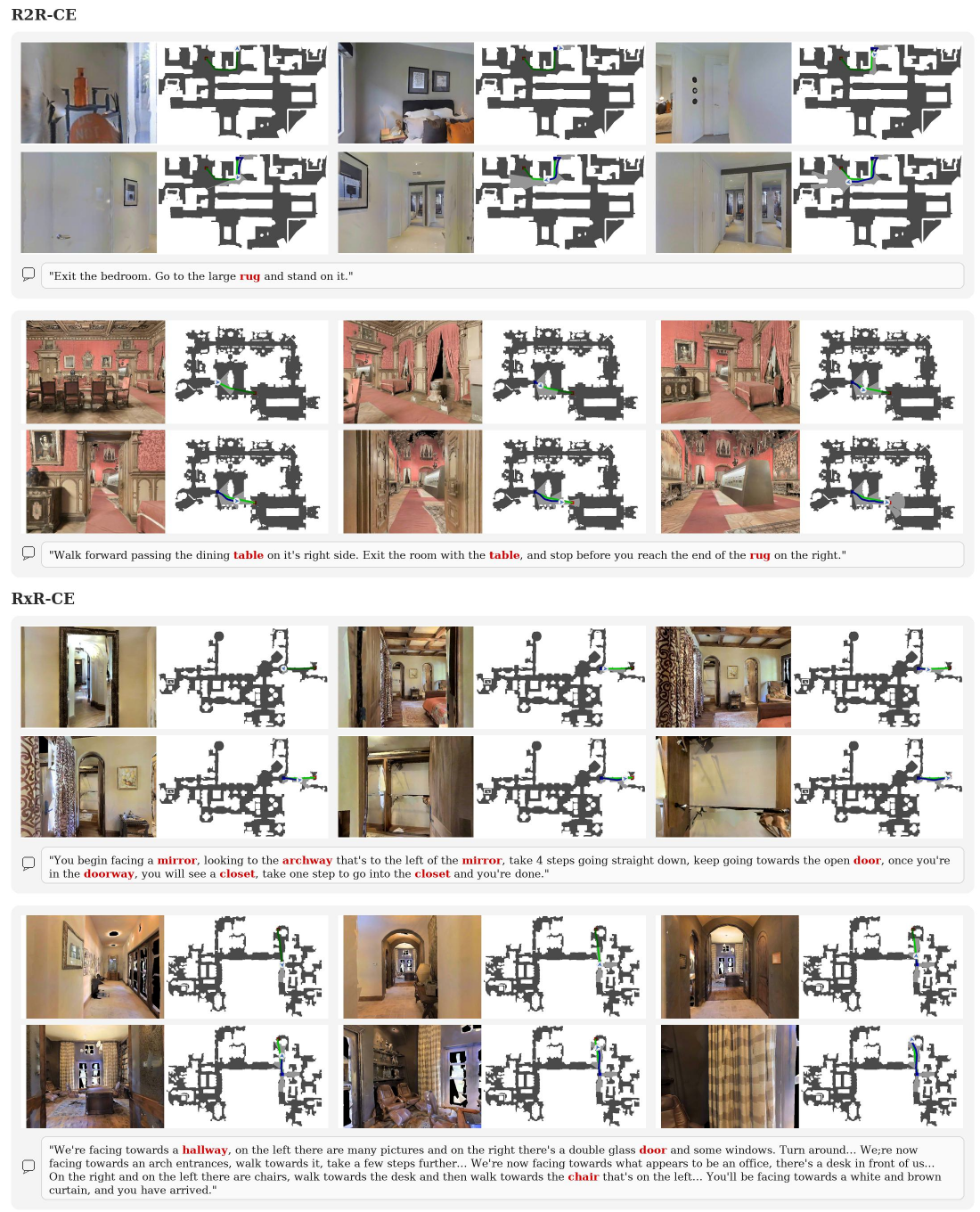}
    \caption{Example of VLN-CE Benchmark.}
    \label{fig:vlnce}
\end{figure}

\clearpage

\begin{figure}[!htbp]
    \centering
    \includegraphics[width=\linewidth]{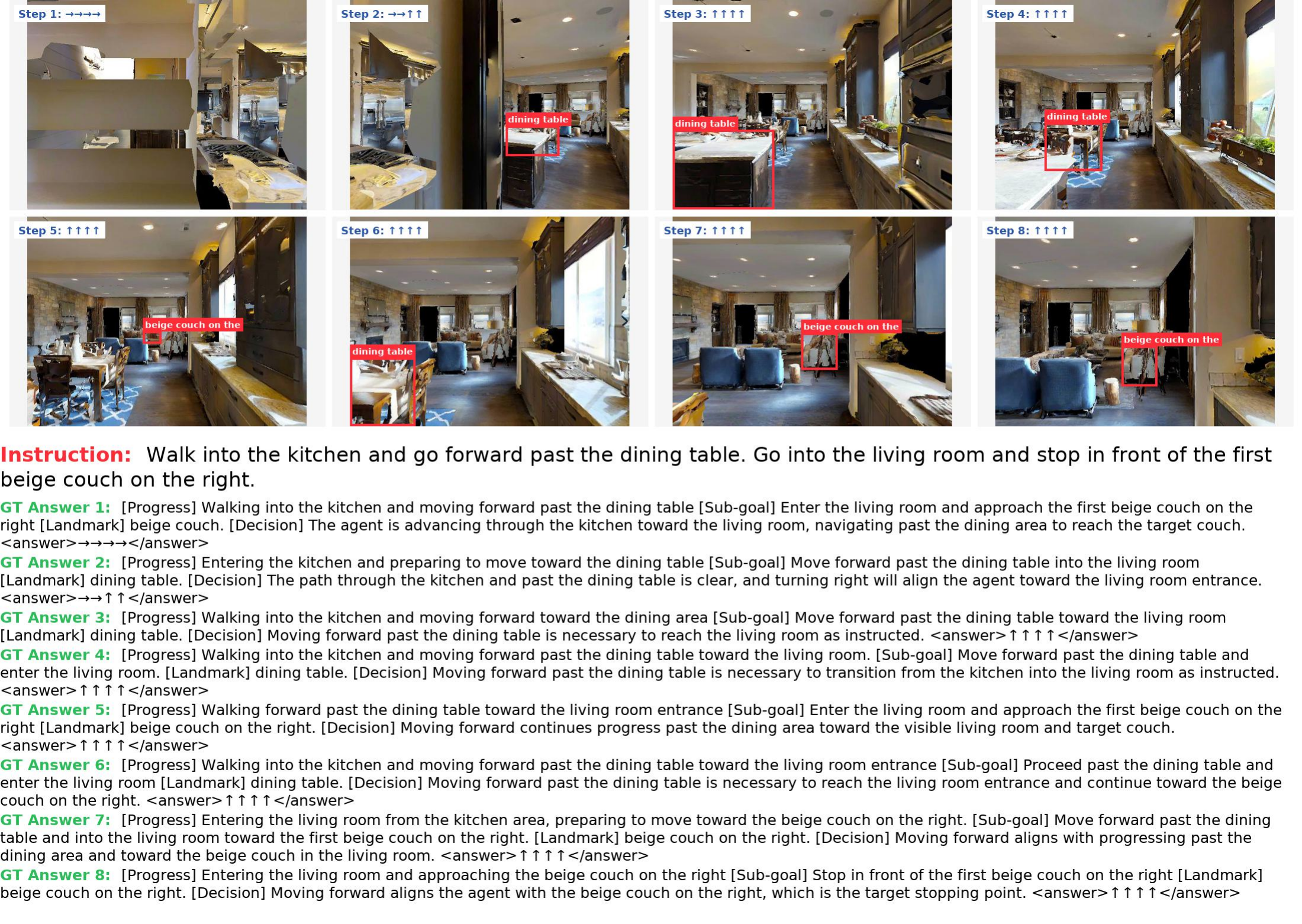}
    \caption{Example of R2R/RxR CoT data.}
    \label{fig:cot_nav}
\end{figure}

\begin{figure}[!htbp]
    \centering
    \includegraphics[width=\linewidth]{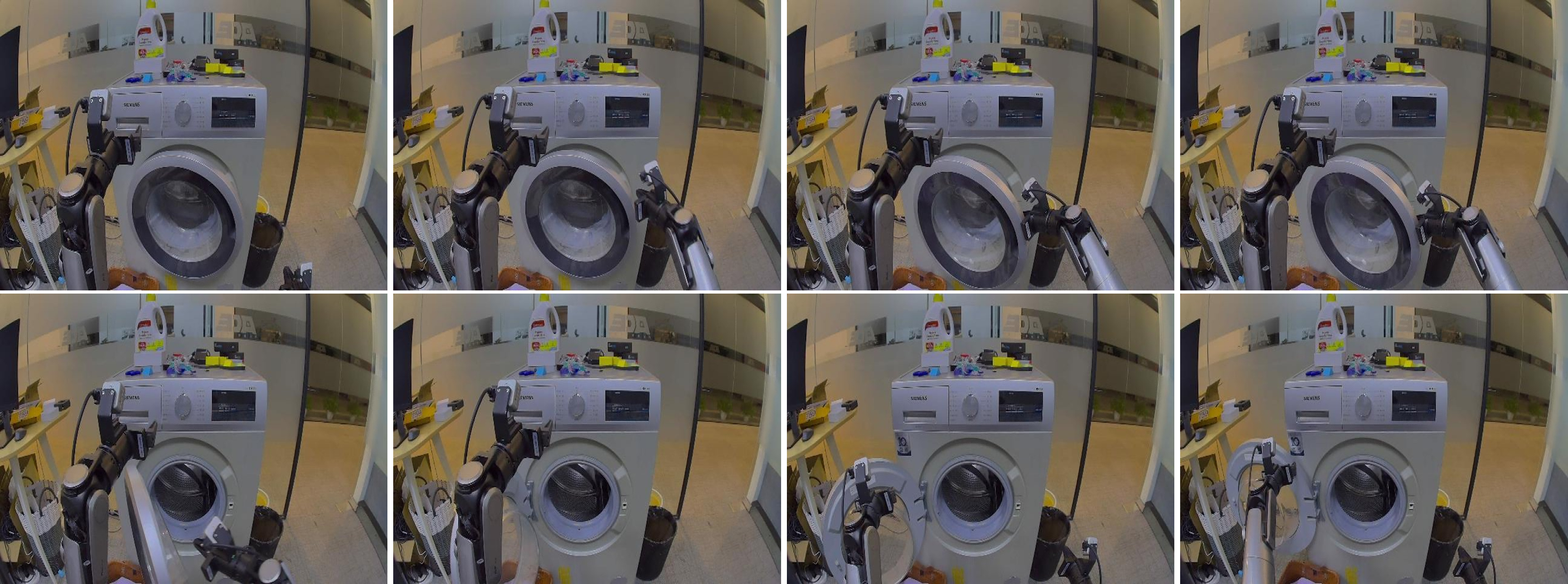}
    \caption{Example of opening the Washing machine.}
    \label{fig:open}
\end{figure}

\begin{figure}[!htbp]
    \centering
    \includegraphics[width=\linewidth]{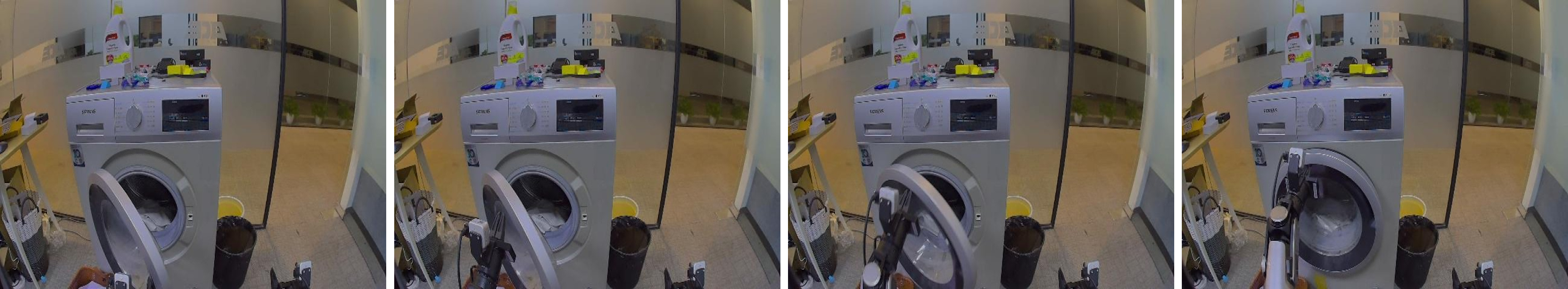}
    \caption{Example of closing the Washing machine.}
    \label{fig:close}
\end{figure}

%% file: tables/rbm_eval_refined_tasks.tex
\begin{table}[htb]
\centering
\caption{\textbf{Selected Tasks for RBM-EVAL-Refined.}
We refine tasks whose progress estimation requires temporal reasoning beyond static final-state recognition.}
\label{tab:rbm_eval_refined_tasks}
\small
\renewcommand{\arraystretch}{1.08}
\begin{tabularx}{\textwidth}{
@{}
>{\centering\arraybackslash}p{0.15\textwidth}
>{\raggedright\arraybackslash}p{0.30\textwidth}
>{\raggedright\arraybackslash}X
@{}
}
\toprule
\textbf{Split} & \textbf{Task} & \textbf{Reason} \\
\midrule

\multirow{8}{=}{\centering\makecell[c]{RBM-EVAL-\\OOD}}
& Pick up the spatula and stir the beans in the pot
& Requires recognizing a stirring process rather than a static state. \\
& Stir the pot
& Progress depends on continuous motion over time. \\
& Press the button
& Button pressing is transient and may leave only subtle visual changes. \\
& Move the orange cup from left to right
& Requires knowing the initial position and motion direction. \\
& Move the orange cup from right to left
& The same state can imply different progress under reversed order. \\
& Separate the purple and orange cups
& Requires observing the transition from close/contact to separated. \\
& Separate the purple and red cups
& Static separation alone does not reveal task progress. \\
& Separate the red and orange cups
& Progress depends on the change in spatial relation over time. \\

\midrule

\multirow{12}{=}{\centering\makecell[c]{RBM-EVAL-\\ID}}
& Drag the pepper across the table
& Requires displacement from an initial to a later position. \\
& Push the eraser across the table
& Progress is defined by motion across the table. \\
& Wipe red bowl with sponge
& Wiping depends on repeated contact and motion. \\
& Wipe the white sponge on the table
& The action cannot be judged from object presence alone. \\
& Wipe tray with sponge
& Requires observing whether wiping has occurred. \\
& Pour
& Pouring is a temporal transfer process. \\
& Make coffee
& Completion depends on a sequence of state changes. \\
& Press the button from top
& Pressing is brief and contact-based. \\
& Press the coffee button
& Pressed and unpressed states may look similar. \\
& Press the handle from side
& Requires action direction and contact dynamics. \\
& Push the coffee cup
& Progress depends on movement relative to the initial state. \\
& Turn door lock counter-clockwise
& Rotation direction is defined relative to the initial pose. \\

\bottomrule
\end{tabularx}
\end{table}

%% file: references.bib
@article{zhang2026pelican,
  title={Pelican-Unify 1.0: A Unified Embodied Intelligence Model for Understanding, Reasoning, Imagination and Action},
  author={Zhang, Yi and Chen, Yinda and Liu, Che and Ding, Zeyuan and Xu, Jin and Zou, Shilong and Liao, Junwei and Hu, Jiayu and Ren, Xiancong and Zhang, Xiaopeng and others},
  journal={arXiv preprint arXiv:2605.15153},
  year={2026}
}

@article{wang2026qwen,
  title={Qwen-vla: Unifying vision-language-action modeling across tasks, environments, and robot embodiments},
  author={Wang, Qiuyue and Li, Mingsheng and Guan, Jian and Ye, Jinhui and Xie, Sicheng and Liu, Yitao and Chen, Junhao and Liang, Zhixuan and Zhang, Jie and Hu, Xintong and others},
  journal={arXiv preprint arXiv:2605.30280},
  year={2026}
}

@article{black2024pi_0,
  title={$\pi_0$: A Vision-Language-Action Flow Model for General Robot Control},
  author={Black, Kevin and Brown, Noah and Driess, Danny and Esmail, Adnan and Equi, Michael and Finn, Chelsea and Fusai, Niccolo and Groom, Lachy and Hausman, Karol and Ichter, Brian and others},
  journal={arXiv preprint arXiv:2410.24164},
  year={2024}
}

@article{yang2026abot,
  title={Abot-m0: Vla foundation model for robotic manipulation with action manifold learning},
  author={Yang, Yandan and Zeng, Shuang and Lin, Tong and Chang, Xinyuan and Qi, Dekang and Xiao, Junjin and Liu, Haoyun and Chen, Ronghan and Chen, Yuzhi and Huo, Dongjie and others},
  journal={arXiv preprint arXiv:2602.11236},
  year={2026}
}

@article{intelligence2604pi0,
  title={$\pi$0. 7: a steerable generalist robotic foundation model with emergent capabilities, 2026},
  author={Intelligence, Physical and Ai, Bo and Amin, Ali and Aniceto, R and Balakrishna, A and Balke, G and Black, K and Bokinsky, G and Cao, S and Charbonnier, T and others},
  journal={URL https://arxiv. org/abs/2604.15483}
}

@article{zhang2026qwen,
  title={Qwen-RobotWorld Technical Report: Unifying Embodied World Modeling through Language-Conditioned Video Generation},
  author={Zhang, Jie and Chen, Xiaoyue and Chen, Anzhe and Lv, Chenxu and Li, Deqing and Zhou, Gengze and Yin, Hang and Yuan, Haoqi and Li, Haoyang and Li, Jiahao and others},
  journal={arXiv preprint arXiv:2606.17030},
  year={2026}
}

@misc{figure2026helix02,
  title={Introducing helix 02: Full-body autonomy},
  author={Figure, AI},
  year={2026}
}

@article{chu2026abotn0,
  title={Abot-n0: Technical report on the vla foundation model for versatile embodied navigation},
  author={Chu, Zedong and Xie, Shichao and Wu, Xiaolong and Shen, Yanfen and Luo, Minghua and Wang, Zhengbo and Liu, Fei and Leng, Xiaoxu and Hu, Junjun and Yin, Mingyang and others},
  journal={arXiv preprint arXiv:2602.11598},
  year={2026}
}

@article{zhang2026qwennav,
  title={Qwen-RobotNav Technical Report: A Scalable Navigation Model Designed for an Agentic Navigation System},
  author={Zhang, Jiazhao and Zhou, Gengze and Yin, Hale and Huang, Yiyang and Lei, Zixing and Peng, Qihang and Yuan, Haoqi and Zhang, Jie and Guo, Xudong and Chen, Xiaoyue and others},
  journal={arXiv preprint arXiv:2606.18112},
  year={2026}
}

@article{yuan2026qwen,
  title={Qwen-robotmanip technical report: Alignment unlocks scale for robotic manipulation foundation models},
  author={Yuan, Haoqi and Liang, Zhixuan and Chen, Anzhe and Wang, Ye and Li, Haoyang and Lin, Pei and Huang, Yiyang and Lei, Zixing and Zhang, Tong and Zhang, Jiazhao and others},
  journal={arXiv preprint arXiv:2606.17846},
  year={2026}
}

@article{fang2026molmoact2,
  title={Molmoact2: Action reasoning models for real-world deployment},
  author={Fang, Haoquan and Duan, Jiafei and Clay, Donovan and Wang, Sam and Liu, Shuo and Huang, Weikai and Fan, Xiang and Tsai, Wei-Chuan and Chen, Shirui and Wang, Yi Ru and others},
  journal={arXiv preprint arXiv:2605.02881},
  year={2026}
}

@article{chen2025robotwin20scalabledata,
  title={Robotwin 2.0: A scalable data generator and benchmark with strong domain randomization for robust bimanual robotic manipulation},
  author={Chen, Tianxing and Chen, Zanxin and Chen, Baijun and Cai, Zijian and Liu, Yibin and Li, Zixuan and Liang, Qiwei and Lin, Xianliang and Ge, Yiheng and Gu, Zhenyu and others},
  journal={arXiv preprint arXiv:2506.18088},
  year={2025}
}

@misc{interleave-vla,
      title={Interleave-VLA: Enhancing Robot Manipulation with Interleaved Image-Text Instructions}, 
      author={Cunxin Fan and Xiaosong Jia and Yihang Sun and Yixiao Wang and Jianglan Wei and Ziyang Gong and Xiangyu Zhao and Masayoshi Tomizuka and Xue Yang and Junchi Yan and Mingyu Ding},
      year={2025},
      eprint={2505.02152},
      archivePrefix={arXiv},
      primaryClass={cs.RO},
      url={https://arxiv.org/abs/2505.02152}, 
}

@article{cheang2024gr2,
  title={Gr-2: A generative video-language-action model with web-scale knowledge for robot manipulation},
  author={Cheang, Chi-Lam and Chen, Guangzeng and Jing, Ya and Kong, Tao and Li, Hang and Li, Yifeng and Liu, Yuxiao and Wu, Hongtao and Xu, Jiafeng and Yang, Yichu and others},
  journal={arXiv preprint arXiv:2410.06158},
  year={2024}
}

@article{ye2026world,
  title={World action models are zero-shot policies},
  author={Ye, Seonghyeon and Ge, Yunhao and Zheng, Kaiyuan and Gao, Shenyuan and Yu, Sihyun and Kurian, George and Indupuru, Suneel and Tan, You Liang and Zhu, Chuning and Xiang, Jiannan and others},
  journal={arXiv preprint arXiv:2602.15922},
  year={2026}
}

@article{guo2026xwam,
  title={Unified 4D world action modeling from video priors with asynchronous denoising},
  author={Guo, Jun and Li, Qiwei and Li, Peiyan and Chen, Zilong and Sun, Nan and Su, Yifei and Wang, Heyun and Zhang, Yuan and Li, Xinghang and Liu, Huaping},
  journal={arXiv preprint arXiv:2604.26694},
  year={2026}
}

@article{ye2026gigaworld,
  title={GigaWorld-Policy: An Efficient Action-Centered World--Action Model},
  author={Ye, Angen and Wang, Boyuan and Ni, Chaojun and Huang, Guan and Zhao, Guosheng and Li, Hao and Li, Hengtao and Li, Jie and Lv, Jindi and Liu, Jingyu and others},
  journal={arXiv preprint arXiv:2603.17240},
  year={2026}
}

@misc{kim2026cosmospolicy,
      title={Cosmos Policy: Fine-Tuning Video Models for Visuomotor Control and Planning}, 
      author={Moo Jin Kim and Yihuai Gao and Tsung-Yi Lin and Yen-Chen Lin and Yunhao Ge and Grace Lam and Percy Liang and Shuran Song and Ming-Yu Liu and Chelsea Finn and Jinwei Gu},
      year={2026},
      eprint={2601.16163},
      archivePrefix={arXiv},
      primaryClass={cs.AI},
      url={https://arxiv.org/abs/2601.16163}, 
}

@misc{sun2026vlajepa,
      title={VLA-JEPA: Enhancing Vision-Language-Action Model with Latent World Model}, 
      author={Jingwen Sun and Wenyao Zhang and Zekun Qi and Shaojie Ren and Zezhi Liu and Hanxin Zhu and Guangzhong Sun and Xin Jin and Zhibo Chen},
      year={2026},
      eprint={2602.10098},
      archivePrefix={arXiv},
      primaryClass={cs.RO},
      url={https://arxiv.org/abs/2602.10098}, 
}

@misc{yuan2026fastwam,
      title={Fast-WAM: Do World Action Models Need Test-time Future Imagination?}, 
      author={Tianyuan Yuan and Zibin Dong and Yicheng Liu and Hang Zhao},
      year={2026},
      eprint={2603.16666},
      archivePrefix={arXiv},
      primaryClass={cs.CV},
      url={https://arxiv.org/abs/2603.16666}, 
}

@inproceedings{walke2023bridgedata,
  title={BridgeData V2: A Dataset for Robot Learning at Scale},
  author={Walke, Homer Rich and Black, Kevin and Zhao, Tony Z and Vuong, Quan and Zheng, Chongyi and Hansen-Estruch, Philippe and He, Andre Wang and Myers, Vivek and Kim, Moo Jin and Du, Max and others},
  booktitle={Conference on Robot Learning (CoRL)},
  year={2023}
}

@misc{vita,
      title={VITA-VLA: Efficiently Teaching Vision-Language Models to Act via Action Expert Distillation}, 
      author={Shaoqi Dong and Chaoyou Fu and Haihan Gao and Yi-Fan Zhang and Chi Yan and Chu Wu and Xiaoyu Liu and Yunhang Shen and Jing Huo and Deqiang Jiang and Haoyu Cao and Yang Gao and Xing Sun and Ran He and Caifeng Shan},
      year={2025},
      eprint={2510.09607},
      archivePrefix={arXiv},
      primaryClass={cs.CV},
      url={https://arxiv.org/abs/2510.09607}, 
}

@misc{shi2025memoryvla,
      title={MemoryVLA: Perceptual-Cognitive Memory in Vision-Language-Action Models for Robotic Manipulation}, 
      author={Hao Shi and Bin Xie and Yingfei Liu and Lin Sun and Fengrong Liu and Tiancai Wang and Erjin Zhou and Haoqiang Fan and Xiangyu Zhang and Gao Huang},
      year={2026},
      eprint={2508.19236},
      archivePrefix={arXiv},
      primaryClass={cs.RO},
      url={https://arxiv.org/abs/2508.19236}, 
}

@misc{graesser2026geminiroboticser,
  author       = {Laura Graesser and Peng Xu},
  title        = {{Gemini Robotics-ER 1.6}: Powering Real-World Robotics Tasks
                  through Enhanced Embodied Reasoning},
  year         = {2026},
  month        = apr,
  howpublished = {\url{https://deepmind.google/blog/gemini-robotics-er-1-6/}},
  note         = {Google DeepMind Blog, accessed June 26, 2026}
}

@article{yuan2026embodied,
  title   = {Embodied-R1. 5: Evolving Physical Intelligence via Embodied Foundation Models},
  author  = {Yuan, Yifu and Huang, Yaoting and Yao, Xianze and Li, Yutong and Zhang, Shuoheng and Han, Linqi and Li, Pengyi and Sun, Jiangeng and Jia, Wenting and Zhang, Zhao and others},
  journal = {arXiv preprint arXiv:2606.11324},
  year    = {2026}
}

@InProceedings{drivemoe,
    author    = {Yang, Zhenjie and Chai, Yilin and Jia, Xiaosong and Li, Qifeng and Shao, Yuqian and Zhu, Xuekai and Su, Haisheng and Yan, Junchi},
    title     = {DriveMoE: Mixture-of-Experts for Vision-Language-Action Model in End-to-End Autonomous Driving},
    booktitle = {Proceedings of the IEEE/CVF Conference on Computer Vision and Pattern Recognition (CVPR)},
    month     = {June},
    year      = {2026},
    pages     = {10678-10688}
}

@misc{llm4drive,
      title={LLM4Drive: A Survey of Large Language Models for Autonomous Driving}, 
      author={Zhenjie Yang and Xiaosong Jia and Hongyang Li and Junchi Yan},
      year={2023},
      eprint={2311.01043},
      archivePrefix={arXiv},
      primaryClass={cs.AI}
}

@inproceedings{
rawdrive,
title={Raw2Drive: Reinforcement Learning with Aligned World Models for End-to-End Autonomous Driving (in {CARLA} v2)},
author={Zhenjie Yang and Xiaosong Jia and Qifeng Li and Xue Yang and Maoqing Yao and Junchi Yan},
booktitle={The Thirty-ninth Annual Conference on Neural Information Processing Systems},
year={2026},
url={https://openreview.net/forum?id=CAz7UGRdLs}
}

@article{intelligence2025pi,
  title   = {{$\pi^{*}_{0.6}$}: a VLA That Learns From Experience},
  author  = {Intelligence, Physical and Amin, Ali and Aniceto, Raichelle and Balakrishna, Ashwin and Black, Kevin and Conley, Ken and Connors, Grace and Darpinian, James and Dhabalia, Karan and DiCarlo, Jared and others},
  journal = {arXiv preprint arXiv:2511.14759},
  year    = {2025}
}

@article{dang2026rynnbrain,
  title   = {Rynnbrain: Open embodied foundation models},
  author  = {Dang, Ronghao and Guo, Jiayan and Hou, Bohan and Leng, Sicong and Li, Kehan and Li, Xin and Liu, Jiangpin and Mao, Yunxuan and Wang, Zhikai and Yuan, Yuqian and others},
  journal = {arXiv preprint arXiv:2602.14979},
  year    = {2026}
}

@article{agarwal2026cosmos,
  title   = {Cosmos 3: Omnimodal world models for physical ai},
  author  = {Agarwal, Niket and Ali, Arslan and Allen, Jon and Antolini, Martin and Aubame, Adeline and Azzolini, Alisson and Bai, Junjie and Bala, Maciej and Balaji, Yogesh and Bapst, Josh and others},
  journal = {arXiv preprint arXiv:2606.02800},
  year    = {2026}
}

@article{tangFusionBenchUnifiedLibrary2025,
  title   = {FusionBench: A Unified Library and Comprehensive Benchmark for Deep Model Fusion},
  author  = {Anke Tang and Li Shen and Yong Luo and Enneng Yang and Han Hu and Lefei Zhang and Bo Du and Dacheng Tao},
  journal = {arXiv preprint arXiv:2406.03280},
  year    = {2025}
}

@article{zhang2024mme,
  title   = {Mme-realworld: Could your multimodal llm challenge high-resolution real-world scenarios that are difficult for humans?},
  author  = {Zhang, Yi-Fan and Zhang, Huanyu and Tian, Haochen and Fu, Chaoyou and Zhang, Shuangqing and Wu, Junfei and Li, Feng and Wang, Kun and Wen, Qingsong and Zhang, Zhang and others},
  journal = {arXiv preprint arXiv:2408.13257},
  year    = {2024}
}

@inproceedings{cao2024maplm,
  title     = {Maplm: A real-world large-scale vision-language benchmark for map and traffic scene understanding},
  author    = {Cao, Xu and Zhou, Tong and Ma, Yunsheng and Ye, Wenqian and Cui, Can and Tang, Kun and Cao, Zhipeng and Liang, Kaizhao and Wang, Ziran and Rehg, James M and others},
  booktitle = {Proceedings of the IEEE/CVF conference on computer vision and pattern recognition},
  pages     = {21819--21830},
  year      = {2024}
}

@article{hao2025driveaction,
  title   = {Driveaction: A benchmark for exploring human-like driving decisions in vla models},
  author  = {Hao, Yuhan and Li, Zhengning and Sun, Lei and Wang, Weilong and Yi, Naixin and Song, Sheng and Qin, Caihong and Zhou, Mofan and Zhan, Yifei and Lang, Xianpeng},
  journal = {arXiv preprint arXiv:2506.05667},
  year    = {2025}
}

@inproceedings{qian2024nuscenes,
  title     = {Nuscenes-qa: A multi-modal visual question answering benchmark for autonomous driving scenario},
  author    = {Qian, Tianwen and Chen, Jingjing and Zhuo, Linhai and Jiao, Yang and Jiang, Yu-Gang},
  booktitle = {Proceedings of the AAAI Conference on Artificial Intelligence},
  volume    = {38},
  number    = {5},
  pages     = {4542--4550},
  year      = {2024}
}

@article{park2025nuplanqa,
  title   = {Nuplanqa: A large-scale dataset and benchmark for multi-view driving scene understanding in multi-modal large language models},
  author  = {Park, Sung-Yeon and Cui, Can and Ma, Yunsheng and Moradipari, Ahmadreza and Gupta, Rohit and Han, Kyungtae and Wang, Ziran},
  journal = {arXiv preprint arXiv:2503.12772},
  year    = {2025}
}

@inproceedings{marcu2024lingoqa,
  title        = {Lingoqa: Visual question answering for autonomous driving},
  author       = {Marcu, Ana-Maria and Chen, Long and H{\"u}nermann, Jan and Karnsund, Alice and Hanotte, Benoit and Chidananda, Prajwal and Nair, Saurabh and Badrinarayanan, Vijay and Kendall, Alex and Shotton, Jamie and others},
  booktitle    = {European Conference on Computer Vision},
  pages        = {252--269},
  year         = {2024},
  organization = {Springer}
}

@article{bai2025qwen3vltechnicalreport,
  title   = {Qwen3-VL Technical Report},
  author  = {Shuai Bai and Yuxuan Cai and Ruizhe Chen and Keqin Chen and Xionghui Chen and Zesen Cheng and Lianghao Deng and Wei Ding and Chang Gao and Chunjiang Ge and Wenbin Ge and Zhifang Guo and Qidong Huang and Jie Huang and Fei Huang and Binyuan Hui and Shutong Jiang and Zhaohai Li and Mingsheng Li and Mei Li and Kaixin Li and Zicheng Lin and Junyang Lin and Xuejing Liu and Jiawei Liu and Chenglong Liu and Yang Liu and Dayiheng Liu and Shixuan Liu and Dunjie Lu and Ruilin Luo and Chenxu Lv and Rui Men and Lingchen Meng and Xuancheng Ren and Xingzhang Ren and Sibo Song and Yuchong Sun and Jun Tang and Jianhong Tu and Jianqiang Wan and Peng Wang and Pengfei Wang and Qiuyue Wang and Yuxuan Wang and Tianbao Xie and Yiheng Xu and Haiyang Xu and Jin Xu and Zhibo Yang and Mingkun Yang and Jianxin Yang and An Yang and Bowen Yu and Fei Zhang and Hang Zhang and Xi Zhang and Bo Zheng and Humen Zhong and Jingren Zhou and Fan Zhou and Jing Zhou and Yuanzhi Zhu and Ke Zhu},
  year    = {2025},
  journal = {arXiv preprint arXiv:2511.21631}
}

@article{vebrain,
  title   = {Visual embodied brain: Let multimodal large language models see, think, and control in spaces},
  author  = {Luo, Gen and Yang, Ganlin and Gong, Ziyang and Chen, Guanzhou and Duan, Haonan and Cui, Erfei and Tong, Ronglei and Hou, Zhi and Zhang, Tianyi and Chen, Zhe and others},
  journal = {arXiv preprint arXiv:2506.00123},
  year    = {2025}
}

@article{vlaser,
  title   = {Vlaser: Vision-Language-Action Model with Synergistic Embodied Reasoning},
  author  = {Yang, Ganlin and Zhang, Tianyi and Hao, Haoran and Wang, Weiyun and Liu, Yibin and Wang, Dehui and Chen, Guanzhou and Cai, Zijian and Chen, Junting and Su, Weijie and others},
  journal = {arXiv preprint arXiv:2510.11027},
  year    = {2025}
}

@inproceedings{robobrain,
  title     = {Robobrain: A unified brain model for robotic manipulation from abstract to concrete},
  author    = {Ji, Yuheng and Tan, Huajie and Shi, Jiayu and Hao, Xiaoshuai and Zhang, Yuan and Zhang, Hengyuan and Wang, Pengwei and Zhao, Mengdi and Mu, Yao and An, Pengju and others},
  booktitle = {Proceedings of the Computer Vision and Pattern Recognition Conference},
  pages     = {1724--1734},
  year      = {2025}
}

@article{robobrain2,
  title   = {Robobrain 2.0 technical report},
  author  = {Team, BAAI RoboBrain and Cao, Mingyu and Tan, Huajie and Ji, Yuheng and Chen, Xiansheng and Lin, Minglan and Li, Zhiyu and Cao, Zhou and Wang, Pengwei and Zhou, Enshen and others},
  journal = {arXiv preprint arXiv:2507.02029},
  year    = {2025}
}

@article{robobrain2_5,
  title   = {RoboBrain 2.5: Depth in Sight, Time in Mind},
  author  = {Tan, Huajie and Zhou, Enshen and Li, Zhiyu and Xu, Yijie and Ji, Yuheng and Chen, Xiansheng and Chi, Cheng and Wang, Pengwei and Jia, Huizhu and Ao, Yulong and others},
  journal = {arXiv preprint arXiv:2601.14352},
  year    = {2026}
}

@article{pelican-vl,
  title   = {Pelican-VL 1.0: A Foundation Brain Model for Embodied Intelligence},
  author  = {Zhang, Yi and Liu, Che and Ren, Xiancong and Ni, Hanchu and Zhang, Shuai and Ding, Zeyuan and Hu, Jiayu and Shan, Hanzhe and Niu, Zhenwei and Liu, Zhaoyang and others},
  journal = {arXiv preprint arXiv:2511.00108},
  year    = {2025}
}

@article{mimo-embodied,
  title   = {MiMo-Embodied: X-Embodied Foundation Model Technical Report},
  author  = {Hao, Xiaoshuai and Zhou, Lei and Huang, Zhijian and Hou, Zhiwen and Tang, Yingbo and Zhang, Lingfeng and Li, Guang and Lu, Zheng and Ren, Shuhuai and Meng, Xianhui and others},
  journal = {arXiv preprint arXiv:2511.16518},
  year    = {2025}
}

@inproceedings{drivelm,
  title        = {Drivelm: Driving with graph visual question answering},
  author       = {Sima, Chonghao and Renz, Katrin and Chitta, Kashyap and Chen, Li and Zhang, Hanxue and Xie, Chengen and Bei{\ss}wenger, Jens and Luo, Ping and Geiger, Andreas and Li, Hongyang},
  booktitle    = {European conference on computer vision},
  pages        = {256--274},
  year         = {2024},
  organization = {Springer}
}

@article{gemini_robotics,
  title   = {Gemini robotics: Bringing ai into the physical world},
  author  = {Team, Gemini Robotics and Abeyruwan, Saminda and Ainslie, Joshua and Alayrac, Jean-Baptiste and Arenas, Montserrat Gonzalez and Armstrong, Travis and Balakrishna, Ashwin and Baruch, Robert and Bauza, Maria and Blokzijl, Michiel and others},
  journal = {arXiv preprint arXiv:2503.20020},
  year    = {2025}
}

@article{drivegpt4,
  title     = {Drivegpt4: Interpretable end-to-end autonomous driving via large language model},
  author    = {Xu, Zhenhua and Zhang, Yujia and Xie, Enze and Zhao, Zhen and Guo, Yong and Wong, Kwan-Yee K and Li, Zhenguo and Zhao, Hengshuang},
  journal   = {IEEE Robotics and Automation Letters},
  year      = {2024},
  publisher = {IEEE}
}

@inproceedings{rovi,
  title     = {Robotic visual instruction},
  author    = {Li, Yanbang and Gong, Ziyang and Li, Haoyang and Huang, Xiaoqi and Kang, Haolan and Bai, Guangping and Ma, Xianzheng},
  booktitle = {Proceedings of the Computer Vision and Pattern Recognition Conference},
  pages     = {12155--12165},
  year      = {2025}
}

@inproceedings{robospatial,
  title     = {Robospatial: Teaching spatial understanding to 2d and 3d vision-language models for robotics},
  author    = {Song, Chan Hee and Blukis, Valts and Tremblay, Jonathan and Tyree, Stephen and Su, Yu and Birchfield, Stan},
  booktitle = {Proceedings of the Computer Vision and Pattern Recognition Conference},
  pages     = {15768--15780},
  year      = {2025}
}

@inproceedings{roboafford,
  title     = {Roboafford: A dataset and benchmark for enhancing object and spatial affordance learning in robot manipulation},
  author    = {Tang, Yingbo and Zhang, Lingfeng and Zhang, Shuyi and Zhao, Yinuo and Hao, Xiaoshuai},
  booktitle = {Proceedings of the 33rd ACM International Conference on Multimedia},
  pages     = {12706--12713},
  year      = {2025}
}

@misc{chatgpt4o,
  title        = {GPT-4o System Card},
  author       = {OpenAI},
  howpublished = {\url{https://openai.com/index/gpt-4o-system-card/}},
  year         = {2025}
}

@article{qwen2.5-vl,
  title   = {Qwen2. 5-vl technical report},
  author  = {Bai, Shuai and Chen, Keqin and Liu, Xuejing and Wang, Jialin and Ge, Wenbin and Song, Sibo and Dang, Kai and Wang, Peng and Wang, Shijie and Tang, Jun and others},
  journal = {arXiv preprint arXiv:2502.13923},
  year    = {2025}
}

@article{qwen3,
  title   = {Qwen3 technical report},
  author  = {Yang, An and Li, Anfeng and Yang, Baosong and Zhang, Beichen and Hui, Binyuan and Zheng, Bo and Yu, Bowen and Gao, Chang and Huang, Chengen and Lv, Chenxu and others},
  journal = {arXiv preprint arXiv:2505.09388},
  year    = {2025}
}

@article{internvl3,
  title   = {Internvl3: Exploring advanced training and test-time recipes for open-source multimodal models},
  author  = {Zhu, Jinguo and Wang, Weiyun and Chen, Zhe and Liu, Zhaoyang and Ye, Shenglong and Gu, Lixin and Tian, Hao and Duan, Yuchen and Su, Weijie and Shao, Jie and others},
  journal = {arXiv preprint arXiv:2504.10479},
  year    = {2025}
}

@article{gemini,
  title   = {Gemini: a family of highly capable multimodal models},
  author  = {Team, Gemini and Anil, Rohan and Borgeaud, Sebastian and Alayrac, Jean-Baptiste and Yu, Jiahui and Soricut, Radu and Schalkwyk, Johan and Dai, Andrew M and Hauth, Anja and Millican, Katie and others},
  journal = {arXiv preprint arXiv:2312.11805},
  year    = {2023}
}

@article{eo1,
  title   = {EO-1: Interleaved Vision-Text-Action Pretraining for General Robot Control},
  author  = {Delin Qu and Haoming Song and Qizhi Chen and Zhaoqing Chen and Xianqiang Gao and Xinyi Ye and Qi Lv and Modi Shi and Guanghui Ren and Cheng Ruan and Maoqing Yao and Haoran Yang and Jiacheng Bao and Bin Zhao and Dong Wang},
  journal = {arXiv preprint},
  year    = {2025},
  url     = {https://arxiv.org/abs/2508.21112}
}

@inproceedings{vsi,
  title     = {Thinking in space: How multimodal large language models see, remember, and recall spaces},
  author    = {Yang, Jihan and Yang, Shusheng and Gupta, Anjali W and Han, Rilyn and Fei-Fei, Li and Xie, Saining},
  booktitle = {Proceedings of the Computer Vision and Pattern Recognition Conference},
  pages     = {10632--10643},
  year      = {2025}
}

@article{gpt4scene,
  title   = {Gpt4scene: Understand 3d scenes from videos with vision-language models},
  author  = {Qi, Zhangyang and Zhang, Zhixiong and Fang, Ye and Wang, Jiaqi and Zhao, Hengshuang},
  journal = {arXiv preprint arXiv:2501.01428},
  year    = {2025}
}

@article{vlm_3r,
  title   = {Vlm-3r: Vision-language models augmented with instruction-aligned 3d reconstruction},
  author  = {Fan, Zhiwen and Zhang, Jian and Li, Renjie and Zhang, Junge and Chen, Runjin and Hu, Hezhen and Wang, Kevin and Qu, Huaizhi and Wang, Dilin and Yan, Zhicheng and others},
  journal = {arXiv preprint arXiv:2505.20279},
  year    = {2025}
}

@inproceedings{embspatial,
  title     = {Embspatial-bench: Benchmarking spatial understanding for embodied tasks with large vision-language models},
  author    = {Du, Mengfei and Wu, Binhao and Li, Zejun and Huang, Xuan-Jing and Wei, Zhongyu},
  booktitle = {Proceedings of the 62nd Annual Meeting of the Association for Computational Linguistics (Volume 2: Short Papers)},
  pages     = {346--355},
  year      = {2024}
}

@inproceedings{multi3drefer,
  title     = {Multi3drefer: Grounding text description to multiple 3d objects},
  author    = {Zhang, Yiming and Gong, ZeMing and Chang, Angel X},
  booktitle = {Proceedings of the IEEE/CVF International Conference on Computer Vision},
  pages     = {15225--15236},
  year      = {2023}
}

@article{mmsi,
  title   = {Mmsi-bench: A benchmark for multi-image spatial intelligence},
  author  = {Yang, Sihan and Xu, Runsen and Xie, Yiman and Yang, Sizhe and Li, Mo and Lin, Jingli and Zhu, Chenming and Chen, Xiaochen and Duan, Haodong and Yue, Xiangyu and others},
  journal = {arXiv preprint arXiv:2505.23764},
  year    = {2025}
}

@article{claude4,
  title  = {Claude Sonnet 4},
  author = {Anthropic},
  year   = {2025}
}

@article{shen2025efficient,
  title     = {Efficient and effective weight-ensembling mixture of experts for multi-task model merging},
  author    = {Shen, Li and Tang, Anke and Yang, Enneng and Guo, Guibing and Luo, Yong and Zhang, Lefei and Cao, Xiaochun and Du, Bo and Tao, Dacheng},
  journal   = {IEEE Transactions on Pattern Analysis and Machine Intelligence},
  year      = {2025},
  publisher = {IEEE}
}

@inproceedings{llava,
  title     = {Improved baselines with visual instruction tuning},
  author    = {Liu, Haotian and Li, Chunyuan and Li, Yuheng and Lee, Yong Jae},
  booktitle = {Proceedings of the IEEE/CVF conference on computer vision and pattern recognition},
  pages     = {26296--26306},
  year      = {2024}
}

@inproceedings{brohan2022rt1,
  title     = {{RT-1}: Robotics Transformer for Real-World Control at Scale},
  author    = {Brohan, Anthony and others},
  booktitle = {Robotics: Science and Systems (RSS)},
  year      = {2023}
}

@inproceedings{zitkovich2023rt2,
  title     = {{RT-2}: Vision-Language-Action Models Transfer Web Knowledge to Robotic Control},
  author    = {Zitkovich, Brianna and others},
  booktitle = {Conference on Robot Learning (CoRL)},
  year      = {2023}
}

@inproceedings{octo2024,
  title     = {Octo: An Open-Source Generalist Robot Policy},
  author    = {{Octo Model Team}},
  booktitle = {Robotics: Science and Systems (RSS)},
  year      = {2024}
}

@inproceedings{kim2024openvla,
  title     = {{OpenVLA}: An Open-Source Vision-Language-Action Model},
  author    = {Kim, Moo Jin and others},
  booktitle = {Conference on Robot Learning (CoRL)},
  year      = {2024}
}

@article{li2024cogact,
  title   = {{CogACT}: A Foundational Vision-Language-Action Model for Synergizing Cognition and Action in Robotic Manipulation},
  author  = {Li, Qixiu and others},
  journal = {arXiv preprint arXiv:2411.19650},
  year    = {2024}
}

@article{pi05_2025,
  title   = {{$\pi_{0.5}$}: A Vision-Language-Action Model with Open-World Generalization},
  author  = {{Physical Intelligence} and Black, Kevin and others},
  journal = {arXiv preprint arXiv:2504.16054},
  year    = {2025}
}

@article{gr3_2025,
  title   = {{GR-3} Technical Report},
  author  = {Cheang, Chilam and others},
  journal = {arXiv preprint arXiv:2507.15493},
  year    = {2025}
}

@article{pani2026gaze,
  title   = {Gaze-Regularized Vision-Language-Action Models for Robotic Manipulation},
  author  = {Pani, Anupam and others},
  journal = {arXiv preprint arXiv:2603.23202},
  year    = {2026}
}

@article{zhang2024vlnsurvey,
  title   = {Vision-and-Language Navigation with Foundation Models: A Survey},
  author  = {Zhang, Yue and others},
  journal = {Transactions on Machine Learning Research (TMLR)},
  year    = {2024}
}

@article{zhang2024uninavid,
  title   = {{Uni-NaVid}: A Video-Based Vision-Language-Action Model for Unifying Embodied Navigation Tasks},
  author  = {Zhang, Jiazhao and others},
  journal = {arXiv preprint arXiv:2412.06224},
  year    = {2024}
}

@article{sidvln,
  title   = {Learning Goal-Oriented Language-Guided Navigation with Self-Improving Demonstrations at Scale},
  author  = {Li, Songze and Wang, Zun and Zhou, Gengze and Li, Jialu and Zeng, Xiangyu and Wang, Limin and Qiao, Yu and Wu, Qi and Bansal, Mohit and Wang, Yi},
  journal = {arXiv preprint arXiv:2509.24910},
  year    = {2025}
}

@article{bai2025endowing,
  title={Endowing embodied agents with spatial reasoning capabilities for vision-and-language navigation},
  author={Bai, Qianqian and Chen, Zhongpu and Luo, Ling and Du, Huaming and Lei, Yuqian and Jiao, Ziyun},
  journal={arXiv preprint arXiv:2504.08806},
  year={2025}
}

@article{kueble2026ssg,
  title   = {Modernising Reinforcement Learning-Based Navigation for Embodied Semantic Scene Graph Generation},
  author  = {K{\"u}ble, Roman and others},
  journal = {arXiv preprint arXiv:2603.25415},
  year    = {2026}
}

@inproceedings{driess2023palme,
  title     = {{PaLM-E}: An Embodied Multimodal Language Model},
  author    = {Driess, Danny and others},
  booktitle = {International Conference on Machine Learning (ICML)},
  year      = {2023}
}

@inproceedings{chen2024spatialvlm,
  title     = {{SpatialVLM}: Endowing Vision-Language Models with Spatial Reasoning Capabilities},
  author    = {Chen, Boyuan and others},
  booktitle = {CVPR},
  year      = {2024}
}

@inproceedings{yuan2024robopoint,
  title     = {{RoboPoint}: A Vision-Language Model for Spatial Affordance Prediction for Robotics},
  author    = {Yuan, Wentao and others},
  booktitle = {Conference on Robot Learning (CoRL)},
  year      = {2024}
}

@article{zhou2025roborefer,
  title   = {{RoboRefer}: Towards Spatial Referring with Reasoning in Vision-Language Models for Robotics},
  author  = {Zhou, Enshen and others},
  journal = {arXiv preprint arXiv:2506.04308},
  year    = {2025}
}

@inproceedings{robocodex2024,
  title     = {{RoboCodeX}: Multimodal Code Generation for Robotic Behavior Synthesis},
  author    = {Mu, Yao and others},
  booktitle = {International Conference on Machine Learning (ICML)},
  year      = {2024}
}

@inproceedings{roboclip,
  title     = {{RoboCLIP}: One Demonstration is Enough to Learn Robot Policies},
  author    = {Sontakke, Sumedh A. and Zhang, Jesse and Arnold, S{\'e}bastien M. R. and Pertsch, Karl and Biyik, Erdem and Sadigh, Dorsa and Finn, Chelsea and Itti, Laurent},
  booktitle = {Advances in Neural Information Processing Systems},
  year      = {2023}
}

@inproceedings{ma2023vipuniversalvisualreward,
  title     = {{VIP}: Towards Universal Visual Reward and Representation via Value-Implicit Pre-Training},
  author    = {Ma, Yecheng Jason and Sodhani, Shagun and Jayaraman, Dinesh and Bastani, Osbert and Kumar, Vikash and Zhang, Amy},
  booktitle = {International Conference on Learning Representations},
  year      = {2023}
}

@article{komatsuzaki2022sparse,
  title={Sparse upcycling: Training mixture-of-experts from dense checkpoints},
  author={Komatsuzaki, Aran and Puigcerver, Joan and Lee-Thorp, James and Ruiz, Carlos Riquelme and Mustafa, Basil and Ainslie, Joshua and Tay, Yi and Dehghani, Mostafa and Houlsby, Neil},
  journal={arXiv preprint arXiv:2212.05055},
  year={2022}
}

@article{yang2026model,
  title={Model merging in llms, mllms, and beyond: Methods, theories, applications, and opportunities},
  author={Yang, Enneng and Shen, Li and Guo, Guibing and Wang, Xingwei and Cao, Xiaochun and Zhang, Jie and Tao, Dacheng},
  journal={ACM Computing Surveys},
  volume={58},
  number={8},
  pages={1--41},
  year={2026},
  publisher={ACM New York, NY}
}

@inproceedings{rocamonde2024visionlanguage,
  title     = {Vision-Language Models are Zero-Shot Reward Models for Reinforcement Learning},
  author    = {Rocamonde, Juan and Montesinos, Victoriano and Nava, Elvis and Perez, Ethan and Lindner, David},
  booktitle = {International Conference on Learning Representations},
  year      = {2024}
}

@inproceedings{ma2023liv,
  title     = {{LIV}: Language-Image Representations and Rewards for Robotic Control},
  author    = {Ma, Yecheng Jason and Kumar, Vikash and Zhang, Amy and Bastani, Osbert and Jayaraman, Dinesh},
  booktitle = {Proceedings of the 40th International Conference on Machine Learning},
  pages     = {23301--23320},
  year      = {2023}
}

@inproceedings{yang2024rank,
  title     = {{Rank2Reward}: Learning Shaped Reward Functions from Passive Video},
  author    = {Yang, Daniel and Tjia, Davin and Berg, Jacob and Damen, Dima and Agrawal, Pulkit and Gupta, Abhishek},
  booktitle = {Proceedings of the 2024 IEEE International Conference on Robotics and Automation},
  year      = {2024}
}

@inproceedings{ma2024generative,
  title     = {Vision Language Models are In-Context Value Learners},
  author    = {Ma, Yecheng Jason and Hejna, Joey and Wahid, Ayzaan and Fu, Chuyuan and Shah, Dhruv and Liang, Jacky and Xu, Zhuo and Kirmani, Sean and Xu, Peng and Driess, Danny and Xiao, Ted and Tompson, Jonathan and Bastani, Osbert and Jayaraman, Dinesh and Yu, Wenhao and Zhang, Tingnan and Sadigh, Dorsa and Xia, Fei},
  booktitle = {International Conference on Learning Representations},
  year      = {2025}
}

@inproceedings{zhang2025rewind,
  title     = {{ReWiND}: Language-Guided Rewards Teach Robot Policies without New Demonstrations},
  author    = {Zhang, Jiahui and Luo, Yusen and Anwar, Abrar and Sontakke, Sumedh Anand and Lim, Joseph J. and Thomason, Jesse and Biyik, Erdem and Zhang, Jesse},
  booktitle = {Proceedings of The 9th Conference on Robot Learning},
  pages     = {460--488},
  year      = {2025}
}

@inproceedings{chen2025sarm,
  title     = {{SARM}: Stage-Aware Reward Modeling for Long Horizon Robot Manipulation},
  author    = {Chen, Qianzhong and Yu, Justin and Schwager, Mac and Abbeel, Pieter and Shentu, Fred and Wu, Philipp},
  booktitle = {International Conference on Learning Representations},
  year      = {2026}
}

@article{chen2026topreward,
  title   = {{TOPReward}: Token Probabilities as Hidden Zero-Shot Rewards for Robotics},
  author  = {Chen, Shirui and Harrison, Cole and Lee, Ying-Chun and Yang, Angela Jin and Ren, Zhongzheng and Ratliff, Lillian J. and Duan, Jiafei and Fox, Dieter and Krishna, Ranjay},
  journal = {arXiv preprint arXiv:2602.19313},
  year    = {2026}
}

@article{zhai2025vision,
  title={A vision-language-action-critic model for robotic real-world reinforcement learning},
  author={Zhai, Shaopeng and Zhang, Qi and Zhang, Tianyi and Huang, Fuxian and Zhang, Haoran and Zhou, Ming and Zhang, Shengzhe and Liu, Litao and Lin, Sixu and Pang, Jiangmiao},
  journal={arXiv preprint arXiv:2509.15937},
  year={2025}
}

@article{lee2026roboreward,
  title   = {{RoboReward}: General-Purpose Vision-Language Reward Models for Robotics},
  author  = {Lee, Tony and others},
  journal = {arXiv preprint arXiv:2601.00675},
  year    = {2026}
}

@article{wang2025univla,
  title   = {Unified Vision-Language-Action Model},
  author  = {Wang, Yuqi and Li, Xinghang and Wang, Wenxuan and Zhang, Junbo and Li, Yingyan and Chen, Yuntao and Wang, Xinlong and Zhang, Zhaoxiang},
  journal = {arXiv preprint arXiv:2506.19850},
  year    = {2025}
}

@article{huang2025thinkact,
  title   = {ThinkAct: Vision-Language-Action Reasoning via Reinforced Visual Latent Planning},
  author  = {Huang, Chi-Pin and Wu, Yueh-Hua and Chen, Min-Hung and Wang, Yu-Chiang Frank and Yang, Fu-En},
  journal = {arXiv preprint arXiv:2507.16815},
  year    = {2025}
}

@article{zheng2025xvla,
  title   = {X-VLA: Soft-Prompted Transformer as Scalable Cross-Embodiment Vision-Language-Action Model},
  author  = {Zheng, Jinliang and Li, Jianxiong and Wang, Zhihao and Liu, Dongxiu and Kang, Xirui and Feng, Yuchun and Zheng, Yinan and Zou, Jiayin and Chen, Yilun and Zeng, Jia and others},
  journal = {arXiv preprint arXiv:2510.10274},
  year    = {2025}
}

@inproceedings{tan2026robodopamine,
  title     = {{Robo-Dopamine}: General Process Reward Modeling for High-Precision Robotic Manipulation},
  author    = {Tan, Huajie and Chen, Sixiang and Xu, Yijie and Wang, Zixiao and Ji, Yuheng and Chi, Cheng and Lyu, Yaoxu and Zhao, Zhongxia and Chen, Xiansheng and Co, Peterson and Xie, Shaoxuan and Yao, Guocai and Wang, Pengwei and Wang, Zhongyuan and Zhang, Shanghang},
  booktitle = {Proceedings of the IEEE/CVF Conference on Computer Vision and Pattern Recognition},
  year      = {2026}
}

@article{robometer2026,
  title   = {{RoboMeter}: Scaling General-Purpose Robotic Reward Models via Trajectory Comparisons},
  author  = {Liang, Anthony and others},
  journal = {arXiv preprint arXiv:2603.02115},
  year    = {2026}
}

@article{team2026hy,
  title   = {HY-Embodied-0.5: Embodied Foundation Models for Real-World Agents},
  author  = {Team, HY and Yu, Xumin and Liu, Zuyan and Wang, Ziyi and Zhang, He and Rao, Yongming and Liu, Fangfu and Zhang, Yani and Zhao, Ruowen and Wang, Oran and others},
  journal = {arXiv preprint arXiv:2604.07430},
  year    = {2026}
}

@article{duan2022survey,
  title   = {A Survey of Embodied AI: From Simulators to Research Tasks},
  author  = {Duan, Jiafei and Yu, Samson and Tan, Hui Li and Zhu, Hongyuan and Tan, Cheston},
  journal = {IEEE Transactions on Emerging Topics in Computational Intelligence},
  volume  = {6},
  number  = {2},
  pages   = {230--244},
  year    = {2022}
}

@article{community2026starvla,
  title={StarVLA: A Lego-like Codebase for Vision-Language-Action Model Developing},
  author={Community, StarVLA},
  journal={arXiv preprint arXiv:2604.05014},
  year={2026},
  eprint={2604.05014},
  archivePrefix={arXiv},
  primaryClass={cs.RO}
}

@article{yang2025mantis,
  title={Mantis: A Versatile Vision-Language-Action Model with Disentangled Visual Foresight},
  author={Yang, Yi and Li, Xueqi and Chen, Yiyang and Song, Jin and Wang, Yihan and Xiao, Zipeng and Su, Jiadi and Qiaoben, You and Liu, Pengfei and Deng, Zhijie},
  journal={arXiv preprint arXiv:2511.16175},
  year={2025}
}

@article{kim2025fine,
  title={Fine-Tuning Vision-Language-Action Models: Optimizing Speed and Success},
  author={Kim, Moo Jin and Finn, Chelsea and Liang, Percy},
  journal={arXiv preprint arXiv:2502.19645},
  year={2025}
}

@article{xu2024robotics,
  title   = {A Survey on Robotics with Foundation Models: Toward Embodied AI},
  author  = {Xu, Zhiyuan and Wu, Kun and Wen, Junjie and Li, Jinming and Liu, Ning and Che, Zhengping and Tang, Jian},
  journal = {arXiv preprint arXiv:2402.02385},
  year    = {2024}
}

@misc{deitke2024molmopixmoopenweights,
  title         = {Molmo and PixMo: Open Weights and Open Data for State-of-the-Art Vision-Language Models},
  author        = {Matt Deitke and Christopher Clark and Sangho Lee and Rohun Tripathi and Yue Yang and Jae Sung Park and Mohammadreza Salehi and Niklas Muennighoff and Kyle Lo and Luca Soldaini and Jiasen Lu and Taira Anderson and Erin Bransom and Kiana Ehsani and Huong Ngo and YenSung Chen and Ajay Patel and Mark Yatskar and Chris Callison-Burch and Andrew Head and Rose Hendrix and Favyen Bastani and Eli VanderBilt and Nathan Lambert and Yvonne Chou and Arnavi Chheda and Jenna Sparks and Sam Skjonsberg and Michael Schmitz and Aaron Sarnat and Byron Bischoff and Pete Walsh and Chris Newell and Piper Wolters and Tanmay Gupta and Kuo-Hao Zeng and Jon Borchardt and Dirk Groeneveld and Crystal Nam and Sophie Lebrecht and Caitlin Wittlif and Carissa Schoenick and Oscar Michel and Ranjay Krishna and Luca Weihs and Noah A. Smith and Hannaneh Hajishirzi and Ross Girshick and Ali Farhadi and Aniruddha Kembhavi},
  year          = {2024},
  eprint        = {2409.17146},
  archiveprefix = {arXiv},
  primaryclass  = {cs.CV},
  url           = {https://arxiv.org/abs/2409.17146}
}

@article{ling2026guide,
  title   = {Guide, Think, Act: Interactive Embodied Reasoning in Vision-Language-Action Models},
  author  = {Ling, Yiran and Lian, Qing and Li, Jinghang and Jiang, Qing and Zhang, Tianming and Jiang, Xiaoke and Liu, Chuanxiu and Liu, Jie and Zhang, Lei},
  journal = {arXiv preprint arXiv:2605.13632},
  year    = {2026}
}

@article{acebrain0,
  title   = {ACE-Brain-0: Spatial Intelligence as a Shared Scaffold for Universal Embodiments},
  author  = {Gong, Ziyang and Luo, Zehang and Tang, Anke and Liu, Zhe and Fu, Shi and Hou, Zhi and Yang, Ganlin and Wang, Weiyun and Wang, Xiaofeng and Liu, Jianbo and Luo, Gen and Kang, Haolan and Luo, Shuang and Zhou, Yue and Luo, Yong and Shen, Li and Jia, Xiaosong and Mu, Yao and Yang, Xue and Liu, Chunxiao and Yan, Junchi and Zhao, Hengshuang and Tao, Dacheng and Wang, Xiaogang},
  journal = {arXiv preprint arXiv:2603.03198},
  year    = {2026}
}

@article{capx,
  title   = {CaP-X: A Framework for Benchmarking and Improving Coding Agents for Robot Manipulation},
  author  = {Fu, Max and Yu, Justin and El-Refai, Karim and Kou, Ethan and Xue, Haoru and Huang, Huang and Xiao, Wenli and Wang, Guanzhi and Fei-Fei, Li and Shi, Guanya and Wu, Jiajun and Sastry, Shankar and Zhu, Yuke and Goldberg, Ken and Fan, Linxi},
  journal = {arXiv preprint arXiv:2603.22435},
  year    = {2026}
}

@inproceedings{nilsson1984shakey,
  title     = {Shakey the Robot},
  author    = {Nilsson, Nils J.},
  booktitle = {SRI International Technical Note 323},
  year      = {1984},
  publisher = {SRI International}
}

@article{brooks1986robust,
  title     = {A Robust Layered Control System for a Mobile Robot},
  author    = {Brooks, Rodney A.},
  journal   = {IEEE Journal on Robotics and Automation},
  volume    = {2},
  number    = {1},
  pages     = {14--23},
  year      = {1986},
  publisher = {IEEE}
}

@book{murphy2000introduction,
  title     = {Introduction to AI Robotics},
  author    = {Murphy, Robin R.},
  year      = {2000},
  publisher = {MIT Press}
}

@article{nvidia2025gr00tn1,
  title   = {GR00T N1: An Open Foundation Model for Generalist Humanoid Robots},
  author  = {NVIDIA and Bjorck, Johan and Casta{\~n}eda, Fernando and Cherniadev, Nikita and Da, Xingye and Ding, Runyu and Fan, Linxi and Fang, Yu and Fox, Dieter and Hu, Fengyuan and Huang, Spencer and Jang, Joel and Jiang, Zhenyu and Kautz, Jan and Kundalia, Kaushil and Lao, Lawrence and Li, Zhiqi and Lin, Zongyu and Lin, Kevin and Liu, Guilin and Llontop, Edith and Magne, Loic and Mandlekar, Ajay and Narayan, Avnish and Nasiriany, Soroush and Reed, Scott and Tan, You Liang and Wang, Guanzhi and Wang, Zu and Wang, Jing and Wang, Qi and Xiang, Jiannan and Xie, Yuqi and Xu, Yinzhen and Xu, Zhenjia and Ye, Seonghyeon and Yu, Zhiding and Zhang, Ao and Zhang, Hao and Zhao, Yizhou and Zheng, Ruijie and Zhu, Yuke},
  journal = {arXiv preprint arXiv:2503.14734},
  year    = {2025},
  doi     = {10.48550/arXiv.2503.14734},
  url     = {https://arxiv.org/abs/2503.14734}
}

@misc{nvidia2025gr00tn15,
  title        = {GR00T N1.5: An Improved Open Foundation Model for Generalist Humanoid Robots},
  author       = {{NVIDIA}},
  year         = {2025},
  howpublished = {\url{https://research.nvidia.com/labs/gear/gr00t-n1_5/}},
  note         = {NVIDIA GEAR Lab technical blog}
}

@inproceedings{ahn2022saycan,
  title     = {Do As I Can, Not As I Say: Grounding Language in Robotic Affordances},
  author    = {Ahn, Michael and Brohan, Anthony and Brown, Noah and Chebotar, Yevgen and Cortes, Omar and David, Byron and Finn, Chelsea and Fu, Chuyuan and Gopalakrishnan, Karol and Hausman, Karol and others},
  booktitle = {Conference on Robot Learning (CoRL)},
  year      = {2022}
}

@inproceedings{shinn2023reflexion,
  title     = {Reflexion: Language Agents with Verbal Reinforcement Learning},
  author    = {Shinn, Noah and Cassano, Federico and Gopinath, Ashwin and Narasimhan, Karthik and Yao, Shunyu},
  booktitle = {Advances in Neural Information Processing Systems (NeurIPS)},
  year      = {2023},
  note      = {arXiv:2303.11366}
}

@inproceedings{belkhale2024rth,
  title     = {{RT-H}: Action Hierarchies Using Language},
  author    = {Belkhale, Suneel and Ding, Tianli and Xiao, Ted and Sermanet, Pierre and Vuong, Quan and Tompson, Jonathan and Chebotar, Yevgen and Dwibedi, Debidatta and Sadigh, Dorsa},
  booktitle = {Robotics: Science and Systems (RSS)},
  year      = {2024},
  note      = {arXiv:2403.01823}
}

@inproceedings{ghasemipour2025selfimproving,
  title     = {Self-Improving Embodied Foundation Models},
  author    = {Ghasemipour, Seyed Kamyar Seyed and Wahid, Ayzaan and Tompson, Jonathan and Sanketi, Pannag and Mordatch, Igor},
  booktitle = {Advances in Neural Information Processing Systems (NeurIPS)},
  year      = {2025},
  note      = {arXiv:2509.15155}
}

@article{li2024simplerenv,
  title   = {Evaluating Real-World Robot Manipulation Policies in Simulation},
  author  = {Li, Xuanlin and Hsu, Kyle and Gu, Jiayuan and Pertsch, Karl and Mees, Oier and Walke, Homer Rich and Fu, Chuyuan and Lunawat, Ishikaa and Sieh, Isabel and Kirmani, Sean and Levine, Sergey and Wu, Jiajun and Finn, Chelsea and Su, Hao and Vuong, Quan and Xiao, Ted},
  journal = {arXiv preprint arXiv:2405.05941},
  year    = {2024}
}

@inproceedings{huang2023voxposer,
  title     = {VoxPoser: Composable 3D Value Maps for Robotic Manipulation with Language Models},
  author    = {Huang, Wenlong and Wang, Chen and Zhang, Ruohan and Li, Yunzhu and Wu, Jiajun and Fei-Fei, Li},
  booktitle = {Conference on Robot Learning (CoRL)},
  year      = {2023}
}

@inproceedings{liu2023libero,
  title     = {LIBERO: Benchmarking Knowledge Transfer for Lifelong Robot Learning},
  author    = {Liu, Bo and Zhu, Yifeng and Gao, Chongkai and Feng, Yihao and Liu, Qiang and Zhu, Yuke and Stone, Peter},
  booktitle = {Advances in Neural Information Processing Systems (NeurIPS)},
  year      = {2023}
}

@article{li2025reflective,
  title   = {Reflection-Based Task Adaptation for Self-Improving {VLA}},
  author  = {Li, Baicheng and others},
  journal = {arXiv preprint arXiv:2510.12710},
  year    = {2025}
}

@article{rise2026,
  title   = {{RISE}: Self-Improving Robot Policy with Compositional World Model},
  author  = {Yang, Jiazhi and Lin, Kunyang and Li, Jinwei and Zhang, Wencong and Lin, Tianwei and Wu, Longyan and Su, Zhizhong and Zhao, Hao and Zhang, Ya-Qin and Chen, Li and Luo, Ping and Yue, Xiangyu and Li, Hongyang},
  journal = {arXiv preprint arXiv:2602.11075},
  year    = {2026}
}

@article{xu2026roboagent,
  title   = {{RoboAgent}: Chaining Basic Capabilities for Embodied Task Planning},
  author  = {Xu, Peiran and Zheng, Jiaqi and Mu, Yadong},
  journal = {arXiv preprint arXiv:2604.07774},
  year    = {2026}
}

@article{yuan2025agentr,
  title   = {{Agent-R}: Training Language Model Agents to Reflect via Iterative Self-Training},
  author  = {Yuan, Siyu and Chen, Zehui and Xi, Zhiheng and Ye, Junjie and Du, Zhengyin and Chen, Jiecao},
  journal = {arXiv preprint arXiv:2501.11425},
  year    = {2025}
}

@inproceedings{tian2025seear1,
  title     = {{SEEA-R1}: Tree-Structured Reinforcement Fine-Tuning for Self-Evolving Embodied Agents},
  author    = {Tian, Wanxin and Zhang, Shijie and Zhang, Kevin and Chi, Xiaowei and Luo, Yulin and Lu, Junyu and Fan, Chunkai and Zhou, Qiang and Zhao, Yiming and Lin, Siyu and Qin, Zhiyuan and Ju, Xiaozhu and Zhang, Shanghang and Tang, Jian},
  booktitle = {Advances in Neural Information Processing Systems (NeurIPS)},
  year      = {2025},
  note      = {arXiv:2506.21669}
}

@inproceedings{lipman2023flow,
  title     = {Flow Matching for Generative Modeling},
  author    = {Lipman, Yaron and Chen, Ricky T. Q. and Ben-Hamu, Heli and Nickel, Maximilian and Le, Matt},
  booktitle = {International Conference on Learning Representations (ICLR)},
  year      = {2023}
}

@inproceedings{open_x_embodiment_rt_x_2023,
  title     = {Open {X-Embodiment}: Robotic Learning Datasets and {RT-X} Models},
  author    = {Open X-Embodiment Collaboration and Abby O'Neill and Abdul Rehman and Abhinav Gupta and ... and Zipeng Lin and Zubair Irshad},
  booktitle = {2024 IEEE International Conference on Robotics and Automation (ICRA)},
  year      = {2024}
}

@article{simeoni2025dinov3,
  title   = {DINOv3},
  author  = {Sim{\'e}oni, Oriane and Vo, Huy V. and Seitzer, Maximilian and others},
  journal = {arXiv preprint arXiv:2508.10104},
  year    = {2025}
}

@inproceedings{liang2023code,
  title     = {Code as Policies: Language Model Programs for Embodied Control},
  author    = {Liang, Jacky and Huang, Wenlong and Xia, Fei and Xu, Peng and Hausman, Karol and Ichter, Brian and Florence, Pete and Zeng, Andy},
  booktitle = {IEEE International Conference on Robotics and Automation (ICRA)},
  pages     = {9493--9500},
  year      = {2023}
}

@misc{mindcube,
  title         = {MindCube: Spatial Mental Modeling from Limited Views},
  author        = {Qineng Wang and Baiqiao Yin and Pingyue Zhang and Jianshu Zhang and Kangrui Wang and Zihan Wang and Jieyu Zhang and Keshigeyan Chandrasegaran and Han Liu and Ranjay Krishna and Saining Xie and Jiajun Wu and Li Fei-Fei and Manling Li},
  year          = {2026},
  eprint        = {2506.21458},
  archiveprefix = {arXiv},
  primaryclass  = {cs.AI},
  url           = {https://arxiv.org/abs/2506.21458}
}

@misc{sqa3d,
  title         = {SQA3D: Situated Question Answering in 3D Scenes},
  author        = {Xiaojian Ma and Silong Yong and Zilong Zheng and Qing Li and Yitao Liang and Song-Chun Zhu and Siyuan Huang},
  year          = {2023},
  eprint        = {2210.07474},
  archiveprefix = {arXiv},
  primaryclass  = {cs.CV},
  url           = {https://arxiv.org/abs/2210.07474}
}

@misc{scanqa,
  title         = {ScanQA: 3D Question Answering for Spatial Scene Understanding},
  author        = {Daichi Azuma and Taiki Miyanishi and Shuhei Kurita and Motoaki Kawanabe},
  year          = {2022},
  eprint        = {2112.10482},
  archiveprefix = {arXiv},
  primaryclass  = {cs.CV},
  url           = {https://arxiv.org/abs/2112.10482}
}

@misc{scanrefer,
  title         = {ScanRefer: 3D Object Localization in RGB-D Scans using Natural Language},
  author        = {Dave Zhenyu Chen and Angel X. Chang and Matthias Nießner},
  year          = {2020},
  eprint        = {1912.08830},
  archiveprefix = {arXiv},
  primaryclass  = {cs.CV},
  url           = {https://arxiv.org/abs/1912.08830}
}

@misc{scan2cap,
  title         = {Scan2Cap: Context-aware Dense Captioning in RGB-D Scans},
  author        = {Dave Zhenyu Chen and Ali Gholami and Matthias Nießner and Angel X. Chang},
  year          = {2020},
  eprint        = {2012.02206},
  archiveprefix = {arXiv},
  primaryclass  = {cs.CV},
  url           = {https://arxiv.org/abs/2012.02206}
}

@misc{pointarena,
  title         = {PointArena: Probing Multimodal Grounding Through Language-Guided Pointing},
  author        = {Long Cheng and Jiafei Duan and Yi Ru Wang and Haoquan Fang and Boyang Li and Yushan Huang and Elvis Wang and Ainaz Eftekhar and Jason Lee and Wentao Yuan and Rose Hendrix and Noah A. Smith and Fei Xia and Dieter Fox and Ranjay Krishna},
  year          = {2025},
  eprint        = {2505.09990},
  archiveprefix = {arXiv},
  primaryclass  = {cs.CV},
  url           = {https://arxiv.org/abs/2505.09990}
}

@misc{skillopt,
  title         = {SkillOpt: Executive Strategy for Self-Evolving Agent Skills},
  author        = {Yifan Yang and Ziyang Gong and Weiquan Huang and Qihao Yang and Ziwei Zhou and Zisu Huang and Yan Li and Xuemei Gao and Qi Dai and Bei Liu and Kai Qiu and Yuqing Yang and Dongdong Chen and Xue Yang and Chong Luo},
  year          = {2026},
  eprint        = {2605.23904},
  archiveprefix = {arXiv},
  primaryclass  = {cs.AI},
  url           = {https://arxiv.org/abs/2605.23904}
}

@misc{skilllens,
  title         = {From Raw Experience to Skill Consumption: A Systematic Study of Model-Generated Agent Skills},
  author        = {Zisu Huang and Jingwen Xu and Yifan Yang and Ziyang Gong and Qihao Yang and Muzhao Tian and Xiaohua Wang and Changze Lv and Xuemei Gao and Qi Dai and Bei Liu and Kai Qiu and Xue Yang and Dongdong Chen and Xiaoqing Zheng and Chong Luo},
  year          = {2026},
  eprint        = {2605.23899},
  archiveprefix = {arXiv},
  primaryclass  = {cs.AI},
  url           = {https://arxiv.org/abs/2605.23899}
}

@misc{continualharness,
  title         = {Continual Harness: Online Adaptation for Self-Improving Foundation Agents},
  author        = {Seth Karten and Joel Zhang and Tersoo Upaa Jr and Ruirong Feng and Wenzhe Li and Chengshuai Shi and Chi Jin and Kiran Vodrahalli},
  year          = {2026},
  eprint        = {2605.09998},
  archiveprefix = {arXiv},
  primaryclass  = {cs.LG},
  url           = {https://arxiv.org/abs/2605.09998}
}

@misc{gca,
  title         = {Geometrically-Constrained Agent for Spatial Reasoning},
  author        = {Zeren Chen and Xiaoya Lu and Zhijie Zheng and Pengrui Li and Lehan He and Yijin Zhou and Jing Shao and Bohan Zhuang and Lu Sheng},
  year          = {2025},
  eprint        = {2511.22659},
  archiveprefix = {arXiv},
  primaryclass  = {cs.AI},
  url           = {https://arxiv.org/abs/2511.22659}
}

@article{zhang2024navid,
  title   = {NaVid: Video-based VLM Plans the Next Step for Vision-and-Language Navigation},
  author  = {Zhang, Jiazhao and Wang, Kunyu and Xu, Rongtao and Zhou, Gengze and Hong, Yicong and Fang, Xiaomeng and Wu, Qi and Zhang, Zhizheng and Wang, He},
  journal = {arXiv preprint arXiv:2402.15852},
  year    = {2024}
}

@article{cheng2024navila,
  title   = {NaVILA: Legged Robot Vision-Language-Action Model for Navigation},
  author  = {Cheng, An-Chieh and Ji, Yandong and Yang, Zhaojing and Gongye, Zaitian and Zou, Xueyan and Kautz, Jan and B{\i}y{\i}k, Erdem and Yin, Hongxu and Liu, Sifei and Wang, Xiaolong},
  journal = {arXiv preprint arXiv:2412.04453},
  year    = {2024}
}

@article{wei2025streamvln,
  title   = {StreamVLN: Streaming Vision-and-Language Navigation via SlowFast Context Modeling},
  author  = {Wei, Meng and Wan, Chenyang and Yu, Xiqian and Wang, Tai and Yang, Yuqiang and Mao, Xiaohan and Zhu, Chenming and Cai, Wenzhe and Wang, Hanqing and Chen, Yilun and Liu, Xihui and Pang, Jiangmiao},
  journal = {arXiv preprint arXiv:2507.05240},
  year    = {2025}
}

@article{zhang2025navfom,
  title   = {Embodied Navigation Foundation Model},
  author  = {Zhang, Jiazhao and Li, Anqi and Qi, Yunpeng and Li, Minghan and Liu, Jiahang and Wang, Shaoan and Liu, Haoran and Zhou, Gengze and Wu, Yuze and Li, Xingxing and others},
  journal = {arXiv preprint arXiv:2509.12129},
  year    = {2025}
}

@inproceedings{vlnce,
  title     = {Beyond the Nav-Graph: Vision-and-Language Navigation in Continuous Environments},
  author    = {Krantz, Jacob and Wijmans, Erik and Majumdar, Arjun and Batra, Dhruv and Lee, Stefan},
  booktitle = {European Conference on Computer Vision},
  year      = {2020}
}

@inproceedings{r2r,
  title     = {Vision-and-Language Navigation: Interpreting Visually-Grounded Navigation Instructions in Real Environments},
  author    = {Anderson, Peter and Wu, Qi and Teney, Damien and Bruce, Jake and Johnson, Mark and Sunderhauf, Niko and Reid, Ian and Gould, Stephen and van den Hengel, Anton},
  booktitle = {Proceedings of the IEEE Conference on Computer Vision and Pattern Recognition},
  year      = {2018}
}

@inproceedings{rxr,
  title     = {Room-Across-Room: Multilingual Vision-and-Language Navigation with Dense Spatiotemporal Grounding},
  author    = {Ku, Alexander and Anderson, Peter and Patel, Roma and Ie, Eugene and Baldridge, Jason},
  booktitle = {Proceedings of the Conference on Empirical Methods in Natural Language Processing},
  year      = {2020}
}

@article{envdrop,
  title={Learning to Navigate Unseen Environments: Back Translation with Environmental Dropout},
  author={Tan, Hao and Yu, Licheng and Bansal, Mohit},
  journal={arXiv preprint arXiv:1904.04195},
  year={2019}
}

@inproceedings{scalevln,
  title={Scaling Data Generation in Vision-and-Language Navigation},
  author={Wang, Zun and Li, Jialu and Hong, Yicong and Wang, Yi and Wu, Qi and Bansal, Mohit and Gould, Stephen and Tan, Hao and Qiao, Yu},
  booktitle={Proceedings of the IEEE/CVF International Conference on Computer Vision},
  year={2023}
}

@article{wang2024srdf,
  title={Bootstrapping Language-Guided Navigation Learning with Self-Refining Data Flywheel},
  author={Wang, Zun and Li, Jialu and Hong, Yicong and Li, Songze and Li, Kunchang and Yu, Shoubin and Wang, Yi and Qiao, Yu and Wang, Yali and Bansal, Mohit and Wang, Limin},
  journal={arXiv preprint arXiv:2412.08467},
  year={2024}
}

@article{qing2024a2po,
  title={A2po: Towards effective offline reinforcement learning from an advantage-aware perspective},
  author={Qing, Yunpeng and Liu, Shunyu and Cong, Jingyuan and Chen, Kaixuan and Zhou, Yihe and Song, Mingli},
  journal={Advances in Neural Information Processing Systems},
  volume={37},
  pages={29064--29090},
  year={2024}
}

@article{qing2025bitrajdiff,
  title={Bitrajdiff: Bidirectional trajectory generation with diffusion models for offline reinforcement learning},
  author={Qing, Yunpeng and Chi, Yixiao and Chen, Shuo and Liu, Shunyu and Yao, Kelu and Lin, Sixu and Liu, Litao and Zou, Changqing},
  journal={arXiv preprint arXiv:2506.05762},
  year={2025}
}

@article{liu2025curricular,
  title={Curricular subgoals for inverse reinforcement learning},
  author={Liu, Shunyu and Qing, Yunpeng and Xu, Shuqi and Wu, Hongyan and Zhang, Jiangtao and Cong, Jingyuan and Chen, Tianhao and Liu, Yun-Fu and Song, Mingli},
  journal={IEEE Transactions on Intelligent Transportation Systems},
  volume={26},
  number={3},
  pages={3016--3027},
  year={2025},
  publisher={IEEE}
}

@article{qing2022survey,
  title={A survey on explainable reinforcement learning: Concepts, algorithms, challenges},
  author={Qing, Yunpeng and Liu, Shunyu and Song, Jie and Zhou, Yang and Chen, Kaixuan and Wang, Huiqiong and Song, Mingli},
  journal={arXiv preprint arXiv:2211.06665},
  year={2022}
}

@inproceedings{kong2024tptu,
  title={Tptu-v2: Boosting task planning and tool usage of large language model-based agents in real-world industry systems},
  author={Kong, Yilun and Ruan, Jingqing and Chen, Yihong and Zhang, Bin and Bao, Tianpeng and Shiwei, Shi and Hu, Xiaoru and Mao, Hangyu and Li, Ziyue and Zeng, Xingyu and others},
  booktitle={Proceedings of the 2024 conference on empirical methods in natural language processing: industry track},
  pages={371--385},
  year={2024}
}

@article{kong2025mastering,
  title={Mastering massive multi-task reinforcement learning via mixture-of-expert decision transformer},
  author={Kong, Yilun and Ma, Guozheng and Zhao, Qi and Wang, Haoyu and Shen, Li and Wang, Xueqian and Tao, Dacheng},
  journal={arXiv preprint arXiv:2505.24378},
  year={2025}
}

@article{kong2024qpo,
  title={Qpo: Query-dependent prompt optimization via multi-loop offline reinforcement learning},
  author={Kong, Yilun and Mao, Hangyu and Zhao, Qi and Zhang, Bin and Ruan, Jingqing and Shen, Li and Chang, Yongzhe and Wang, Xueqian and Zhao, Rui and Tao, Dacheng},
  journal={arXiv preprint arXiv:2408.10504},
  year={2024}
}

@inproceedings{chen2022think,
  title={Think global, act local: Dual-scale graph transformer for vision-and-language navigation},
  author={Chen, Shizhe and Guhur, Pierre-Louis and Tapaswi, Makarand and Schmid, Cordelia and Laptev, Ivan},
  booktitle={Proceedings of the IEEE/CVF conference on computer vision and pattern recognition},
  pages={16537--16547},
  year={2022}
}

@inproceedings{zhou2024navgpt,
  title={Navgpt: Explicit reasoning in vision-and-language navigation with large language models},
  author={Zhou, Gengze and Hong, Yicong and Wu, Qi},
  booktitle={Proceedings of the AAAI Conference on Artificial Intelligence},
  volume={38},
  number={7},
  pages={7641--7649},
  year={2024}
}

@inproceedings{gao2026octonav,
  title={Octonav: Towards generalist embodied navigation},
  author={Gao, Chen and Jin, Liankai and Peng, Xingyu and Zhang, Jiazhao and Deng, Yue and Li, Annan and Wang, He and Liu, Si},
  booktitle={Proceedings of the IEEE/CVF Conference on Computer Vision and Pattern Recognition},
  pages={40074--40084},
  year={2026}
}

@article{xin2026agentvln,
  title={Agentvln: Towards agentic vision-and-language navigation},
  author={Xin, Zihao and Li, Wentong and Jiang, Yixuan and Huang, Ziyuan and Wang, Bin and Li, Piji and Zhu, Jianke and Qin, Jie and Huang, Shengjun},
  journal={arXiv preprint arXiv:2603.17670},
  year={2026}
}

@article{cen2025rynnvla,
  title={Rynnvla-002: A unified vision-language-action and world model},
  author={Cen, Jun and Huang, Siteng and Yuan, Yuqian and Li, Kehan and Yuan, Hangjie and Yu, Chaohui and Jiang, Yuming and Guo, Jiayan and Li, Xin and Luo, Hao and others},
  journal={arXiv preprint arXiv:2511.17502},
  year={2025}
}

@article{luo2026being,
  title={Being-h0. 7: A latent world-action model from egocentric videos},
  author={Luo, Hao and Zhang, Wanpeng and Feng, Yicheng and Zheng, Sipeng and Xu, Haiweng and Xu, Chaoyi and Xi, Ziheng and Fu, Yuhui and Lu, Zongqing},
  journal={arXiv preprint arXiv:2605.00078},
  year={2026}
}

@misc{chen2026abotm05,
      title={ABot-M0.5: Unified Mobility-and-Manipulation World Action Model}, 
      author={Ronghan Chen and Yandan Yang and Zuojin Tang and Dongjie Huo and Tong Lin and Haoning Wu and Haoyun Liu and Yuzhi Chen and Lulu Zheng and Botai Yuan and Tianlun Li and Mingxin Wang and Dekang Qi and Bin Hu and Wei Mei and Yuze Xuan and Haolong Yang and Yanqing Zhu and Mu Xu and Zhiheng Ma and Xinyuan Chang},
      year={2026},
      eprint={2607.00678},
      archivePrefix={arXiv},
      primaryClass={cs.CV},
      url={https://arxiv.org/abs/2607.00678}, 
}

@article{zhang2026dreamvla,
  title={Dreamvla: a vision-language-action model dreamed with comprehensive world knowledge},
  author={Zhang, Wenyao and Liu, Hongsi and Qi, Zekun and Wang, Yunnan and Yu, Xinqiang and Zhang, Jiazhao and Dong, Runpei and He, Jiawei and Wang, He and Zhang, Zhizheng and others},
  journal={Advances in Neural Information Processing Systems},
  volume={38},
  pages={24195--24228},
  year={2026}
}
